\documentclass[10pt,journal,cspaper,compsoc]{IEEEtran}
\ifCLASSOPTIONcompsoc
\usepackage[nocompress]{cite}
\else
\usepackage{cite}
\fi

\ifCLASSINFOpdf
\else
\fi

\usepackage{graphicx}
\usepackage{amsmath,amssymb} % define this before the line numbering.
\usepackage{color}
\usepackage{paralist}
\usepackage{booktabs}
\usepackage{times}
\usepackage{epsfig}

\usepackage{lineno}
\usepackage{algorithmic}
\usepackage{algorithm}
\usepackage{url}
\usepackage{listings}
\usepackage{mathrsfs}
\usepackage{array}
\usepackage{subfigure}
\usepackage{amsthm}
\usepackage{multirow}

\newcommand{\revyj}[1]{\textcolor{black}{#1}} %magenta
\newcommand{\revyjnew}[1]{\textcolor{black}{#1}} %magenta

\newcommand{\revyjnewminor}[1]{\textcolor{black}{#1}} %magenta

\begin{document}

\title{Joint Image Filtering with \\ Deep Convolutional Networks}

\author{Yijun~Li,~
Jia-Bin~Huang,
Narendra~Ahuja,
and~Ming-Hsuan~Yang~% <-this % stops a space

\IEEEcompsocitemizethanks{
\IEEEcompsocthanksitem Yijun Li is with School of Engineering, University of California, Merced, CA, US, E-mail: yli62@ucmerced.edu

\IEEEcompsocthanksitem Jia-Bin Huang is with Department of Electrical and Computer Engineering, Virginia Tech, VA, US, E-mail: jbhuang@vt.edu

\IEEEcompsocthanksitem Narendra Ahuja is with Department of Electrical and Computer Engineering, University of Illinois at Urbana-Champaign, IL, US, E-mail: n-ahuja@illinois.edu

\IEEEcompsocthanksitem Ming-Hsuan Yang is with School of Engineering, University of California, Merced, CA, US, E-mail: mhyang@ucmerced.edu
}
}

% ^------------^------------^----Do not want these spaces!

\markboth{IEEE Transactions on Pattern Analysis and Machine Intelligence}{}
% *** name in the headers of peer review papers. ***



\IEEEtitleabstractindextext{%
\begin{abstract}
Joint image filters leverage the guidance image as a prior and transfer the structural details from the guidance image to the target image for suppressing noise or enhancing spatial resolution.
Existing methods either rely on various explicit filter constructions
or hand-designed objective functions, thereby making it
difficult to understand, improve, and accelerate these filters in a coherent framework.
In this paper, we propose a learning-based approach for constructing joint filters based on Convolutional Neural Networks.
In contrast to existing methods that consider only the guidance image, the proposed algorithm can selectively transfer salient structures that are consistent with both guidance and target images.
We show that the model trained on a certain type of data, e.g., RGB and depth images, generalizes well to other modalities, e.g., flash/non-Flash and RGB/NIR images.
We validate the effectiveness of the proposed joint filter through extensive experimental evaluations with state-of-the-art methods.
\end{abstract}

% Note that keywords are not normally used for peer-review papers.
}

\maketitle

\IEEEdisplaynontitleabstractindextext

% \fi
\IEEEpeerreviewmaketitle

\IEEEraisesectionheading{\section{Introduction}\label{sec:introduction}}

\begin{figure*}[t]
\centering
\begin{tabular}{c@{\hspace{0.005\linewidth}}c@{\hspace{0.005\linewidth}}c@{\hspace{0.005\linewidth}}c@{\hspace{0.005\linewidth}}c@{\hspace{0.005\linewidth}}c@{\hspace{0.005\linewidth}}c@{\hspace{0.005\linewidth}}c@{\hspace{0.005\linewidth}}c@{\hspace{0.005\linewidth}}c@{\hspace{0.005\linewidth}}c@{\hspace{0.005\linewidth}}c@{\hspace{0.005\linewidth}}c}
\includegraphics[height=.12\linewidth, width = .156\linewidth]{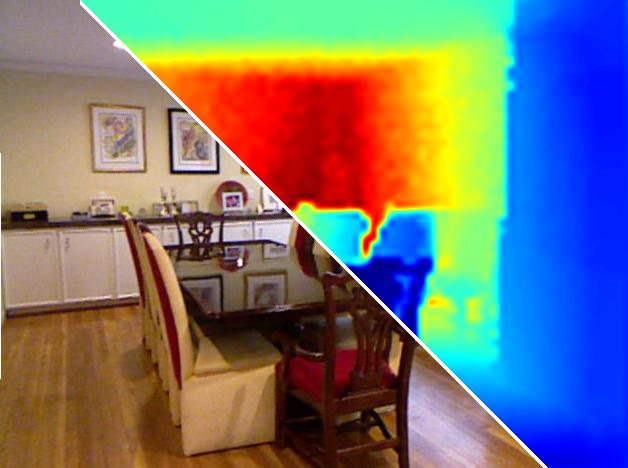} &
\includegraphics[height=.12\linewidth, width = .156\linewidth]{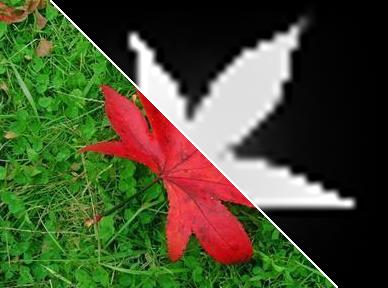} &
\includegraphics[height=.12\linewidth, width = .156\linewidth]{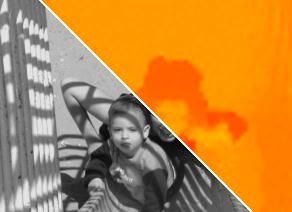} &
\includegraphics[height=.12\linewidth, width = .156\linewidth]{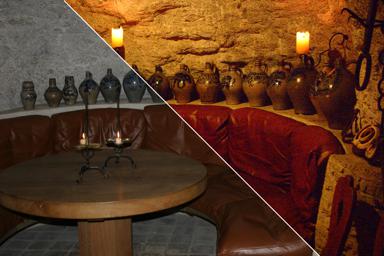} &
\includegraphics[height=.12\linewidth, width = .156\linewidth]{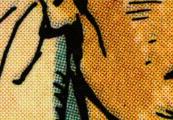} &
\includegraphics[height=.12\linewidth, width = .156\linewidth]{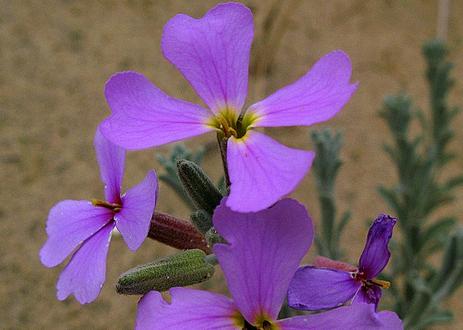} & \\
\includegraphics[height=.12\linewidth, width = .156\linewidth]{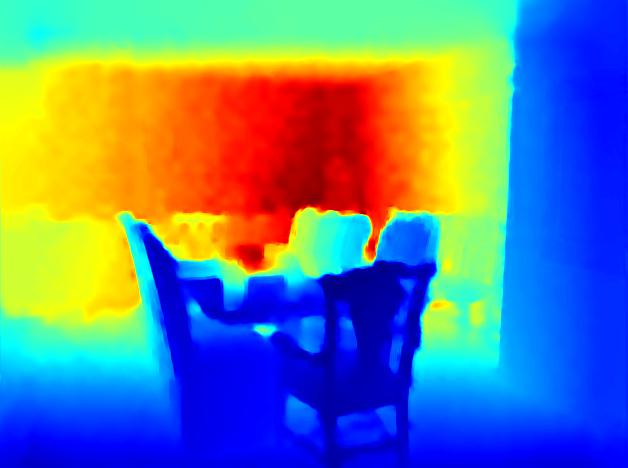} &
\includegraphics[height=.12\linewidth, width = .156\linewidth]{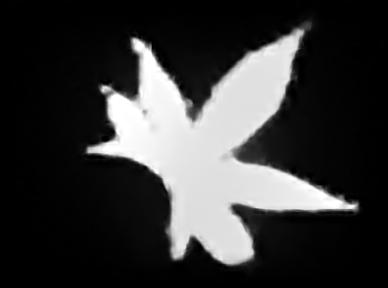} &
\includegraphics[height=.12\linewidth, width = .156\linewidth]{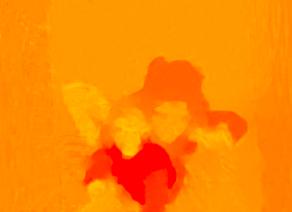} &
\includegraphics[height=.12\linewidth, width = .156\linewidth]{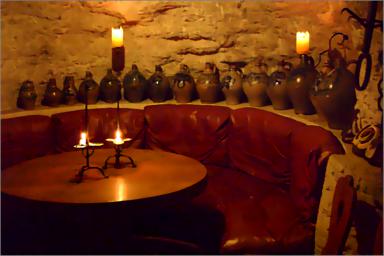} &
\includegraphics[height=.12\linewidth, width = .156\linewidth]{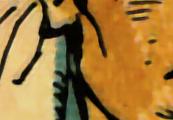} &
\includegraphics[height=.12\linewidth, width = .156\linewidth]{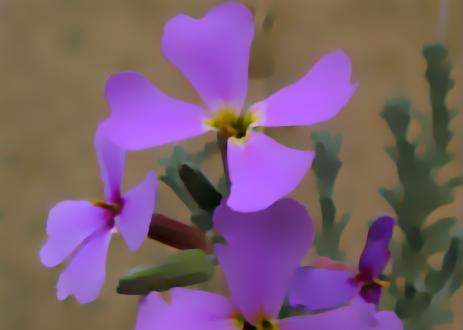} & \\
{Depth map} & {Saliency map} & {Chromaticity map} & {Flash/Non-Flash} & {Inverse} & {Texture} \\
{upsampling} & {upsampling} & {upsampling} & {noise reduction} & {halftoning} & {removal} \\

\end{tabular}
\caption{
\textbf{Sample applications of joint image filtering.}
The target/guidance pair (top) can be various types of cross-modality visual data.
With the help of the guidance image, important structures can be transferred to the degraded target image to help enhance the spatial resolution or suppress noises (bottom).
The guidance image can either be high-resolution RGB images, images from different sensing modalities, or the target image itself.
}
\label{fig:Task}
\end{figure*}

\IEEEPARstart{I}{mage} filtering with guidance signals, known as \emph{joint} or \emph{guided filtering}, has been successfully applied to numerous computer vision and computer graphics tasks, such as depth map enhancement~\cite{Yang-CVPR-2007,Park-ICCV-2011,TGV-ICCV-2013}, joint upsampling~\cite{JBU-TOG-2007,Yang-CVPR-2007}, cross-modality noise reduction~\cite{NIR-ICCV-2013,He-PAMI-2013,Shen-ICCV-2015}, and structure-texture separation~\cite{Xu-TOG-2012,Rolling-ECCV-2014}.
The wide applicability of joint filters can be attributed to the adaptability in handling visual signals in various image domains and modalities, as shown in Figure~\ref{fig:Task}.
For a target image, the guidance image can either be the target image itself~\cite{BF-ICCV-1998,He-PAMI-2013}, high-resolution RGB images~\cite{He-PAMI-2013,Park-ICCV-2011,TGV-ICCV-2013}, images from different sensing modalities~\cite{Eisemann-TOG-2004,Details-TOG-2004,NIR-ICCV-2013}, or filter outputs from previous iterations~\cite{Rolling-ECCV-2014}.
The basic idea behind joint image filtering is that we can transfer the important structural details contained in the guidance image to the target image.
The main goal of joint filtering is to enhance the degraded target image due to noise or low spatial resolution while avoiding transferring extraneous structures that do not originally exist
in the target image, e.g., texture-copying artifacts.

Several approaches have been developed to transfer structures in the guidance image to the target image.
One category of algorithms is to construct joint filters for specific tasks.
For example, the bilateral filtering algorithm~\cite{BF-ICCV-1998} constructs spatially-varying filters that reflect local image structures (e.g., smooth regions, edges, textures) in the guidance image.
Such filters can then be applied to the target image for edge-aware smoothing~\cite{BF-ICCV-1998} or joint upsampling~\cite{JBU-TOG-2007}.
On the other hand, the guided image filter~\cite{He-PAMI-2013}
assumes a locally linear model over the guidance image for filtering.
However, these filters share one common drawback.
That is, the filter construction considers only the information contained in
the guidance image and remains fixed (i.e., static guidance).
When the local structures in the guidance and target images are not consistent, these methods may transfer incorrect or extraneous contents to the target image.

To address this issue, recent efforts focus on considering the common structures existing in both the target and guidance images~\cite{Rolling-ECCV-2014,Ham-CVPR-2015,Shen-ICCV-2015}.
These frameworks typically build on iterative methods for minimizing global objective functions.
The guidance signals are updated at each iteration (i.e., dynamic guidance) towards preserving the mutually consistent structures while suppressing contents that are not commonly shared in both images.
However, these global optimization based methods often use hand-crafted objective functions that may not reflect natural image priors well and typically require a heavy computational load.

\begin{figure*}[t!]
\begin{center}
\includegraphics[width=.8\textwidth]{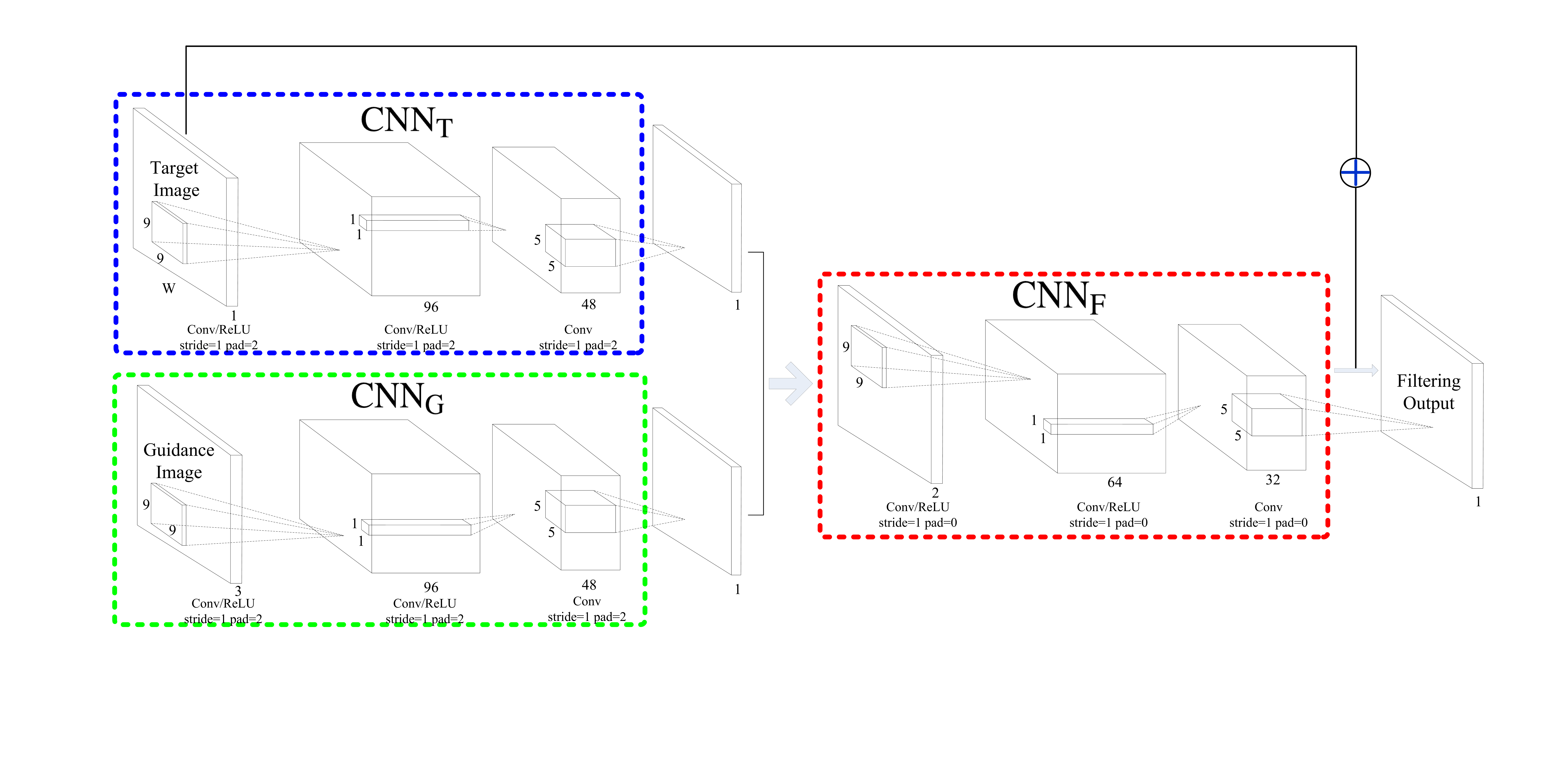}
\end{center}
\caption{
\textbf{Network architecture for joint image filter.}
The proposed deep joint image filter model consists of three major components.
Each component is a three-layer network.
The sub-networks $\mathrm{CNN_T}$ and $\mathrm{CNN_G}$ aim to extract informative feature responses from the target and guidance images, respectively.
We then concatenate these features responses together and use them as input for the network $\mathrm{CNN_F}$.
In addition, we introduce a skip connection so that the network $\mathrm{CNN_F}$ learns to predict the residuals between the input target image and the desired ground truth output.
We train the network to selectively transfer main structures while suppressing inconsistent structures using an RGB/depth dataset.
While we describe these sub-networks individually, the parameters of all three sub-networks are updated simultaneously during the training stage.
}
\label{fig:framework}
\end{figure*}

In this work, we propose a learning-based joint filter based on Convolutional Neural Networks (CNNs).
We propose a network architecture that consists of three sub-networks and a skip connection, as shown in Figure~\ref{fig:framework}.
The first two sub-networks $\mathrm{CNN_T}$ and $\mathrm{CNN_G}$
extract informative features from both target and guidance images.
These feature responses are then concatenated as inputs for the network $\mathrm{CNN_F}$ to selectively transfer common structures.
As the target input and output images are largely similar, we introduce a skip connection, together with the output of $\mathrm{CNN_F}$ to reconstruct the filtered output.
In other words, we enforce the network to focus on learning the residuals between the degraded target and the ground truth images.
We train the network using large quantities of RGB/depth data and learn all the network parameters simultaneously without stage-wise training.

Our algorithm differs from existing methods in that the proposed joint image filter is
purely data-driven.
This
allows the network to handle complicated scenarios that may be difficult to capture through hand-crafted objective functions.
While the network is trained using the RGB/depth data, the network learns how to selectively transfer structures by leveraging the prior from the guidance image, rather than predicting specific values.
As a result, the learned network generalizes well for handling images in various domains and modalities.

We make the following contributions in this paper:
\begin{itemize}
\item We propose a learning-based framework for constructing a generic joint image filter.
Our network takes both target and guidance images into consideration and naturally handles the inconsistent structure problem.
\item \revyj{Using the learned joint image filter for depth upsampling, we demonstrate the state-of-the-art performance on the NYU 
v2~\cite{NYU-ECCV-2012} and SUN RGB-D~\cite{Song-CVPR-2015} dataset and achieve competitive performance on the Middlebury 
dataset~\cite{Midd1-CVPR-2007,Midd2-CVPR-2007}.
}
\item We show that the model trained on a certain type of data (e.g., RGB/depth) generalizes well to handle image data in a variety of domains.
\end{itemize}

A preliminary version of this work was presented earlier in~\cite{DJF-ECCV-2016}.
In this paper, we significantly extend our work and summarize the main differences as follows.
First, we propose an improved network architecture for joint image filtering.
Instead of directly predicting filtered pixel values (as in~\cite{DJF-ECCV-2016}), we
predict a residual image by adding a skip connection from the input target image to the output (Figure~\ref{fig:framework}).
As the residual learning alleviates the need for restoring specific target image contents (which complicates the learning process), we show significant improvement in transferring accurate details from the guidance to the target image.
Second, in \cite{DJF-ECCV-2016}, we train the model only using an RGB/depth dataset and then evaluate its generalization ability on other domains.
In this work, we show that the model trained using an RGB/flow dataset also generalizes well on other visual domains.
This demonstrates that our network design is insensitive to the modality of the training data.
Third, we evaluate our approach on various joint image filter applications, compare against several state-of-the-art joint image filters (including concurrent work~\cite{Tai-2016-depth,Barron-2016-solver}), and conduct a detailed ablation study by analyzing the
performance of all methods under different hyper-parameter settings (e.g., filter number, filter size, network depth).

%%--------------------------------------------------------------------------------------------------------------------------------
%%--------------------------------------------------------------------------------------------------------------------------------
%%---------------Related work--------------------------------------------------------------------------------------------------
%%--------------------------------------------------------------------------------------------------------------------------------
%%--------------------------------------------------------------------------------------------------------------------------------

\section{Related Work}

{\flushleft \textbf{Joint image filters.}} Joint image filters can be categorized into two main classes based on explicit filter construction or global optimization of data fidelity and regularization terms.

Explicit joint filters compute the filtered output as a weighted average of neighboring pixels in the target image.
The bilateral filters~\cite{BF-ICCV-1998,Yang-CVPR-2007,JBU-TOG-2007,JGU-CVPR-2013,Rolling-ECCV-2014,Barron-2016-solver} and guided filters~\cite{He-PAMI-2013} are
representative algorithms in this class.
The filter weights, however, depend solely on the local structure of the guidance image.
Therefore, erroneous or extraneous structures may be transferred to the target image
due to the lack of consistency constraints.
In contrast, our model considers the contents of both images
based on feature maps and enforces consistency implicitly through learning from examples.

Numerous approaches formulate joint filtering based on a global optimization framework.
The objective function typically consists of two terms: data fidelity and regularization terms.
The data fidelity term ensures that the filtering output is close to the input target image.
These techniques differ from each other mainly in the regularization term that encourages
the output to have a similar structure with the guidance image.
The regularization term can be defined according to texture derivatives~\cite{MRF-NIPS-2005}, mid-level representations~\cite{Park-ICCV-2011} such as segmentation and saliency,
filtering outputs~\cite{Ham-CVPR-2015}, or mutual structures shared by the target and guidance image~\cite{Shen-ICCV-2015}.
However, global optimization based methods rely on hand-designed objective functions that may not reflect the complexities of natural images.
Furthermore, these approaches involve iterative optimization are often time-consuming.
In contrast, our method learns how to selectively transfer important details directly from the RGB/depth data.
Although the training process is time-consuming, the learned model is efficient during run-time.

{\flushleft \textbf{Learning-based image filters.}}
With significant success in high-level vision tasks~\cite{Imagenet-NIPS-2012}, substantial efforts have been made to construct image filters using learning algorithms and CNNs.
For example, the conventional bilateral filter can be improved by replacing the predefined filter weights with those learned from a large amount of data~\cite{Jam-2016-HDF,gharbi2017deep,chen-ICCV-2017}.
In the context of joint depth upsampling, Tai et al.~\cite{Tai-2016-depth} use a multi-scale guidance strategy to improve upsampling performance.
Gu et al.~\cite{gu2017learning} adjust the original guidance dynamically to account for the iterative updates of the filtering results.
However, these methods~\cite{Tai-2016-depth,gu2017learning} are limited to the application of depth map upsampling.
In contrast, our goal is to construct a generic joint filter for various applications using
target/guidance image pairs in different visual domains.

%%\vspace{1em}
{\flushleft \textbf{Deep models for low-level vision.}} In addition to filtering, deep learning models have also been applied to other low-level vision and computational photography tasks.
Examples include image denoising~\cite{Denoise-CVPR-2012}, raindrop removal~\cite{Rain-ICCV-2013}, image super-resolution~\cite{SRCNN-ECCV-2014}, image deblurring~\cite{zhang-2016-learning} and optical flow estimation~\cite{Flownet-ICCV-2015}.
Existing deep learning models for low-level vision use either one input image~\cite{SRCNN-ECCV-2014,Denoise-CVPR-2012,Rain-ICCV-2013,Xu-ICML-2015} or two images in the same domain \cite{Flownet-ICCV-2015}.
In contrast, our network can accommodate two streams of inputs by \emph{heterogeneous} domains, e.g., RGB/NIR, flash/non-flash, RGD/Depth, intensity/color.
Our network architecture bears some resemblance to that in Dosovitskiy et al.~\cite{Flownet-ICCV-2015}.
The main difference is that the merging layer used in~\cite{Flownet-ICCV-2015} is a correlation operator while our model integrates the inputs through concatenating the feature responses.
Furthermore, we adopt the residual learning by introducing the skip connection.

Another closely related work is by Xu et al.~\cite{Xu-ICML-2015}, which learns a CNN to approximate existing edge-aware filters from example images.
Our method differs from \cite{Xu-ICML-2015} in two aspects.
First, the goal of~\cite{Xu-ICML-2015} is to use CNN for approximating existing edge-aware filters. In contrast, our goal is to learn a new joint image filter.
Second, unlike the network in~\cite{Xu-ICML-2015} that takes only one single RGB image, the proposed joint filter handles two images from different domains and modalities.

{\flushleft \textbf{Skip connections.}}
% JB: discussing purposes of skip connections.
% Learning deep network
% - ResNet, DenseNet
% Learning only high-frequency
% - VDSR, LapSR
As deeper networks have been developed for vision tasks,
the information contained in the input or gradients can vanish and wash out by the time
it reaches the end (or beginning) of the network.
He et al.~\cite{resnet-2016-deep} address this problem through bypassing the signals
from one layer to the next via skip connections.
This residual learning method facilitates us to train very deep networks effectively.
The work of~\cite{huang2017densely} further strengthens its effectiveness with dense connections across all the layers.
For low-level vision tasks, skip connection have been shown to be useful to restore high-frequency details~\cite{VDSR-2016-accurate,lai2017deep} by enforcing the network
to learn the residual signals only.

%%--------------------------------------------------------------------------------------------------------------------------------
%%--------------------------------------------------------------------------------------------------------------------------------
%%--------------Learning Deep Joint Image Filters--------------------------------------------------------------------------
%%--------------------------------------------------------------------------------------------------------------------------------
%%--------------------------------------------------------------------------------------------------------------------------------

\section{Learning Joint Image Filters}

In this section, we introduce a learning-based joint image filter based on CNNs.
We first present the network design (Section~\ref{sec:design}) and skip connection (Section~\ref{sec:skip}).
Next, we describe the network training process (Section~\ref{sec:training}) and visualize the guidance map generated by the network (Section~\ref{sec:vis}).

Our CNN model consists of three sub-networks:
the target network $\mathrm{CNN_T}$, the guidance network $\mathrm{CNN_G}$, and the filter network $\mathrm{CNN_F}$ as shown in Figure~\ref{fig:framework}.
First, the sub-network $\mathrm{CNN_T}$ takes the target image as input and extracts a feature map.
Second, similar to $\mathrm{CNN_T}$, the sub-network $\mathrm{CNN_G}$ extracts a feature map from the guidance image.
Third, the sub-network $\mathrm{CNN_F}$ takes the concatenated feature responses from the sub-networks $\mathrm{CNN_T}$ and $\mathrm{CNN_G}$ as input and generates the residual, i.e., the difference between the degraded target image and ground truth.
By adding the target input through the skip connection, we obtain the final joint filtering result.
Here, the main roles of the two sub-networks $\mathrm{CNN_T}$ and $\mathrm{CNN_G}$ are to serve as non-linear feature extractors that capture the local structural details in the respective target and guidance images.
The sub-network $\mathrm{CNN_F}$ can be viewed as a non-linear regression function that maps the feature responses from both target and guidance images to the desired residuals.
Note that the information from target and guidance images is simultaneously considered when predicting the final filtered result.
Such a design allows us to selectively transfer structures and avoid texture-copying artifacts.

\begin{figure}[t]
\centering

\begin{tabular}{c@{\hspace{0.005\linewidth}}c@{\hspace{0.005\linewidth}}c@{\hspace{0.005\linewidth}}c@{\hspace{0.005\linewidth}}c@{\hspace{0.005\linewidth}}c@{\hspace{0.005\linewidth}}c}
\includegraphics[width = .48\linewidth]{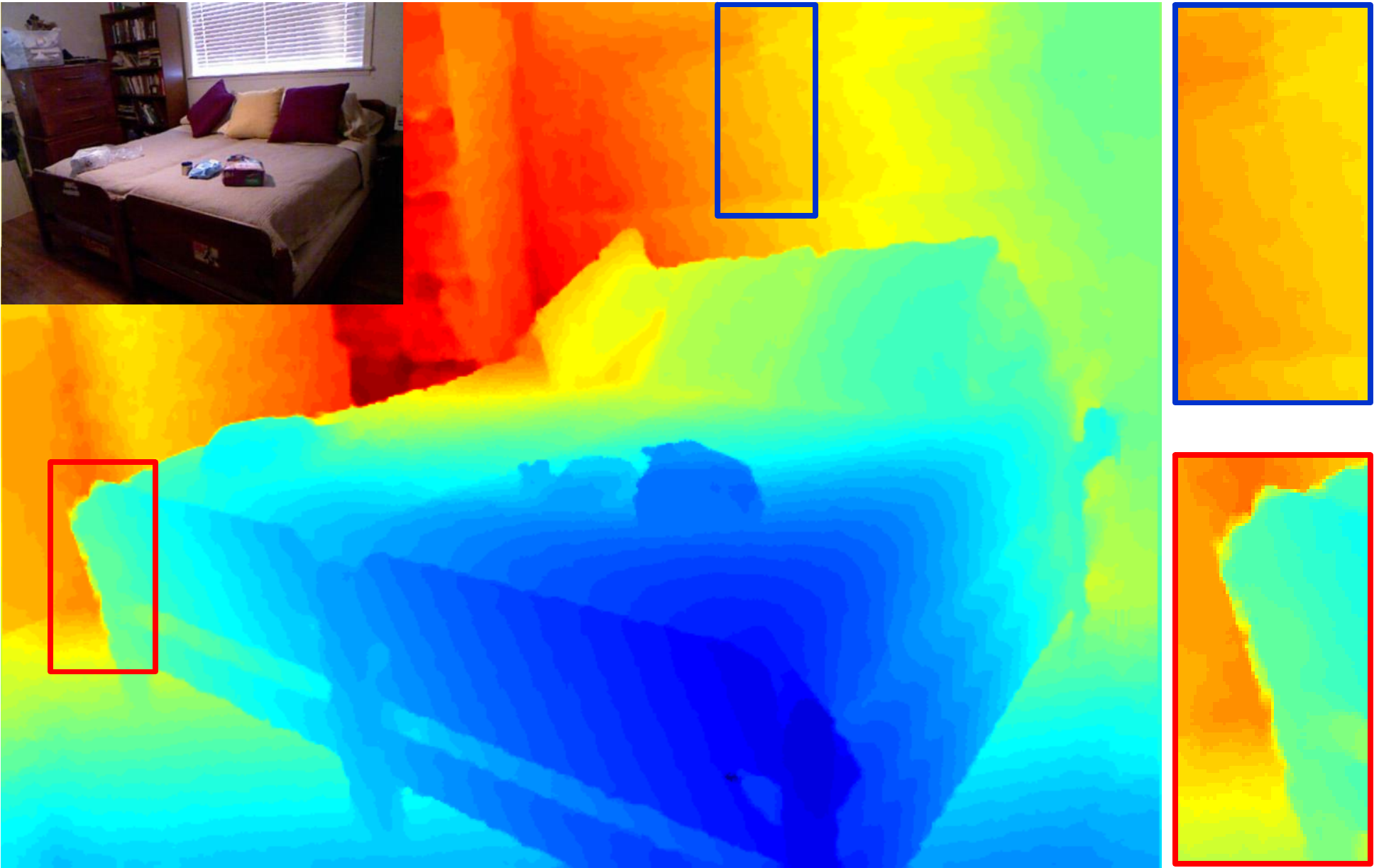} &
\includegraphics[width = .48\linewidth]{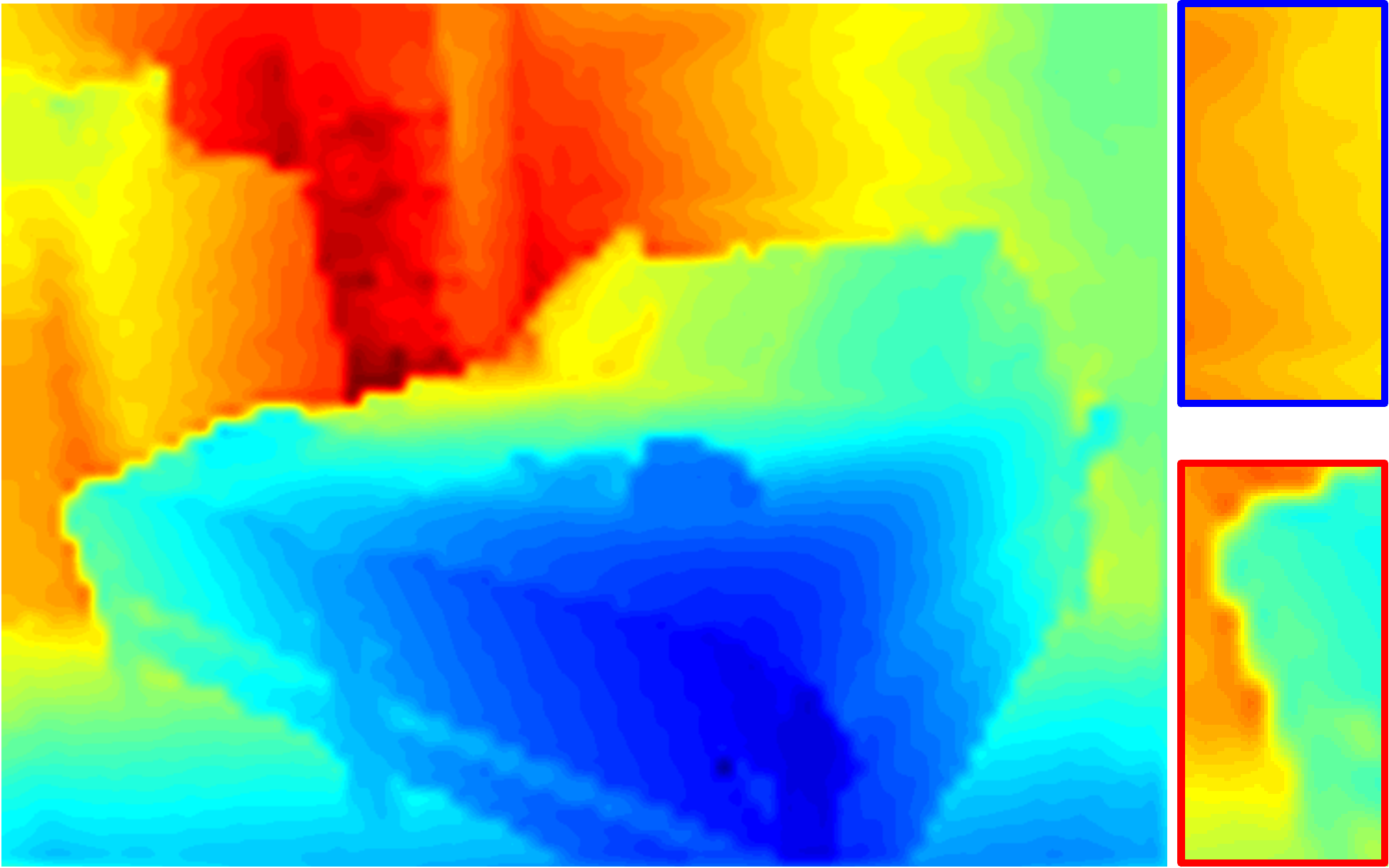} & \\

{ (a) Ground truth}& {(b) Bicubic upsampling, 5.82} & \\

\includegraphics[width = .48\linewidth]{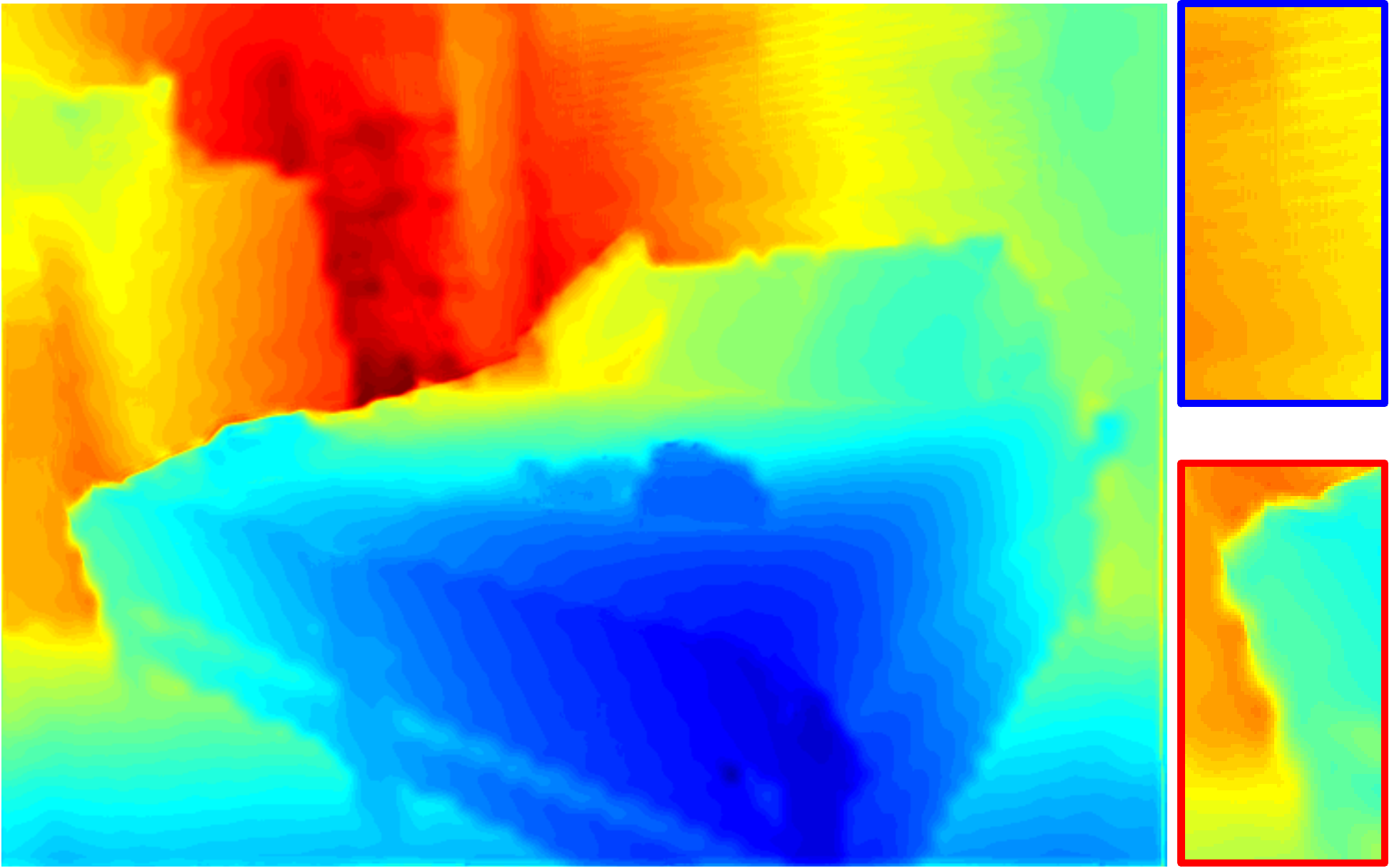} &
\includegraphics[width = .48\linewidth]{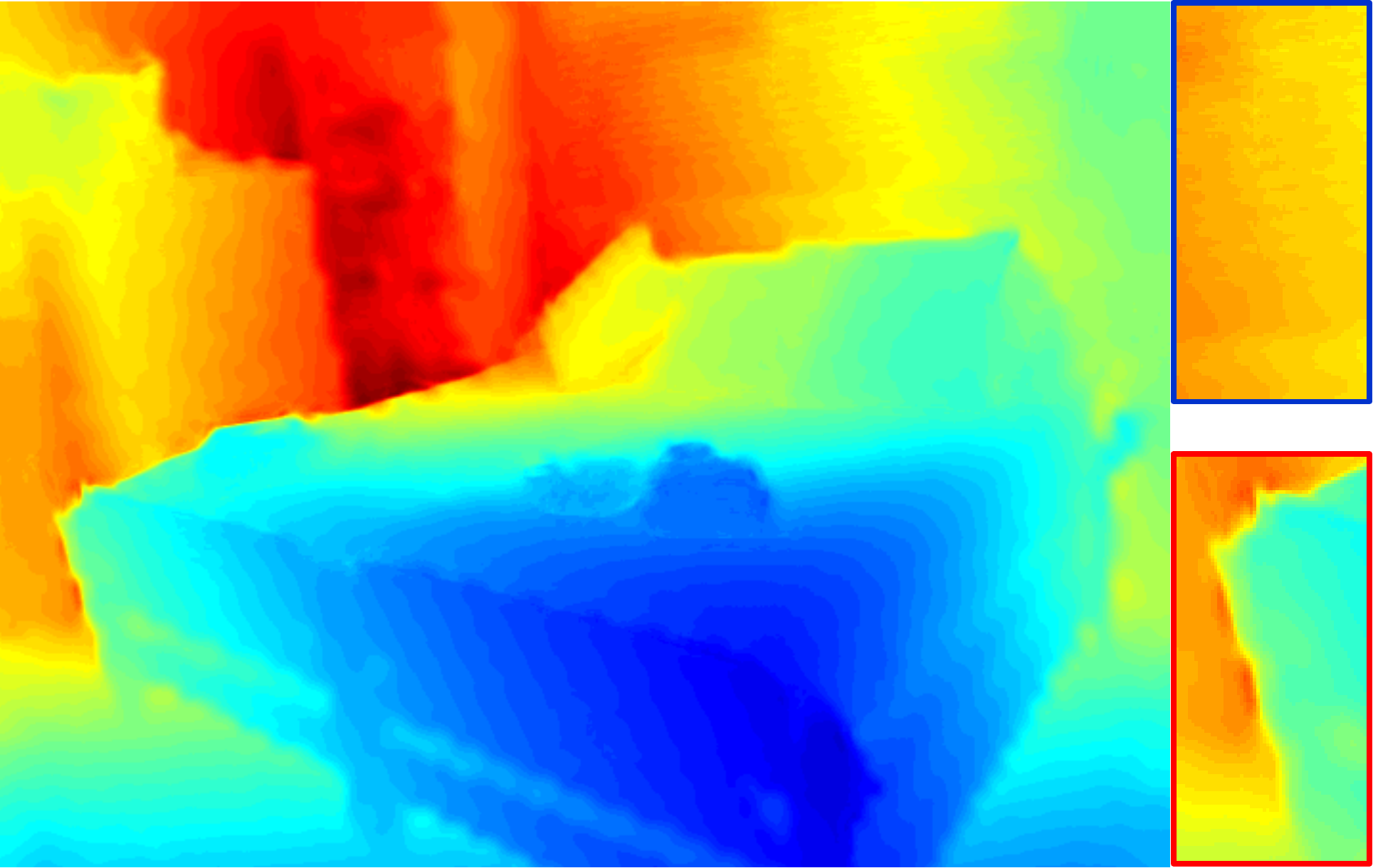} & \\

{ (c) 3-layer $\mathrm{CNN_F}$, 4.05} & {(d) 3-layer $\mathrm{CNN_F\_R}$, 3.96} \\

\includegraphics[width = .48\linewidth]{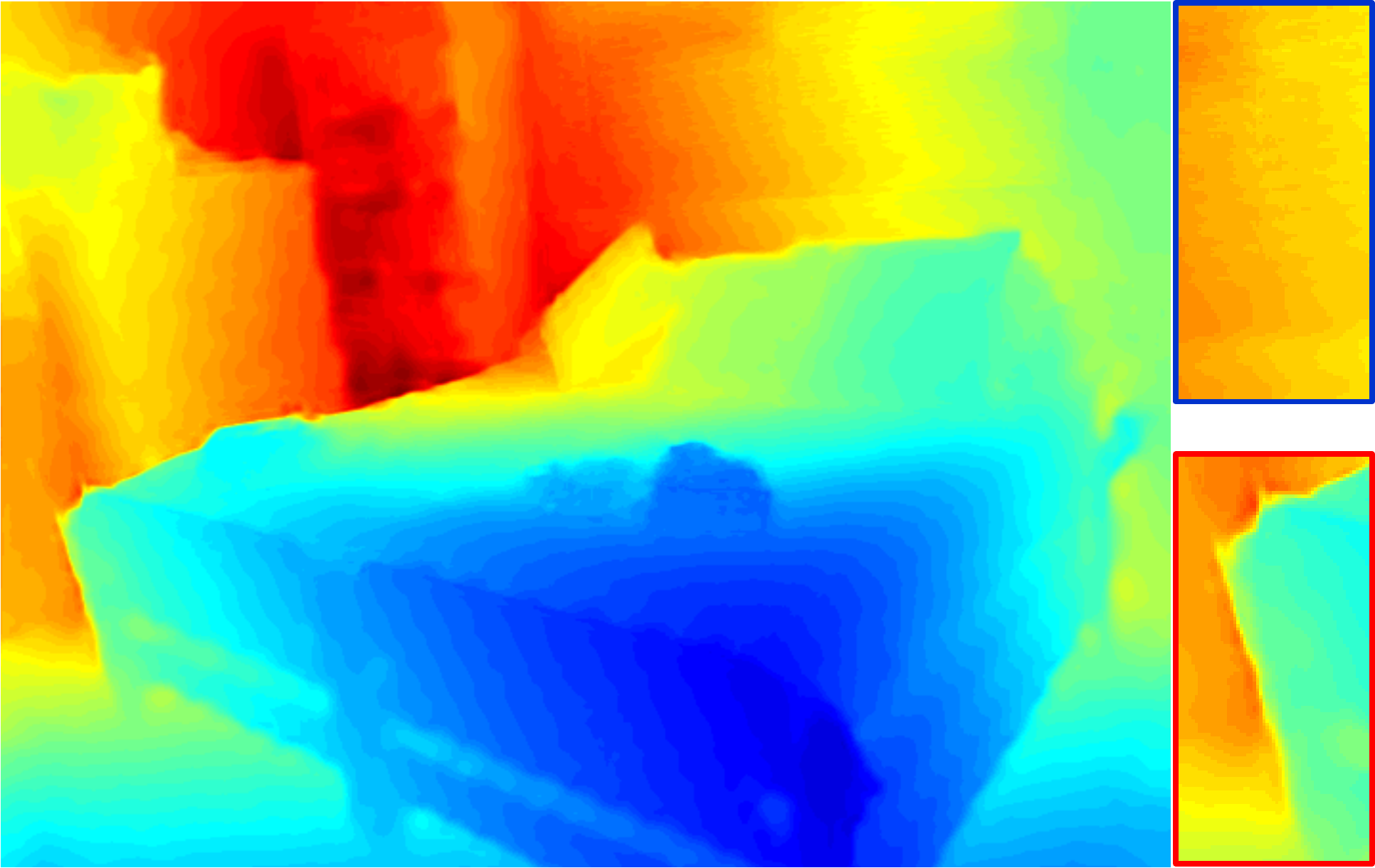} &
\includegraphics[width = .48\linewidth]{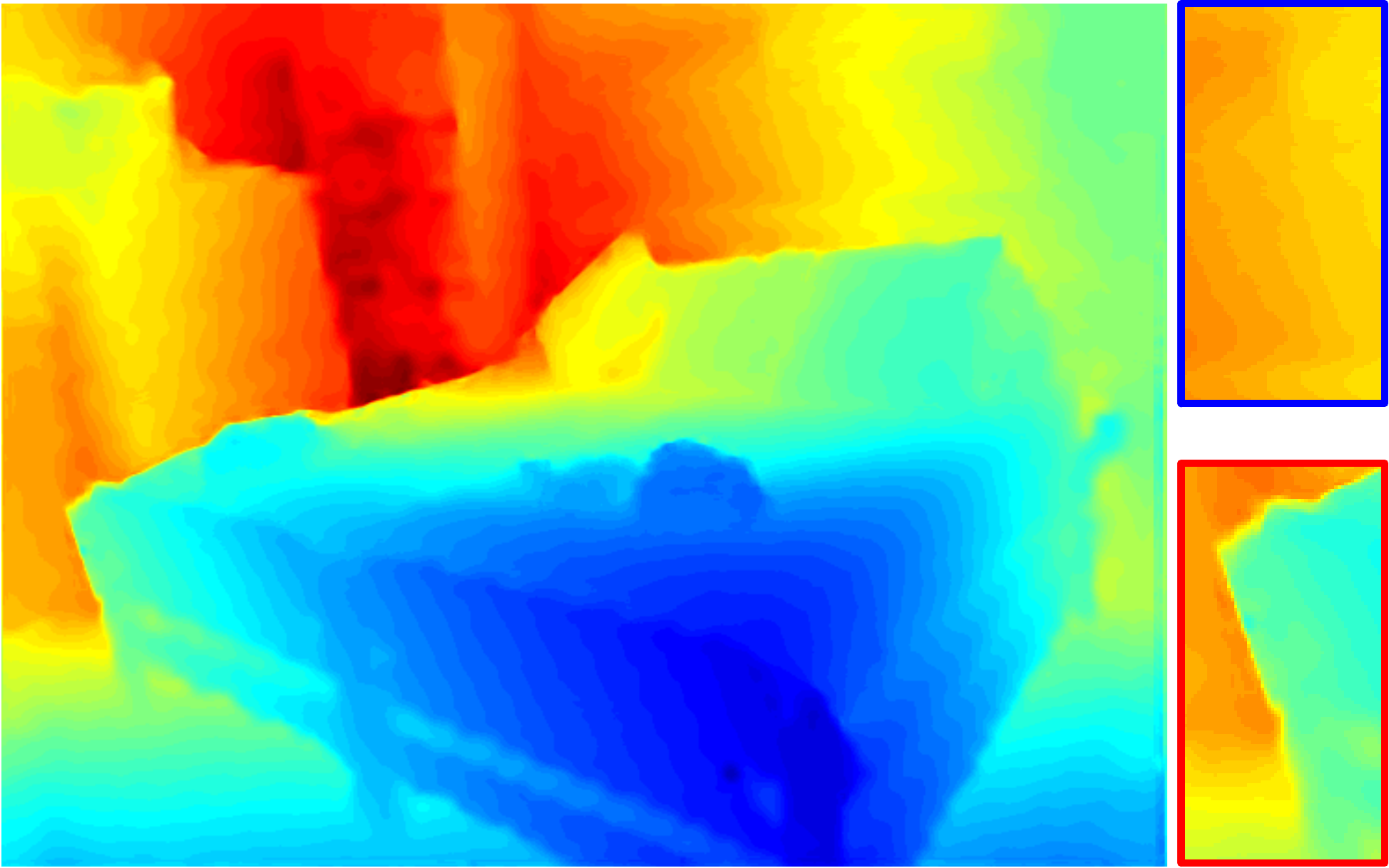} & \\

{(e) 4-layer $\mathrm{CNN_F\_R}$, 3.91} & {(f) Our network, 2.62}\\

\end{tabular}

\caption{
\textbf{Comparison of network design.}
Joint depth upsampling (8$\times$) results of using different network architectures. % $f_1$-$f_2$-... where $f_i$ is the filter size of the $i$-th layer.
(a) GT depth map (inset: guidance image).
(b) Bicubic upsampling.
\revyj{(c)-(e) Results from the straightforward implementation using $\mathrm{CNN_F}$ and $\mathrm{CNN_F\_R}$.
(f) Our results. Note the difference on the bed corner and curtain. The numbers are the RMSE metric based on the GT in (a).}
}
\label{fig:problem}
\end{figure}

\subsection{Network architecture design} \label{sec:design}
To design a joint filter using CNNs, a straightforward implementation is to concatenate the target and guidance images together and directly train a generic CNN similar to the filter network $\mathrm{CNN_F}$.
While in theory, we can train a generic CNN to approximate the desired function for joint filtering, our empirical results show that such a network generates poor results.
Figure~\ref{fig:problem}(c)-(d) shows an example of joint depth upsampling using the network $\mathrm{CNN_F}$ \revyj{and its residual-based variant $\mathrm{CNN_F\_R}$}.
The main structures (e.g., the bed corner) contained in the guidance image are not well transferred to the target depth image, thereby resulting in blurry boundaries.
In addition, inconsistent texture structures in the guidance image (e.g., the stripe pattern of the curtain on the wall) are also incorrectly copied to the target image.
A potential approach that may improve the results is to adjust the architecture of $\mathrm{CNN_F}$, such as increasing the network depth or using larger filter sizes.
However, as shown in Figure~\ref{fig:problem}(e), these variants do not show notable improvement.
Blurry boundaries and the texture-copying problem still occur.
We note that similar observations have also been reported in~\cite{SRCNN-PAMI-2015}, which indicate that the effectiveness of deeper structures for low-level tasks is not as apparent as that shown in high-level tasks (e.g., image classification).

\begin{figure}[t]
\centering

\begin{tabular}{c@{\hspace{0.005\linewidth}}c@{\hspace{0.005\linewidth}}c@{\hspace{0.005\linewidth}}c@{\hspace{0.005\linewidth}}c@{\hspace{0.005\linewidth}}c@{\hspace{0.005\linewidth}}c}
\includegraphics[width = .315\linewidth]{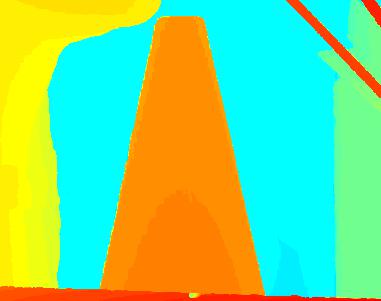} &
\includegraphics[width = .315\linewidth]{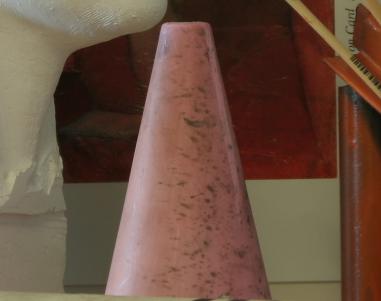} &
\includegraphics[width = .315\linewidth]{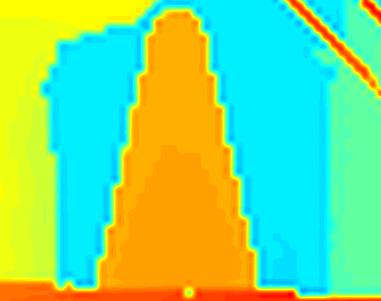} &\\

{(a) GT depth} & {(b) Guidance} & {(c) Bicubic}\\

\includegraphics[width = .315\linewidth]{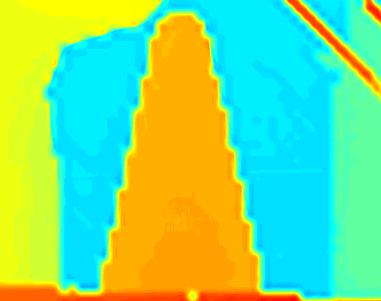} &
\includegraphics[width = .315\linewidth]{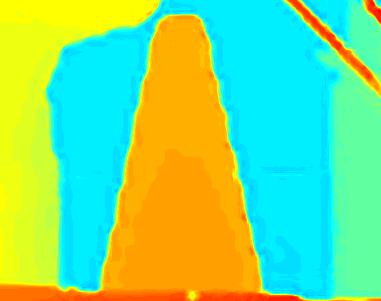} &
\includegraphics[width = .315\linewidth]{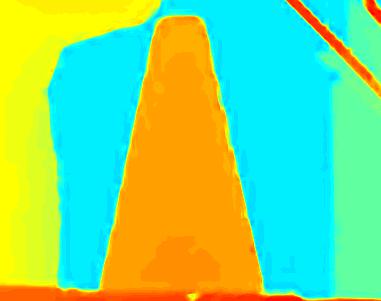} & \\

{(d) RGB guided} & {(e) Edge guided} & {(f) Ours}\\

\end{tabular}

\caption{
\textbf{Comparison of different types of guidance.}
Joint depth upsampling (8$\times$) results using different types of guidance images.
Both (d) and (e) are trained using the $\mathrm{CNN_F}$ network.
Our method generates sharper boundary of the sculpture (left) and the cone (middle).
}
\label{fig:edgemap}
\end{figure}

We attribute the limitation of using a generic network for joint filtering to the fact that the original RGB guidance image fails to provide direct and effective guidance as it mixes a variety of information (e.g., texture, intensity, and edges).
To validate this intuition, we show in Figure~\ref{fig:edgemap} one example where we replace the original RGB guidance image with its edge map extracted using~\cite{Dollar-ICCV-2013}.
Compared to the results guided by the RGB image (Figure~\ref{fig:edgemap}(d)), the
upsampled image using the edge map guidance (Figure~\ref{fig:edgemap}(e))
shows substantial improvement in preserving the sharp edges.

Based on the above observation, we introduce two sub-networks $\mathrm{CNN_T}$ and $\mathrm{CNN_G}$ to first construct two separate processing streams for the two images
before concatenation.
With the proposed architecture, we constrain the network to extract effective features from both images separately first and then fuse them at a later stage to generate the final filtering output.
This differs from conventional joint filters where the guidance information is mainly computed from the pixel-level intensity/color differences in the local neighborhoods.
As our models are jointly trained in an end-to-end fashion, our result (Figure~\ref{fig:edgemap}(f)) shows further improvements over that of using the edge guided filtering (Figure~\ref{fig:edgemap}(e)).

In this work, we adopt a three-layer structure for each sub-network as shown in Figure~\ref{fig:framework}.
Given $M$ training image samples $\{I_{i}^{T}, I_{i}^{G}, I_{i}^{gt}\}_{i=1}^{M}$, we learn the network parameters by minimizing the sum of the squared losses:
\begin{equation}
\label{formula1}
\| I_{}^{gt}-\Phi(I_{}^{T}, ~I_{}^{G}) \|_2^2~,
\end{equation}
where $\Phi$ denotes the joint image filtering operator.
In addition, $I_{}^{T}$, $I_{}^{G}$, and $I_{}^{gt}$
denote the target image, the guidance image and the ground truth output, respectively.

\begin{figure}[t]
\centering

\begin{tabular}{c@{\hspace{0.005\linewidth}}c@{\hspace{0.005\linewidth}}c@{\hspace{0.005\linewidth}}c@{\hspace{0.005\linewidth}}c@{\hspace{0.005\linewidth}}c@{\hspace{0.005\linewidth}}c}

\includegraphics[width = .48\linewidth]{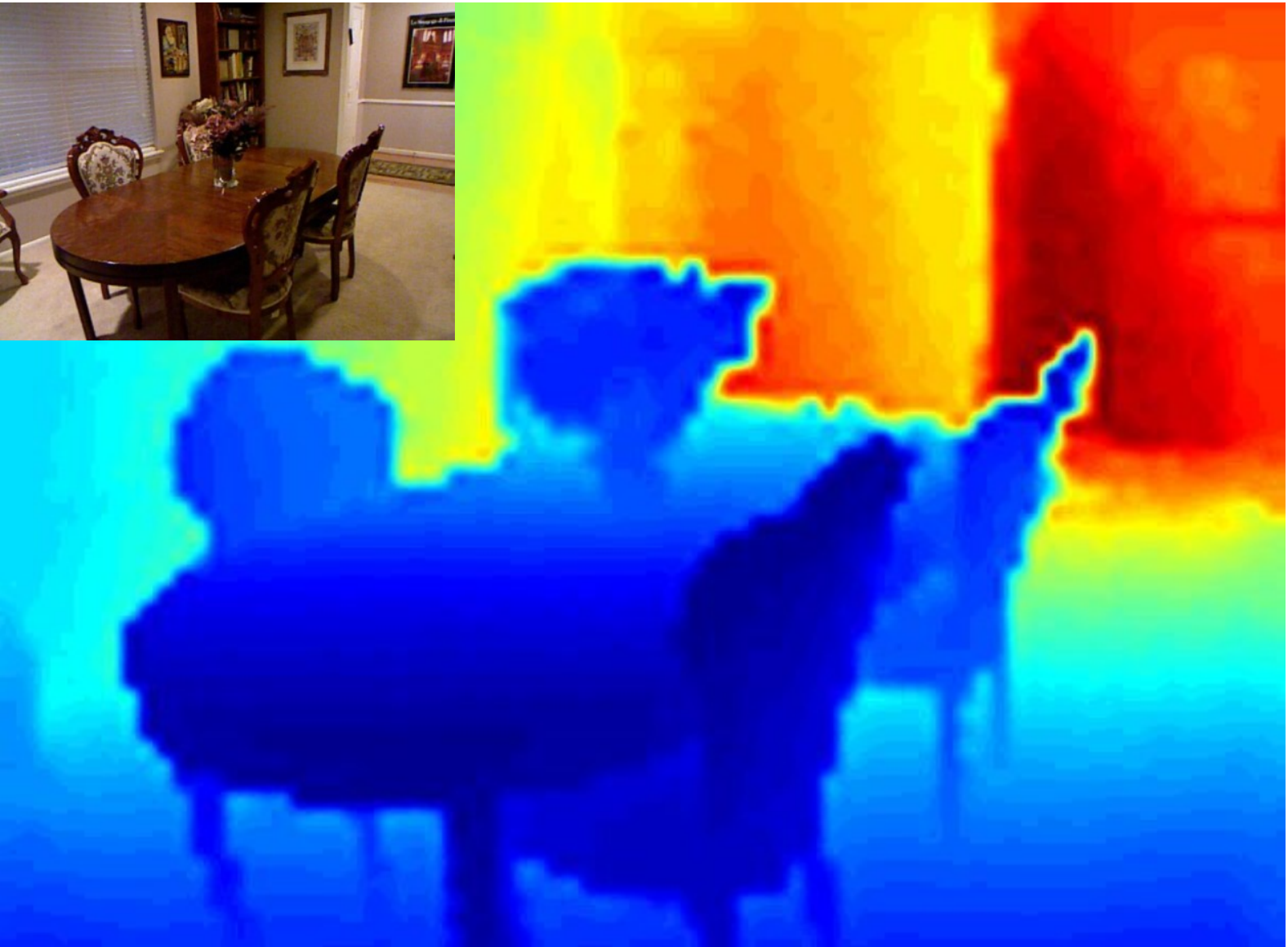} &
\includegraphics[width = .48\linewidth]{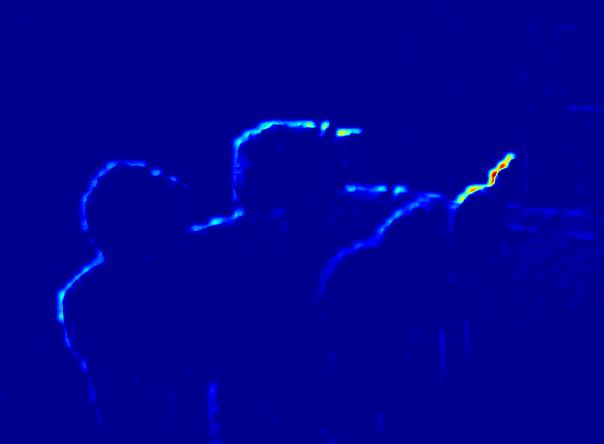} & \\

{(a) Target input} & {(b) Residual output} \\

\includegraphics[width = .48\linewidth]{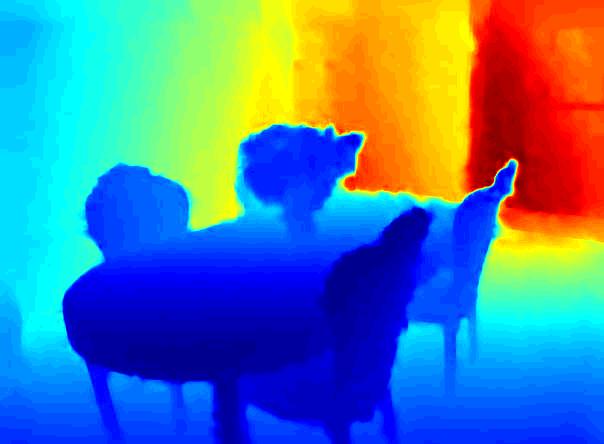} &
\includegraphics[width = .48\linewidth]{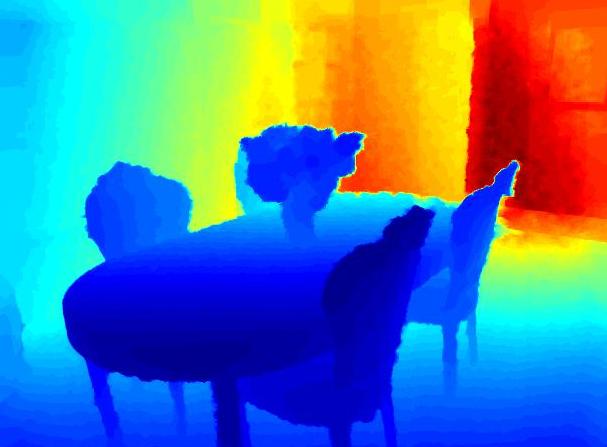} &\\

{(c) Filtering output} & {(d) Ground truth} \\

\end{tabular}

\caption{
\textbf{Residual prediction.}
Joint depth upsampling results (8$\times$) of using our network with a skip connection.
The filtering output (c) is the summation of (a) the target input and (b) the predicted output.}
\label{fig:res1}
\end{figure}

\begin{figure*}[t]
\centering

\begin{tabular}{c@{\hspace{0.005\linewidth}}c@{\hspace{0.005\linewidth}}c@{\hspace{0.005\linewidth}}c@{\hspace{0.005\linewidth}}c@{\hspace{0.005\linewidth}}c@{\hspace{0.005\linewidth}}c}

\includegraphics[width = .235\linewidth]{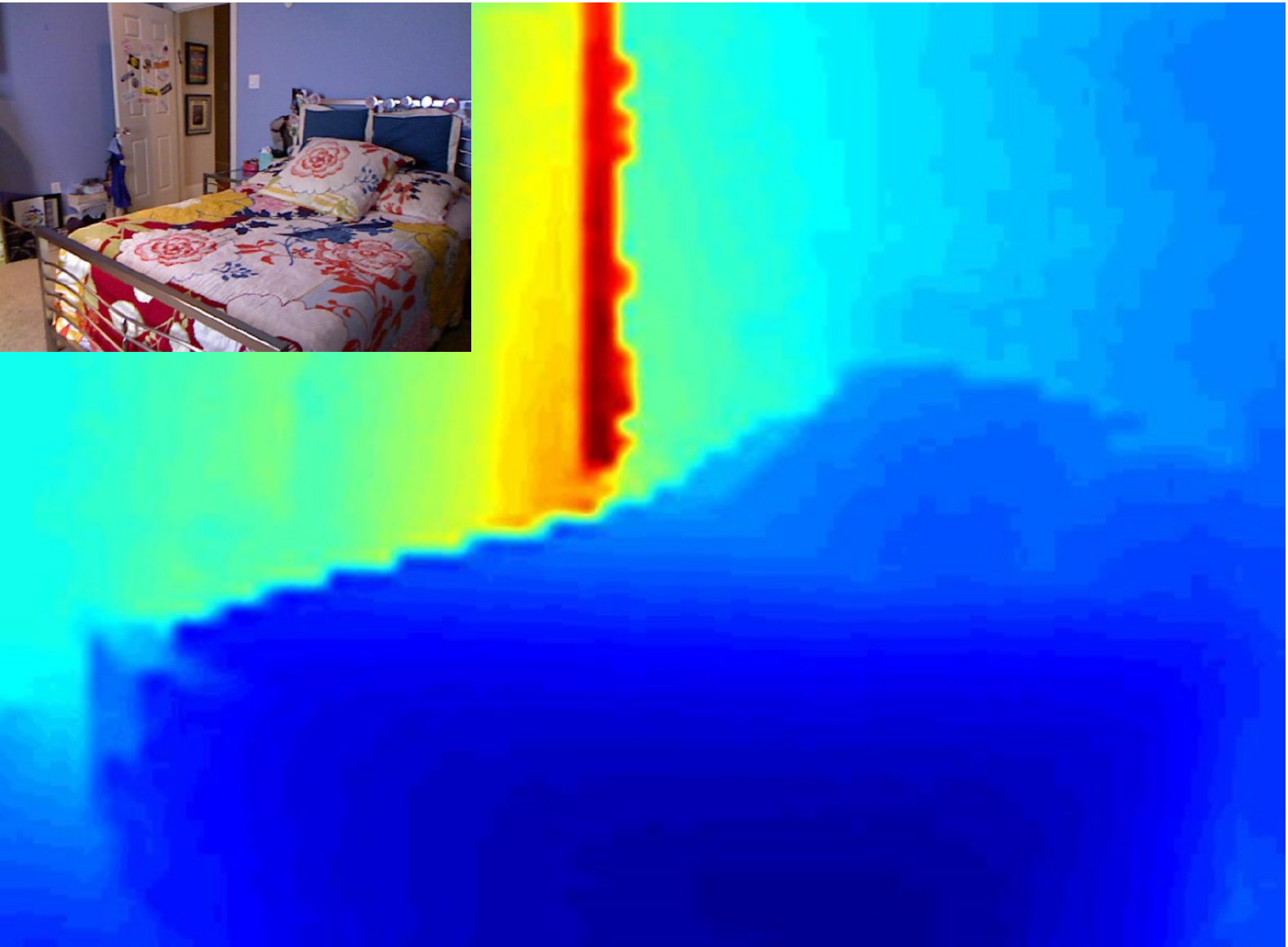} &
\includegraphics[width = .235\linewidth]{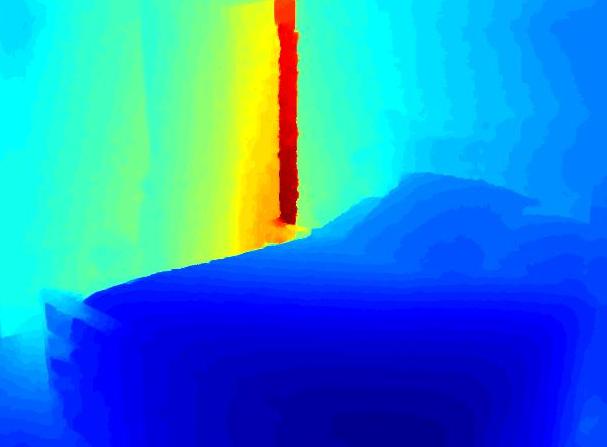} &

\hspace{1pt}\vrule\hspace{1pt}

\includegraphics[width = .235\linewidth, height= .173\linewidth]{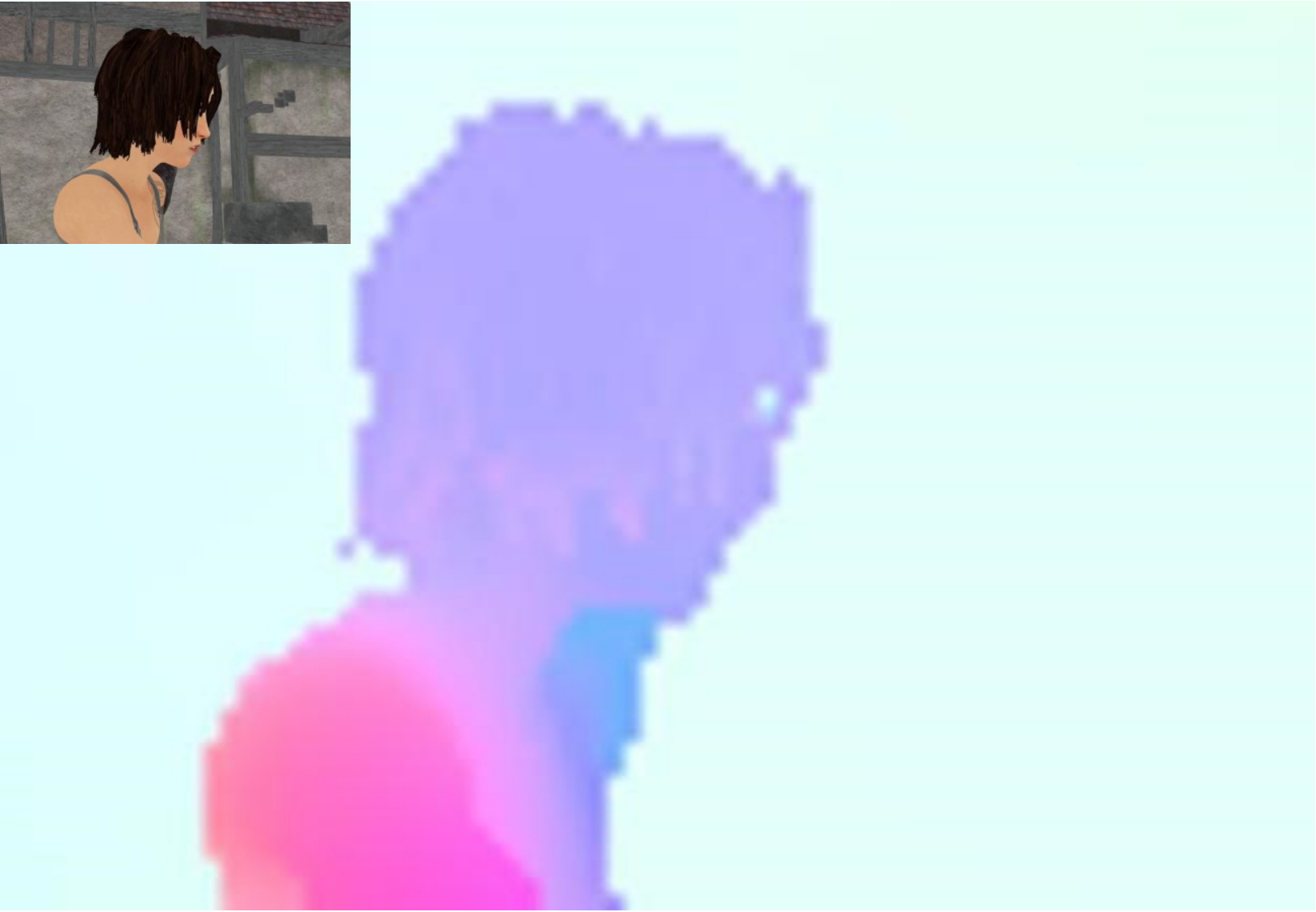} &
\includegraphics[width = .235\linewidth, height= .173\linewidth]{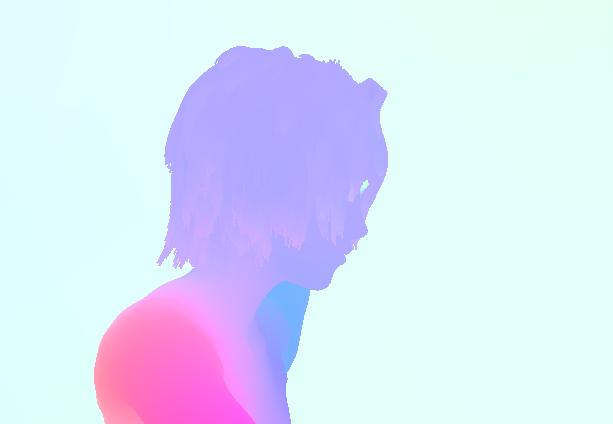} & \\

{(a) Target input depth} & {(b) GT depth} & {(e) Target input flow} & {(f) GT flow} \\

\includegraphics[width = .235\linewidth]{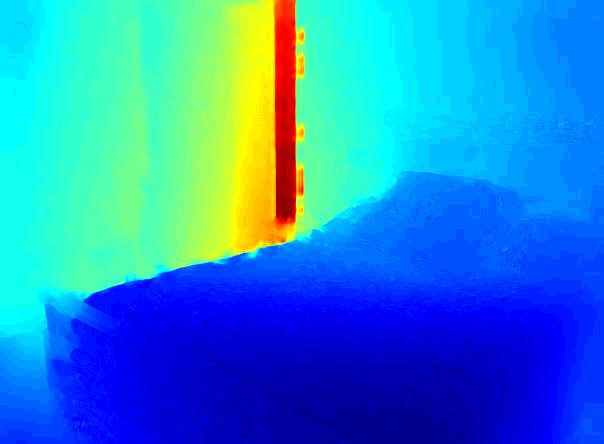} &
\includegraphics[width = .235\linewidth]{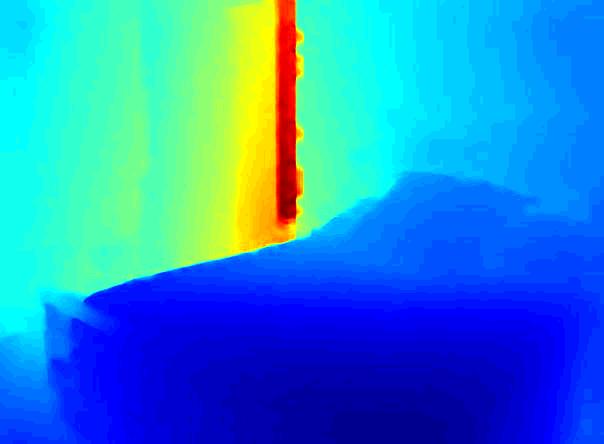} &

\hspace{1pt}\vrule\hspace{1pt}

\includegraphics[width = .235\linewidth, height= .173\linewidth]{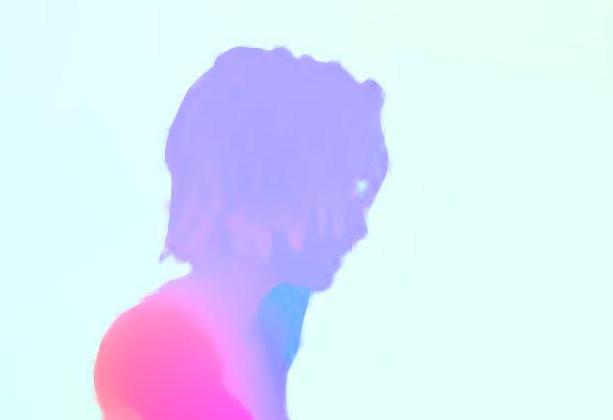} &
\includegraphics[width = .235\linewidth, height= .173\linewidth]{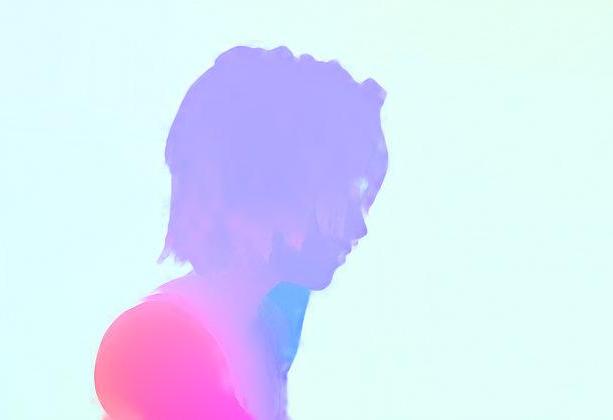} &\\

{(c) Upsample by the flow model} & {(d) Upsample by the depth model} & {(g) Upsample by the depth model} & {(h) Upsample by the flow model} \\

{3.54} & {3.36} & {5.63} & {5.35} \\

\end{tabular}
\caption{
\textbf{Effect of training data modalities.} (a)-(d) Joint depth map upsampling (8$\times$).
The model trained with RGB/flow data generates similar results when compared with the model trained with RGB/depth data. (e)-(h) Joint flow map upsampling (8$\times$). (g) The model trained with RGB/depth data and (h) The model trained with RGB/flow data. The numbers are the RMSE metric comparing against the GT.}
\label{fig:flow_on_depth}
\end{figure*}

\subsection{Skip connection} \label{sec:skip}

As the goal of the joint image filter is to leverage the signals from the guidance image to enhance the degraded target image, the input target image and the desired output share the same low-resolution frequency components.
We thus introduce a skip connection to enforce the network to focus on learning the residuals rather than predicting the actual pixel values.
With the skip connection, the network does not need to learn the identity mapping function from the input target image to the desired output in order to preserve the low-frequency contents.
Instead, the network learns to predict the sparse residuals in important regions (e.g., object contours).
In Figure~\ref{fig:res1}, we show an example of the predicted residuals, which highlights the estimated difference between the target input (Figure~\ref{fig:res1}(a)) and the ground truth (Figure~\ref{fig:res1}(d)).
Quantitative results in Table~\ref{table:depth} show that with the skip connection, the proposed algorithm obtains notable improvements over~\cite{DJF-ECCV-2016}.

\begin{figure*}[t]
\centering

\begin{tabular}{l@{\hspace{0.005\linewidth}}c@{\hspace{0.005\linewidth}}c@{\hspace{0.005\linewidth}}c@{\hspace{0.005\linewidth}}c@{\hspace{0.005\linewidth}}c@{\hspace{0.005\linewidth}}c@{\hspace{0.005\linewidth}}c@{\hspace{0.005\linewidth}}c@{\hspace{0.005\linewidth}}c@{\hspace{0.005\linewidth}}c@{\hspace{0.005\linewidth}}c@{\hspace{0.005\linewidth}}c}
\includegraphics[width = .19\linewidth, height = .145\linewidth]{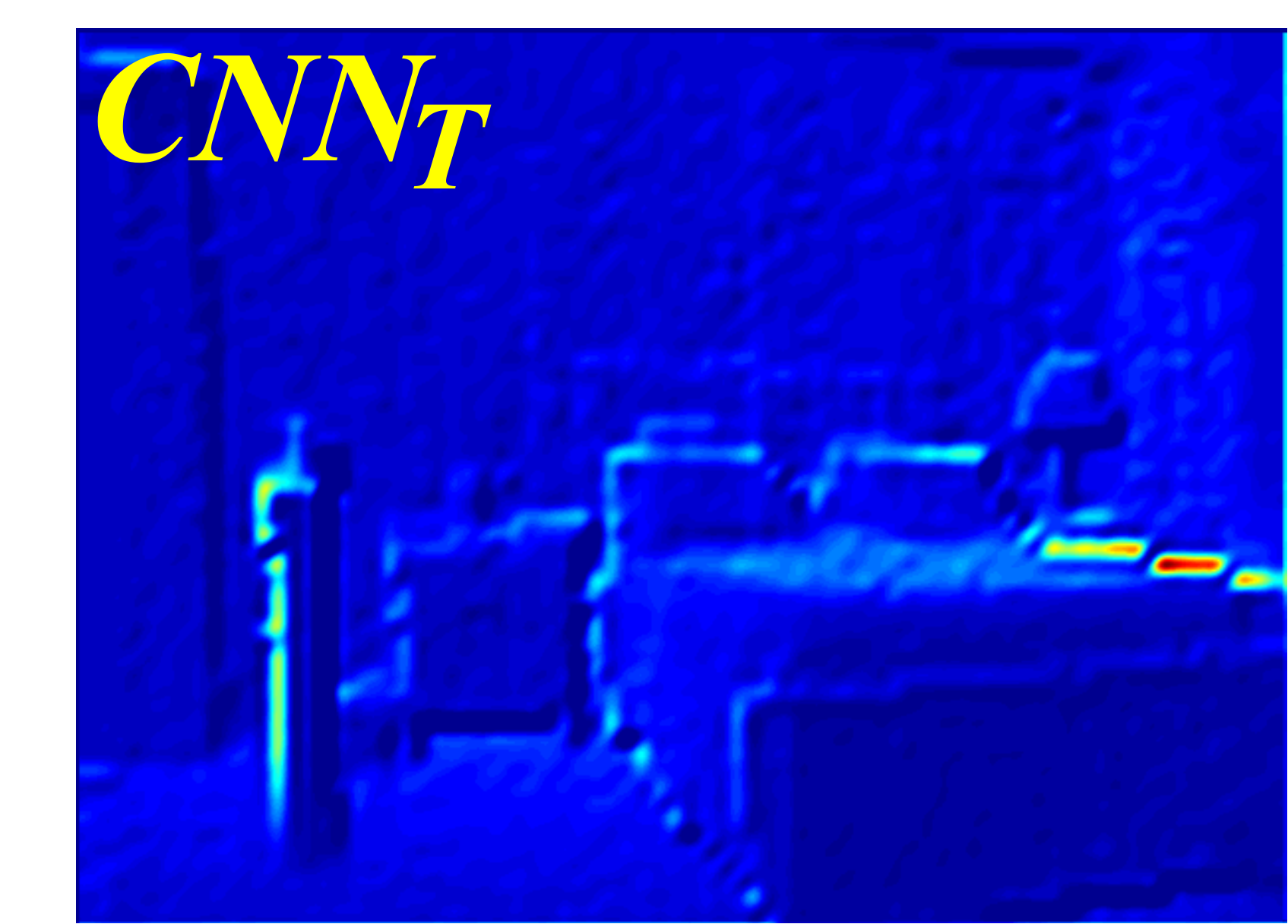} &
\includegraphics[width = .19\linewidth]{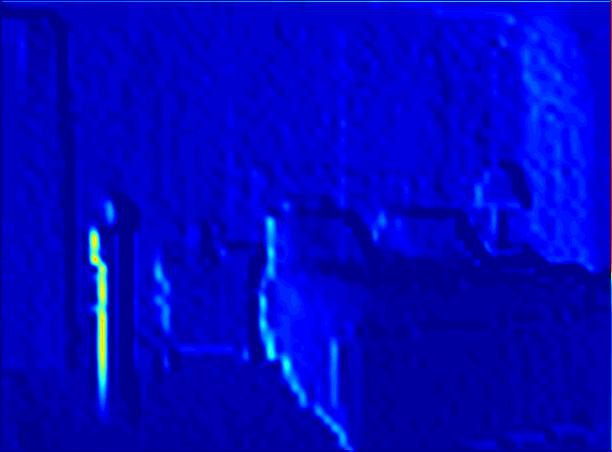} &
\includegraphics[width = .19\linewidth]{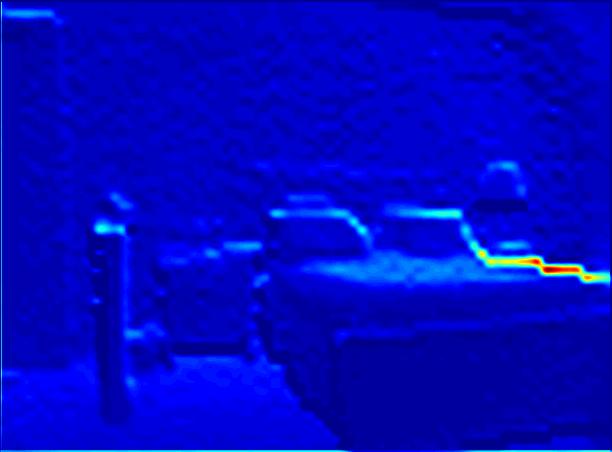} &
\includegraphics[width = .19\linewidth]{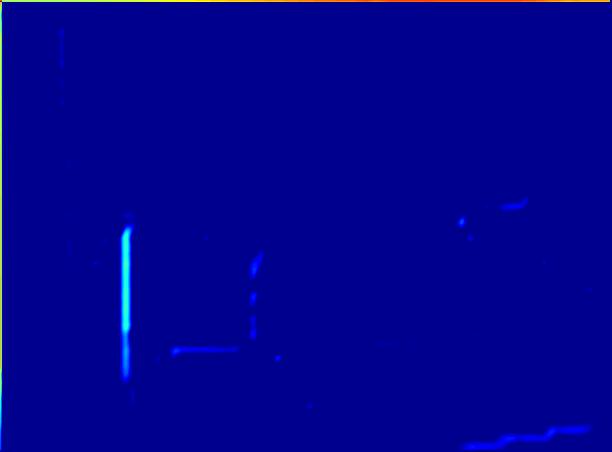} &
\includegraphics[width = .19\linewidth]{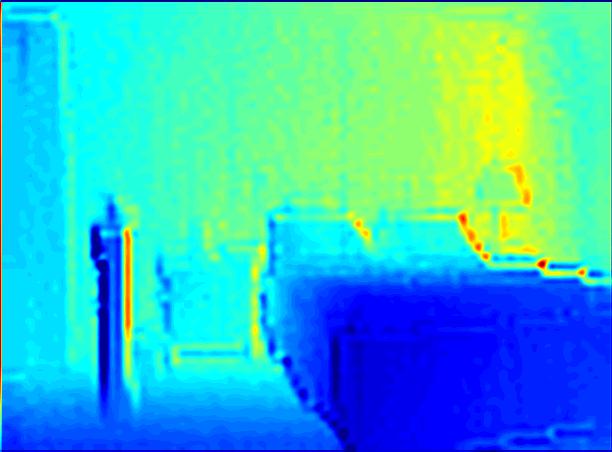} &
\includegraphics[height =.14\linewidth, width = .01\linewidth]{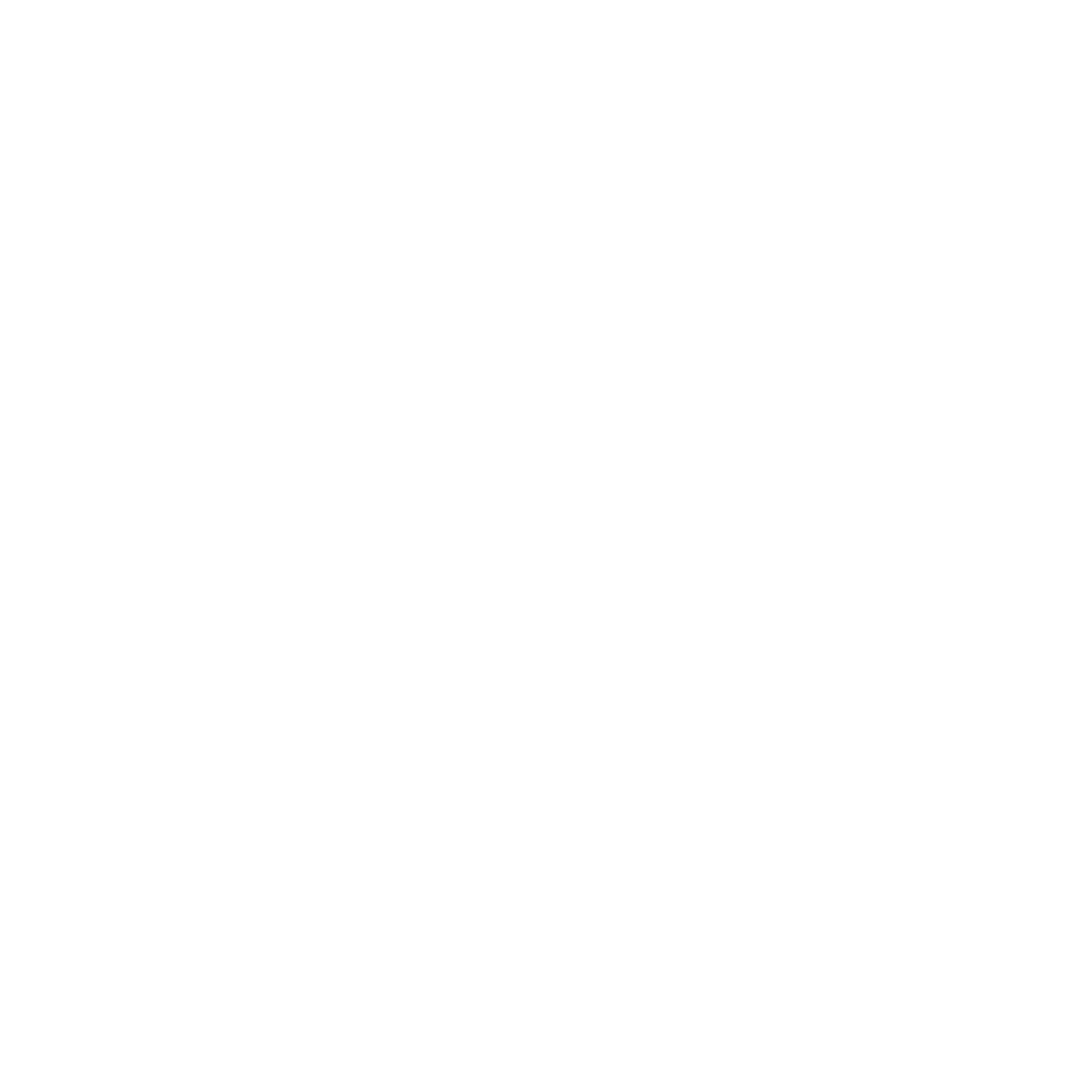} & \\

\includegraphics[width = .19\linewidth, height = .145\linewidth]{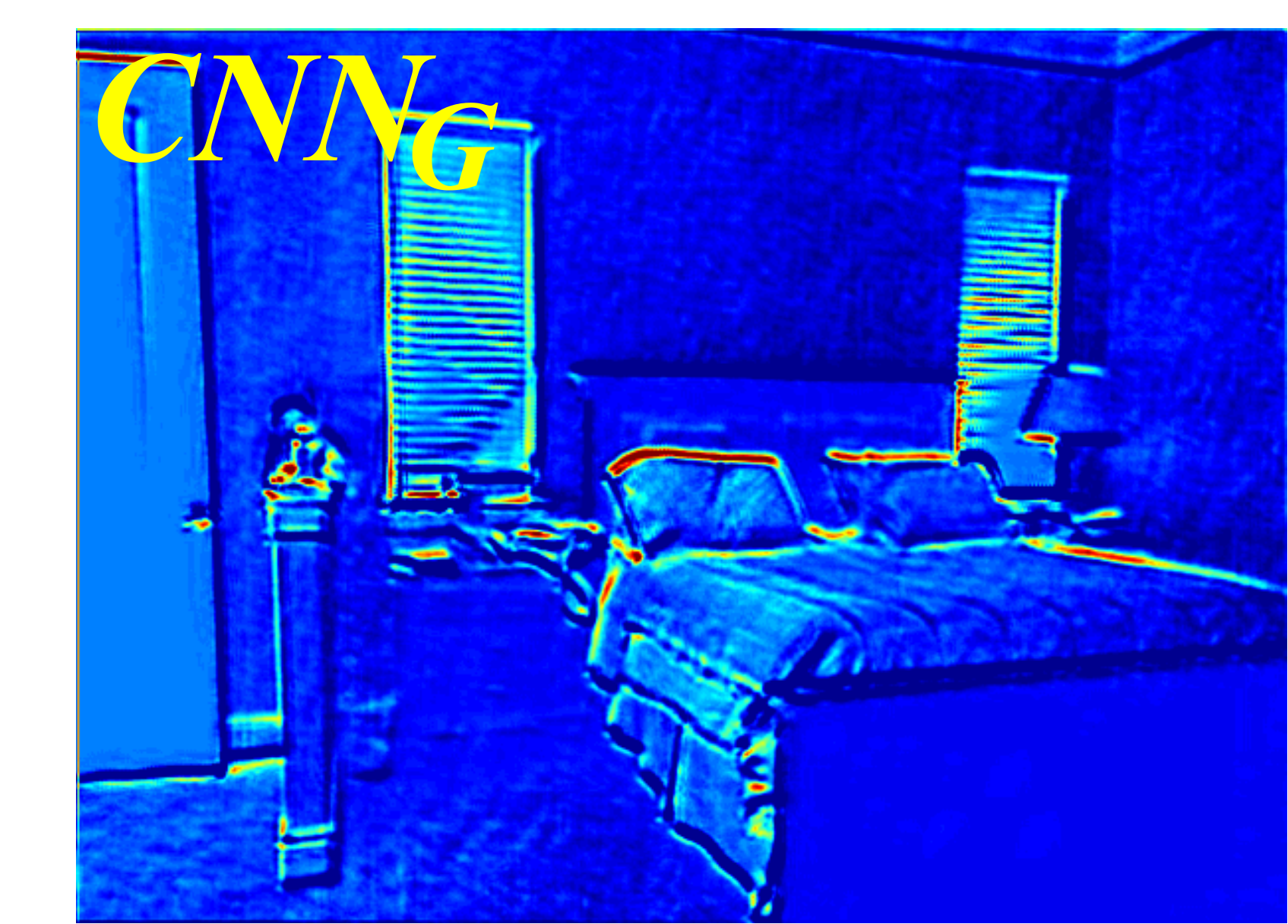} &
\includegraphics[width = .19\linewidth]{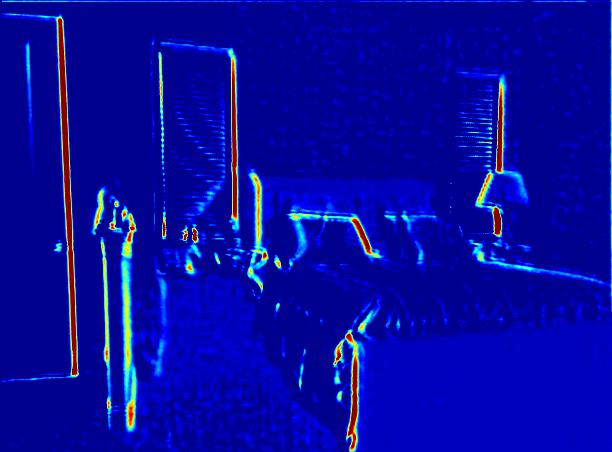} &
\includegraphics[width = .19\linewidth]{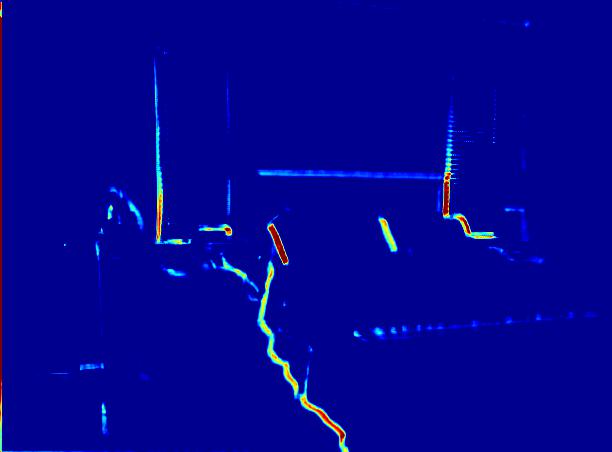} &
\includegraphics[width = .19\linewidth]{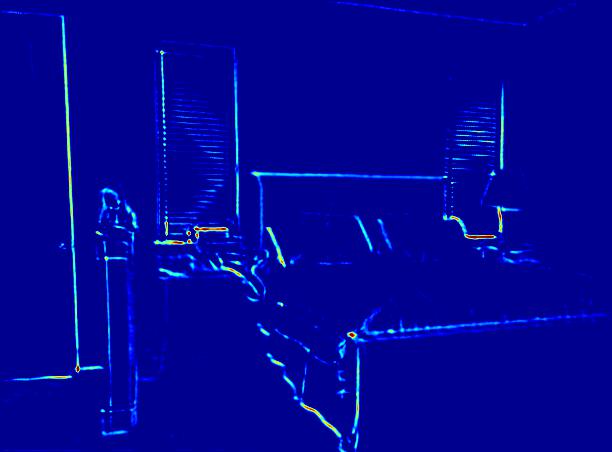} &
\includegraphics[width = .19\linewidth]{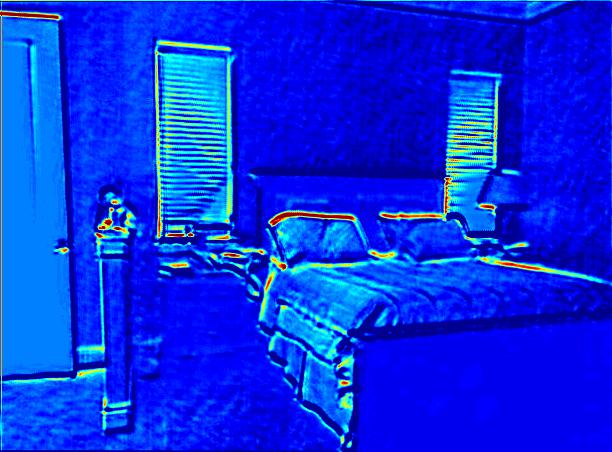} &
\includegraphics[height =.14\linewidth, width = .01\linewidth]{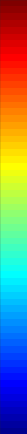} & \\

\includegraphics[width = .19\linewidth, height = .145\linewidth]{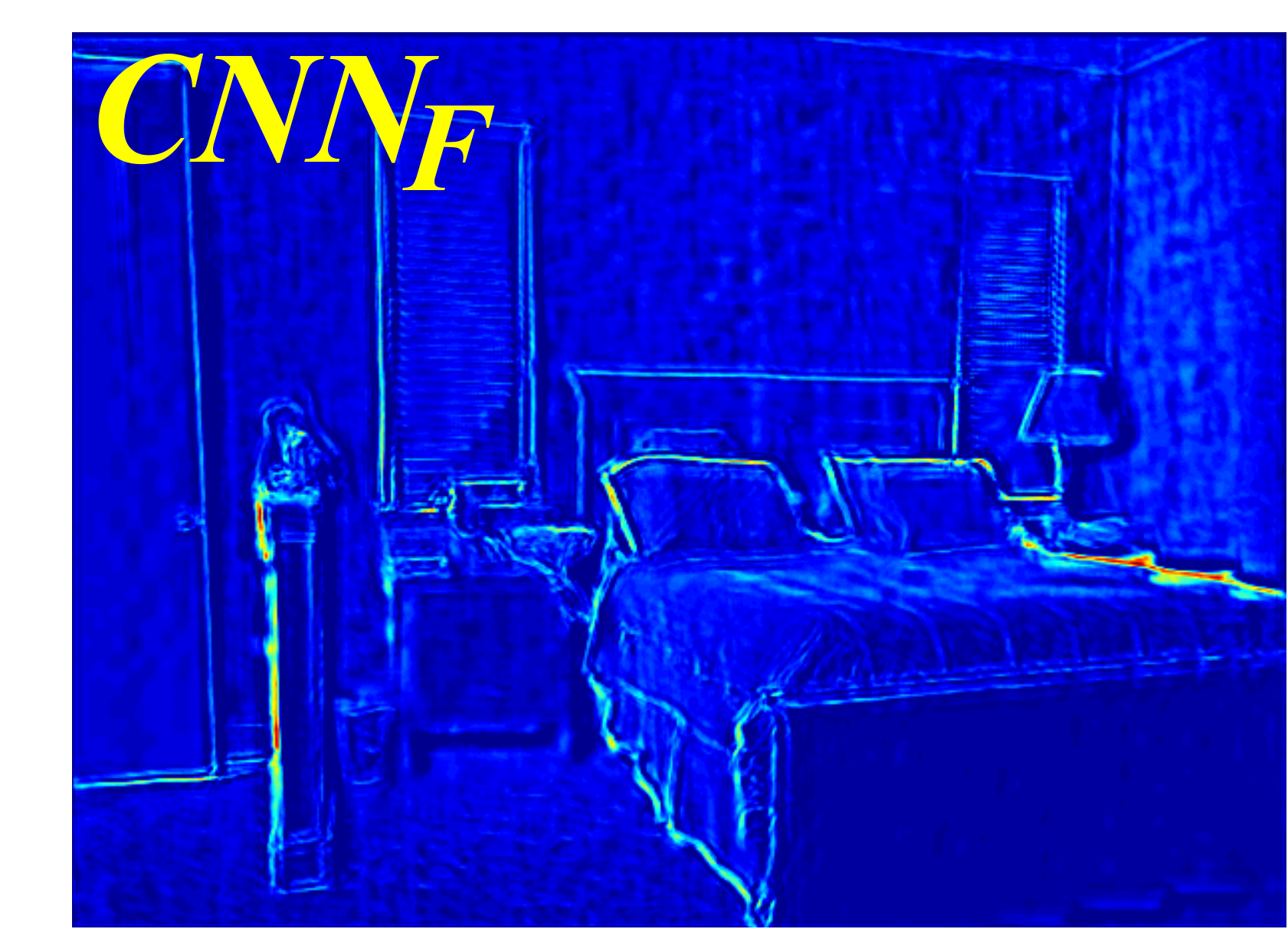} &
\includegraphics[width = .19\linewidth]{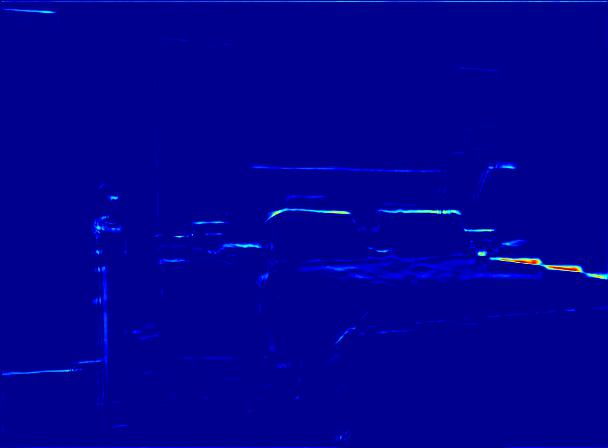} &
\includegraphics[width = .19\linewidth]{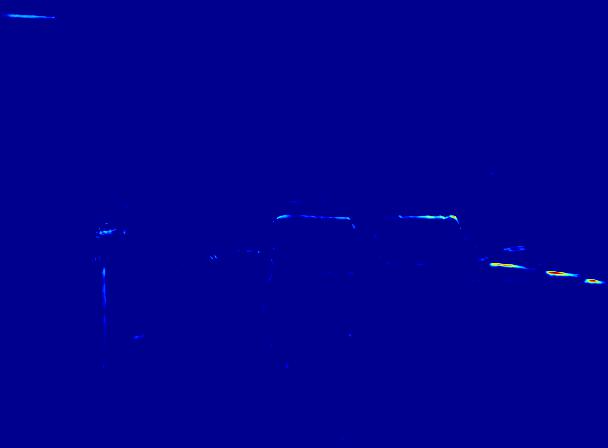} &
\includegraphics[width = .19\linewidth]{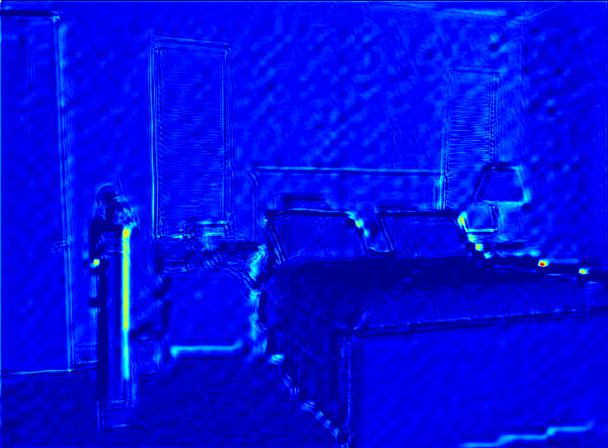} &
\includegraphics[width = .19\linewidth]{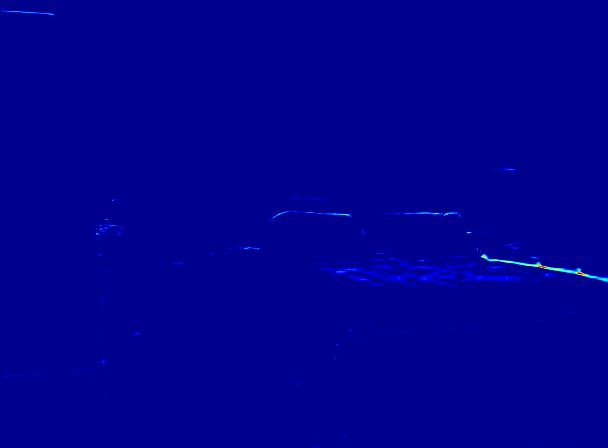} &
\includegraphics[height =.14\linewidth, width = .01\linewidth]{fig/04-6/white.jpg} & \\

\end{tabular}

\caption{
\textbf{Visualization of feature responses.}
Sample feature responses of the input in Figure~\ref{fig:guidance}(a) at the first layer of $\mathrm{CNN_T}$ (top) and $\mathrm{CNN_G}$ (middle), and the second layer of $\mathrm{CNN_F}$ (bottom). For each subnetwork, we select five feature channels and visualize the responses through the colormap.
The corresponding colorbar is shown in the rightmost.
% JB: Just give the COLORBAR directly on the right hand side directly.
% Pixels with \revyj{warmer colors} indicate stronger responses.
%
Note that with the help of $\mathrm{CNN_F}$, inconsistent structures (e.g., the window on the wall) are correctly suppressed.
}
\label{fig:featuremap}
\end{figure*}

\begin{figure*}[t]
\centering

\begin{tabular}{c@{\hspace{0.01\linewidth}}c@{\hspace{0.01\linewidth}}c@{\hspace{0.01\linewidth}}c@{\hspace{0.01\linewidth}}c@{\hspace{0.01\linewidth}}c@{\hspace{0.01\linewidth}}c}
\includegraphics[width = .157\linewidth]{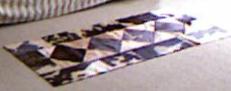} &
\includegraphics[width = .157\linewidth]{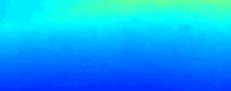} &
\includegraphics[width = .157\linewidth]{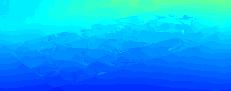} &
\includegraphics[width = .157\linewidth]{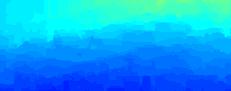} &
\includegraphics[width = .157\linewidth]{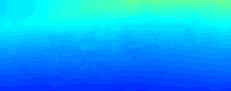} &
\includegraphics[width = .157\linewidth]{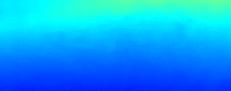} & \\

{(a) Guidance} & {(b) Ground truth}& {(c) JBU \cite{JBU-TOG-2007}} & {(d) Park \cite{Park-ICCV-2011}} & {(e) DJF~\cite{DJF-ECCV-2016}} & {(f) Ours}\\

{} & {RMSE} & {3.64} & {4.84} & {3.24} & {\textbf{2.95}} \\

\end{tabular}
\caption{
\textbf{Selective transfer.}
Comparisons of different joint upsampling methods on handling the texture-copying issue.
The carpet on the floor contains grid-like texture structures that may be incorrectly transferred to the target image. The numbers are the RMSE metric comparing against the GT.
}
\label{fig:block}
\end{figure*}

\subsection{Network training} \label{sec:training}

Since the target and guidance image pair can be from various modalities (e.g., RGB/depth, RGB/NIR), it is infeasible and costly to collect large datasets and train one network for each type of data pair separately.
The goal of our network training, however, is not predicting specific pixel values in
one particular modality.
Instead, we aim to train the network so that it can selectively transfer structures by leveraging the prior from the guidance image.
Hence, we only need to train the network with only one type of image data and then apply the network to other domains.

To demonstrate that the proposed method is insensitive to the training data modality,
we train the network with either the RGB/depth dataset~\cite{NYU-ECCV-2012} or RGB/flow dataset~\cite{Sintel-ECCV-2012}.
We conduct a cross-dataset evaluation (training with one type and evaluate on the other) and show the exemplary results in Figure~\ref{fig:flow_on_depth}.
Figure~\ref{fig:flow_on_depth} (a)-(d) shows the upsampled depth maps
using models trained with different domains of image data.
The flow model refers to the one trained with RGB/flow data for flow map upsampling, while the depth model is trained with RGB/depth data for depth map upsampling.
In Figure~\ref{fig:flow_on_depth}(c), we apply the flow model to upsample the degraded depth map and show competitive results obtained by the depth model (Figure~\ref{fig:flow_on_depth}(d)).
Similar observations on flow map upsampling are also found in Figure~\ref{fig:flow_on_depth} (e)-(h).
%s
Both the models trained with the flow and depth data achieve similar performance.
More filtering results are shown in Section~\ref{sec:exp}, where we evaluate the model with different image data from various domains.
More quantitative results are presented in Table~\ref{table:depth}.

\subsection{What has the network learned?} \label{sec:vis}

%%\vspace{1em}
{\flushleft \textbf{Selective transferring.}}
%MH: check this sentence
Using the learned guidance model $\mathrm{CNN_G}$
alone to transfer details may sometimes be erroneous.
In particular, the structures extracted from the guidance image may not exist in the target image.
The top and middle rows of Figure~\ref{fig:featuremap} show
typical responses at the first layer of $\mathrm{CNN_T}$ and $\mathrm{CNN_G}$.
These two sub-networks show strong responses to edges from the target and guidance images respectively.
Note that there are inconsistent structures in the guidance and target images, e.g., the window on the wall.
The bottom row of Figure~\ref{fig:featuremap} shows sample responses at the second layer of $\mathrm{CNN_F}$.
We observe that the sub-network $\mathrm{CNN_F}$ suppresses inconsistent details.

\begin{figure*}[t]
\centering

\begin{tabular}{c@{\hspace{0.005\linewidth}}c@{\hspace{0.005\linewidth}}c@{\hspace{0.005\linewidth}}c@{\hspace{0.005\linewidth}}c@{\hspace{0.005\linewidth}}c@{\hspace{0.005\linewidth}}c}
\includegraphics[width = .24\linewidth]{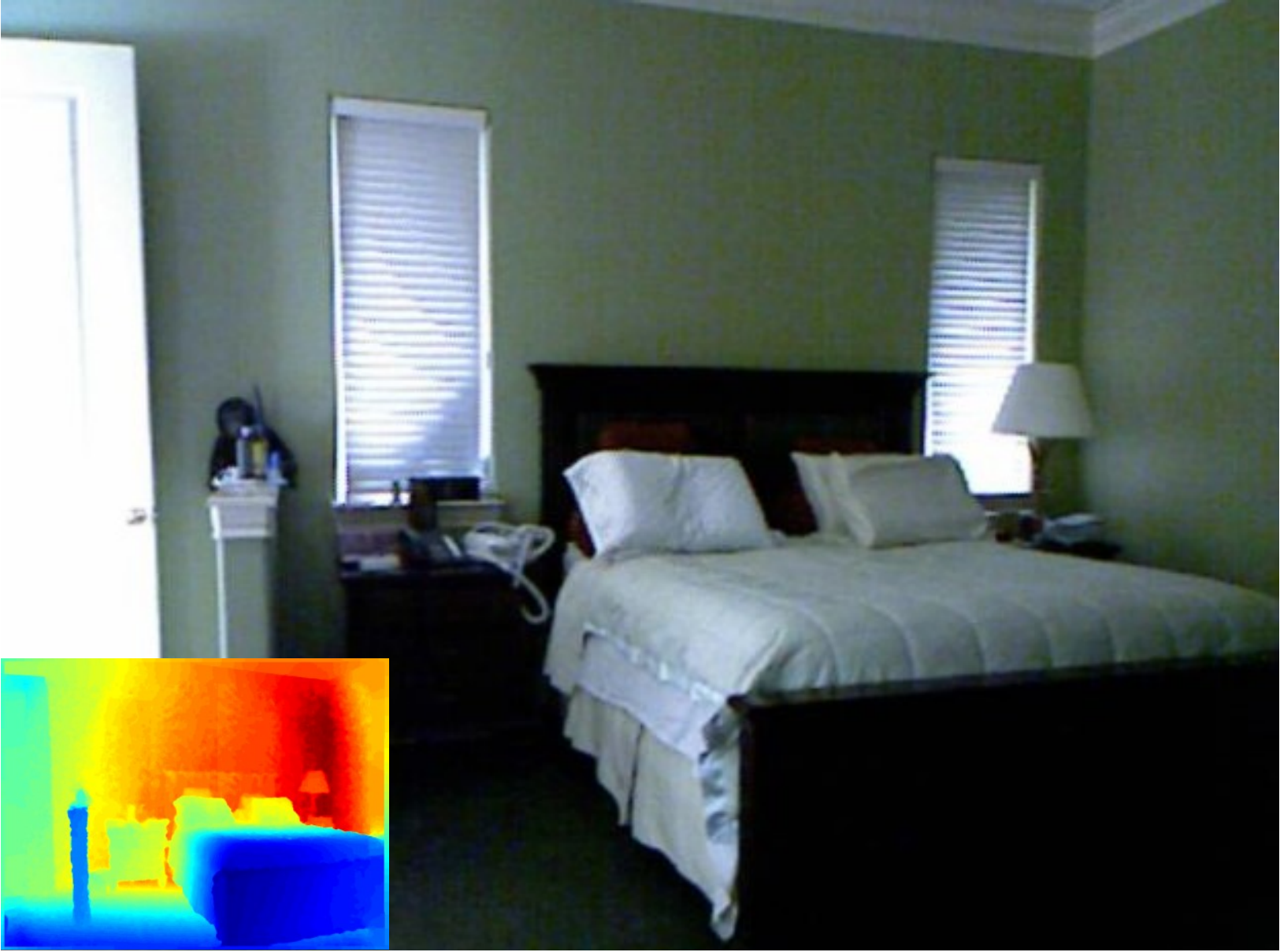} &
\includegraphics[width = .24\linewidth]{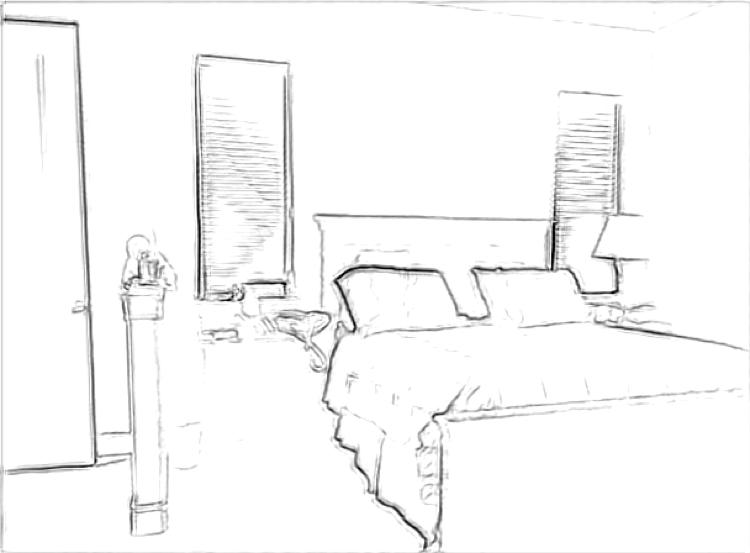} &
\includegraphics[width = .24\linewidth]{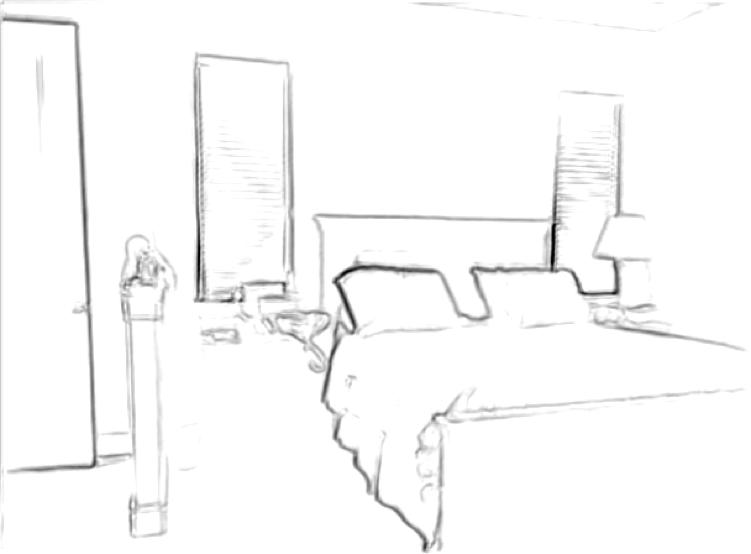} &
\includegraphics[width = .24\linewidth]{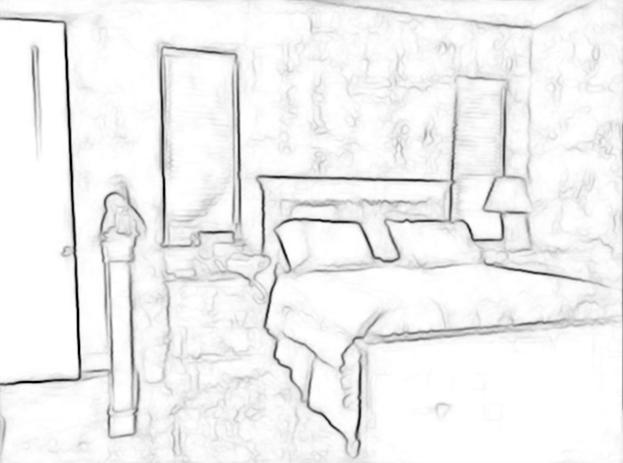} & \\

\includegraphics[width = .24\linewidth]{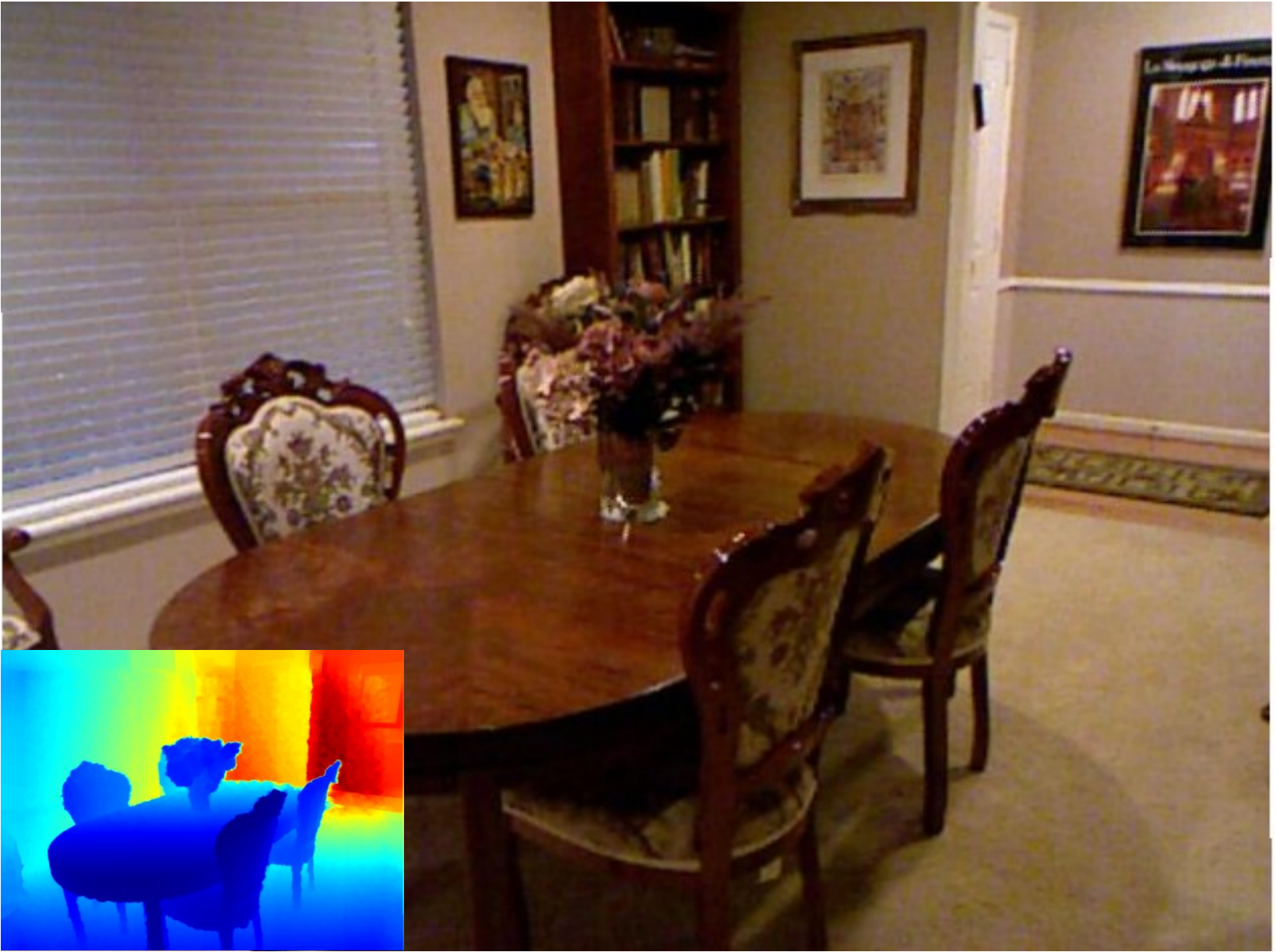} &
\includegraphics[width = .24\linewidth]{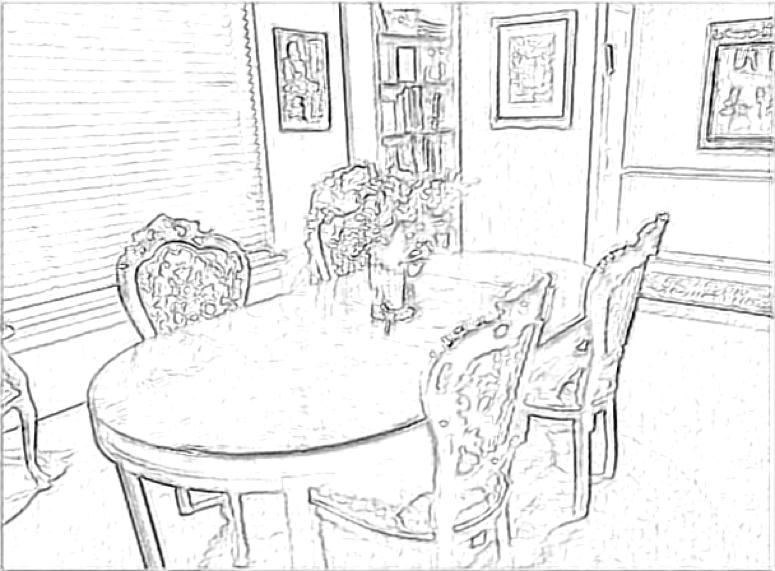} &
\includegraphics[width = .24\linewidth]{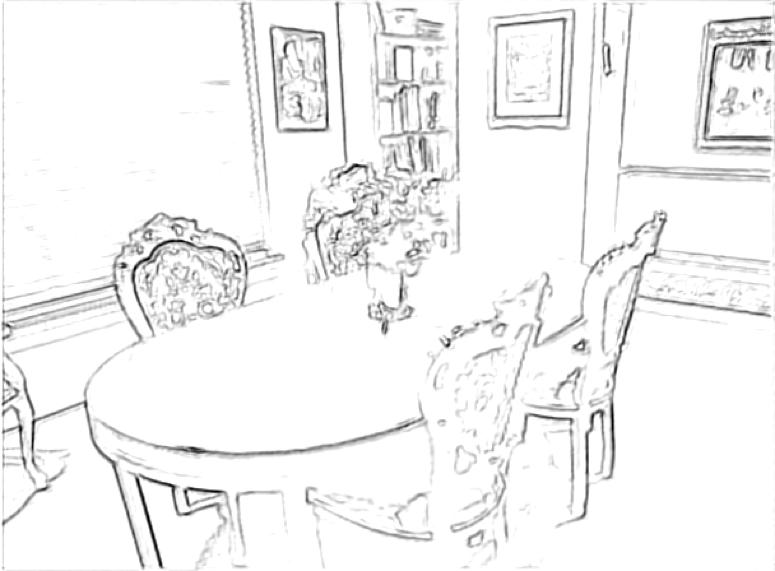} &
\includegraphics[width = .24\linewidth]{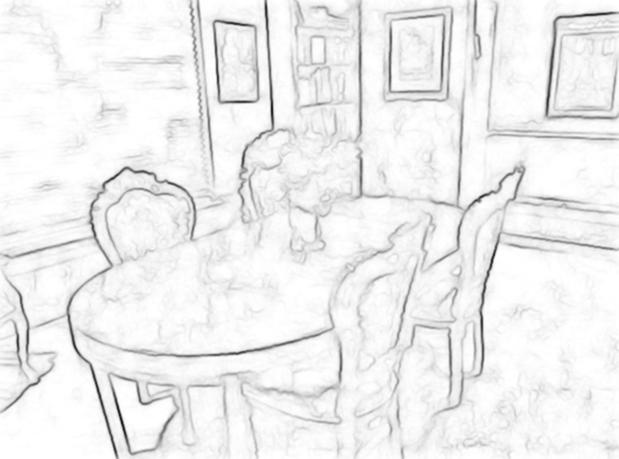} & \\

{(a) Guidance } & {(b) Learned guidance } &{(c) Our learned guidance}& {(d) Edge map~\cite{Dollar-ICCV-2013}} \\
{ (inset: GT depth)} & {w/o skip connection~\cite{DJF-ECCV-2016}} &{w/ skip connection}& {} \\

\end{tabular}
\caption{
\textbf{Visualization of the learned guidance map.}
Comparison between the learned guidance feature maps from $\mathrm{CNN_G}$ and edge maps from~\cite{Dollar-ICCV-2013}.
The network $\mathrm{CNN_G}$ is capable of extracting informative, salient structures from the guidance image for content transfer.
\revyj{Furthermore, with the skip connection, the learned guidance maps in (c) are cleaner than that in (b) by suppressing inconsistent structures (edges on the window and wall) in the target/guidance pair.}
}
\label{fig:guidance}
\end{figure*}

We present another example in Figure~\ref{fig:block}.
We note that the ground truth depth map of the selected region is smooth.
However, due to the high contrast patterns on the mat in the guidance image, several methods, e.g., \cite{JBU-TOG-2007,Park-ICCV-2011}, incorrectly transfer the mat structure to the upsampled depth map.
The reason is that these methods~\cite{JBU-TOG-2007,Park-ICCV-2011} rely only on structures in the guidance image.
The problem, commonly known as texture-copying artifacts, often occurs when the texture in the guidance image has strong color contrast.
With the help of the $\mathrm{CNN_F}$, our method successfully blocks the texture structure in the guidance image (Figure~\ref{fig:block}(f)).

%\vspace{1em}
{\flushleft \textbf{Output of $\mathrm{CNN_G}$.}}
In Figure \ref{fig:guidance}(c), we show the learned guidance from $\mathrm{CNN_G}$ using two examples from the NYU v2 dataset \cite{NYU-ECCV-2012}.
In general, the learned guidance appears to be similar to
an edge map highlighting the salient structures in the guidance image.
We show edge detection results from~\cite{Dollar-ICCV-2013} in Figure~\ref{fig:guidance}(d).
Both results show strong responses to the main structures, but the guidance map generated by $\mathrm{CNN_G}$ appears to detect sharper boundaries while suppressing responses to small-scale textures, e.g., the wall in the first example.
The result suggests that using only $\mathrm{CNN_F}$ (Figure~\ref{fig:problem}(c)) does not perform well due to lack of the salient feature extraction step from the sub-network $\mathrm{CNN_G}$.

\revyj{To demonstrate the effectiveness of the skip connection, we compare the learned guidance without and with 
the skip connection in Figure~\ref{fig:guidance}(b) and (c).
Adding the skip connection helps suppress more inconsistent structures (e.g., edges on the bed, wall, table) in the target/guidance pair, and consequently the residual-based model effectively alleviates texture-copying artifacts.
}

\subsection{Relationship to prior work}

The proposed framework is closely related to weighted-average, optimization-based, and CNN-based models.
In each layer of the network, the convolutional filters also perform the
weighted-average process.
In this context, our filter is similar to the weighted-average filters.
\revyj{The key difference is that the weights in this work are learned from data %by considering both contents of the target and guidance images.
while those of the weighted-average filters~\cite{BF-ICCV-1998,JBU-TOG-2007} 
are pre-defined based on color or gradient features.
The proposed network plays a similar role in 
the fidelity and regularization terms defined in the optimization-based joint filters.
Specifically, the training objective in~\eqref{formula1} corresponds to the fidelity term 
of the weighted-average filters~\cite{BF-ICCV-1998,JBU-TOG-2007} 
as it encourages the output to be as close to the ground truth as possible.
The skip connection implicitly serves as the regularization term by enforcing adjacent pixels to share similar values (e.g., depth) as it directly bypasses the low-quality target input to the output of the network.
%Through a data-driven approach and the incorporation of skip connection, our model implicitly ensures that the output does not deviate significantly from the target image while sharing salient structures with the guidance image.
}
For CNN-based models, our network architecture can be viewed as a unified model for different tasks.
For example, if we remove $\mathrm{CNN_G}$ and use only $\mathrm{CNN_T}$ and $\mathrm{CNN_F}$, the resulting network architecture resembles an image restoration model, e.g., SRCNN~\cite{SRCNN-ECCV-2014}.
On the other hand, in cases of removing $\mathrm{CNN_T}$, the remaining $\mathrm{CNN_G}$ and $\mathrm{CNN_F}$ can be viewed as one
using CNNs for depth prediction~\cite{David-NIPS-2014}.

\begin{table*}[t]
\caption{\textbf{Quantitative comparisons on depth upsampling.}
Comparisons with the state-of-the-art methods in terms of RMSE.
The depth values are scaled to the range $[0, 255]$ for
the Middlebury\cite{Midd1-CVPR-2007,Midd2-CVPR-2007}, %Lu~\cite{Lu-CVPR-2014}
and SUN RGB-D~\cite{Song-CVPR-2015} datasets. For the NYU v2 dataset~\cite{NYU-ECCV-2012}, the depth values are measured in \textbf{centimeter}.
Note that the depth maps in the SUN RGB-D dataset may contain missing regions due to the limitation of depth sensors.
We ignore these pixels in calculating the RMSE.
Numbers in bold indicate the best performance and underscored numbers indicate the second best.
\revyj{The mean and standard deviation of the RMSE values are shown in each entry.}
}
\label{table:depth}
\centering

\begin{tabular}{lccc}
\toprule
~ & \textbf{Middlebury \cite{Midd1-CVPR-2007,Midd2-CVPR-2007}} & \textbf{NYU v2 \cite{NYU-ECCV-2012}} & \textbf{SUN RGB-D \cite{Song-CVPR-2015}} \\
\multicolumn{1}{c}{} & {4$\times$~~~~~~~~~~~~ ~8$\times$~~~~~~~~~~~~ 16$\times$} & {4$\times$~~~~~~~~~~~ ~~~8$\times$~~~~~~~~~~~~ 16$\times$} & {4$\times$~~~~~~~~~~~~ ~8$\times$~~~~~~~~~~~~ 16$\times$}\\

\midrule
Bicubic &4.44~{\scriptsize $\pm$}~{\scriptsize 1.59}~ 7.58~{\scriptsize $\pm$}~{\scriptsize 2.69}~ 11.87~{\scriptsize $\pm$}~{\scriptsize 4.04} &~~~8.16~{\scriptsize $\pm$}~{\scriptsize 4.37}~ 14.22~{\scriptsize $\pm$}~{\scriptsize 7.56}~ 22.32~{\scriptsize $\pm$}~{\scriptsize 11.68} & 2.09~{\scriptsize $\pm$}~{\scriptsize 1.56}~ 3.45~{\scriptsize $\pm$}~{\scriptsize 2.23}~ 5.48~{\scriptsize $\pm$}~{\scriptsize 3.21} \\
MRF \cite{MRF-NIPS-2005} &4.26~{\scriptsize $\pm$}~{\scriptsize 1.52}~ 7.43~{\scriptsize $\pm$}~{\scriptsize 2.63}~ 11.80~{\scriptsize $\pm$}~{\scriptsize 4.01} &~~~7.84~{\scriptsize $\pm$}~{\scriptsize 4.20}~ 13.98~{\scriptsize $\pm$}~{\scriptsize 7.42}~ 22.20~{\scriptsize $\pm$}~{\scriptsize 11.61} & 1.99~{\scriptsize $\pm$}~{\scriptsize 1.57}~ 3.38~{\scriptsize $\pm$}~{\scriptsize 2.19}~ 5.45~{\scriptsize $\pm$}~{\scriptsize 3.18} \\
GF \cite{He-PAMI-2013} &4.01~{\scriptsize $\pm$}~{\scriptsize 1.42}~ 7.22~{\scriptsize $\pm$}~{\scriptsize 2.55}~ 11.70~{\scriptsize $\pm$}~{\scriptsize 3.97} &~~~7.32~{\scriptsize $\pm$}~{\scriptsize 3.86}~ 13.62~{\scriptsize $\pm$}~{\scriptsize 7.20}~ 22.03~{\scriptsize $\pm$}~{\scriptsize 11.51} & 1.91~{\scriptsize $\pm$}~{\scriptsize 1.43}~ 3.31~{\scriptsize $\pm$}~{\scriptsize 2.15}~ 5.41~{\scriptsize $\pm$}~{\scriptsize 3.17} \\

JBU \cite{JBU-TOG-2007} &2.44~{\scriptsize $\pm$}~{\scriptsize 0.86}~ 3.81~{\scriptsize $\pm$}~{\scriptsize 1.49}~ ~6.13~{\scriptsize $\pm$}~{\scriptsize 2.34} &~4.07~{\scriptsize $\pm$}~{\scriptsize 2.22}~~ ~8.29~{\scriptsize $\pm$}~{\scriptsize 4.47}~ 13.35~{\scriptsize $\pm$}~{\scriptsize 7.47} & 1.37~{\scriptsize $\pm$}~{\scriptsize 1.12}~ 2.01~{\scriptsize $\pm$}~{\scriptsize 1.76}~ 3.15~{\scriptsize $\pm$}~{\scriptsize 2.58} \\

TGV \cite{TGV-ICCV-2013} &3.39~{\scriptsize $\pm$}~{\scriptsize 1.25}~ 5.41~{\scriptsize $\pm$}~{\scriptsize 1.99}~ 12.03~{\scriptsize $\pm$}~{\scriptsize 4.17} &~~6.98~{\scriptsize $\pm$}~{\scriptsize 3.61}~ 11.23~{\scriptsize $\pm$}~{\scriptsize 5.46}~ 28.13~{\scriptsize $\pm$}~{\scriptsize 10.47} & 1.94~{\scriptsize $\pm$}~{\scriptsize 1.31}~ 3.01~{\scriptsize $\pm$}~{\scriptsize 2.46}~ 5.87~{\scriptsize $\pm$}~{\scriptsize 3.46} \\
Park \cite{Park-ICCV-2011} &2.82~{\scriptsize $\pm$}~{\scriptsize 0.94}~ 4.08~{\scriptsize $\pm$}~{\scriptsize 1.43}~ ~7.26~{\scriptsize $\pm$}~{\scriptsize 2.41} &~5.21~{\scriptsize $\pm$}~{\scriptsize 2.64}~ ~9.56~{\scriptsize $\pm$}~{\scriptsize 4.41}~ 18.10~{\scriptsize $\pm$}~{\scriptsize 8.29} & 1.78~{\scriptsize $\pm$}~{\scriptsize 1.33}~ 2.76~{\scriptsize $\pm$}~{\scriptsize 1.99}~ 4.77~{\scriptsize $\pm$}~{\scriptsize 2.97} \\
Ham \cite{Ham-CVPR-2015} &3.14~{\scriptsize $\pm$}~{\scriptsize 1.24}~ 5.03~{\scriptsize $\pm$}~{\scriptsize 2.08}~ ~8.83~{\scriptsize $\pm$}~{\scriptsize 3.96} &~5.27~{\scriptsize $\pm$}~{\scriptsize 2.86}~~12.31~{\scriptsize $\pm$}~{\scriptsize 6.07}~ 19.24~{\scriptsize $\pm$}~{\scriptsize 9.64} & 1.67~{\scriptsize $\pm$}~{\scriptsize 1.41}~ 2.60~{\scriptsize $\pm$}~{\scriptsize 2.31}~ 4.36~{\scriptsize $\pm$}~{\scriptsize 3.32} \\

DMSG \cite{Tai-2016-depth} &\textbf{1.79~{\scriptsize $\pm$}~{\scriptsize 0.66}}~ \textbf{3.39~{\scriptsize $\pm$}~{\scriptsize 1.28}}~ ~\textbf{5.87~{\scriptsize $\pm$}~{\scriptsize 2.38}} &~\underline{3.48~{\scriptsize $\pm$}~{\scriptsize 1.96}}~~~\underline{6.07~{\scriptsize $\pm$}~{\scriptsize 3.26}}~ ~10.27~{\scriptsize $\pm$}~{\scriptsize 5.79} & 1.30~{\scriptsize $\pm$}~{\scriptsize 1.12}~ \underline{1.80~{\scriptsize $\pm$}~{\scriptsize 1.31}}~ 2.81~{\scriptsize $\pm$}~{\scriptsize 1.92} \\

FBS \cite{Barron-2016-solver} &2.58~{\scriptsize $\pm$}~{\scriptsize 0.88}~ 4.19~{\scriptsize $\pm$}~{\scriptsize 1.48}~ 7.30~{\scriptsize $\pm$}~{\scriptsize 2.49} &~4.29~{\scriptsize $\pm$}~{\scriptsize 2.53}~~~8.94~{\scriptsize $\pm$}~{\scriptsize 4.68}~ ~14.59~{\scriptsize $\pm$}~{\scriptsize 8.32} & 1.58~{\scriptsize $\pm$}~{\scriptsize 1.41}~ 2.27~{\scriptsize $\pm$}~{\scriptsize 2.33}~ 3.76~{\scriptsize $\pm$}~{\scriptsize 3.01} \\

\midrule

Ours-flow~ &2.31~{\scriptsize $\pm$}~{\scriptsize 0.84}~~ 3.95~{\scriptsize $\pm$}~{\scriptsize 1.45}~ 6.34~{\scriptsize $\pm$}~{\scriptsize 2.44} & ~4.42~{\scriptsize $\pm$}~{\scriptsize 2.77}~~ 7.32~{\scriptsize $\pm$}~{\scriptsize 3.91}~ 11.62~{\scriptsize $\pm$}~{\scriptsize 6.58} &~1.36~{\scriptsize $\pm$}~{\scriptsize 1.14}~ 1.91~{\scriptsize $\pm$}~{\scriptsize 1.37}~ 2.90~{\scriptsize $\pm$}~{\scriptsize 2.04}\\

DJF~\cite{DJF-ECCV-2016}~ &2.14~{\scriptsize $\pm$}~{\scriptsize 0.69}~~ 3.77~{\scriptsize $\pm$}~{\scriptsize 1.32}~ 6.12~{\scriptsize $\pm$}~{\scriptsize 2.19} & ~3.54~{\scriptsize $\pm$}~{\scriptsize 1.86}~~ 6.20~{\scriptsize $\pm$}~{\scriptsize 3.26}~ \underline{10.21~{\scriptsize $\pm$}~{\scriptsize 5.57}} &~\underline{1.28~{\scriptsize $\pm$}~{\scriptsize 1.02}}~ 1.81~{\scriptsize $\pm$}~{\scriptsize 1.35}~ \underline{2.78~{\scriptsize $\pm$}~{\scriptsize 1.93}} \\

Ours~ &\underline{1.98~{\scriptsize $\pm$}~{\scriptsize 0.67}}~~ \underline{3.61~{\scriptsize $\pm$}~{\scriptsize 1.39}}~ \underline{6.07~{\scriptsize $\pm$}~{\scriptsize 2.20}} & ~\textbf{3.38~{\scriptsize $\pm$}~{\scriptsize 1.95}}~~ \textbf{5.86~{\scriptsize $\pm$}~{\scriptsize 3.14}}~ \textbf{10.11~{\scriptsize $\pm$}~{\scriptsize 5.49}} &~\textbf{1.27~{\scriptsize $\pm$}~{\scriptsize 0.98}}~ \textbf{1.77~{\scriptsize $\pm$}~{\scriptsize 1.30}}~ \textbf{2.75~{\scriptsize $\pm$}~{\scriptsize 1.94}} \\
\bottomrule
\end{tabular}
\end{table*}

\begin{figure*}[t]
\centering
\begin{tabular}{c@{\hspace{0.005\linewidth}}c@{\hspace{0.005\linewidth}}c@{\hspace{0.005\linewidth}}c@{\hspace{0.005\linewidth}}c@{\hspace{0.005\linewidth}}c@{\hspace{0.005\linewidth}}c@{\hspace{0.005\linewidth}}c}

\includegraphics[height=.1\linewidth, width = .131\linewidth]{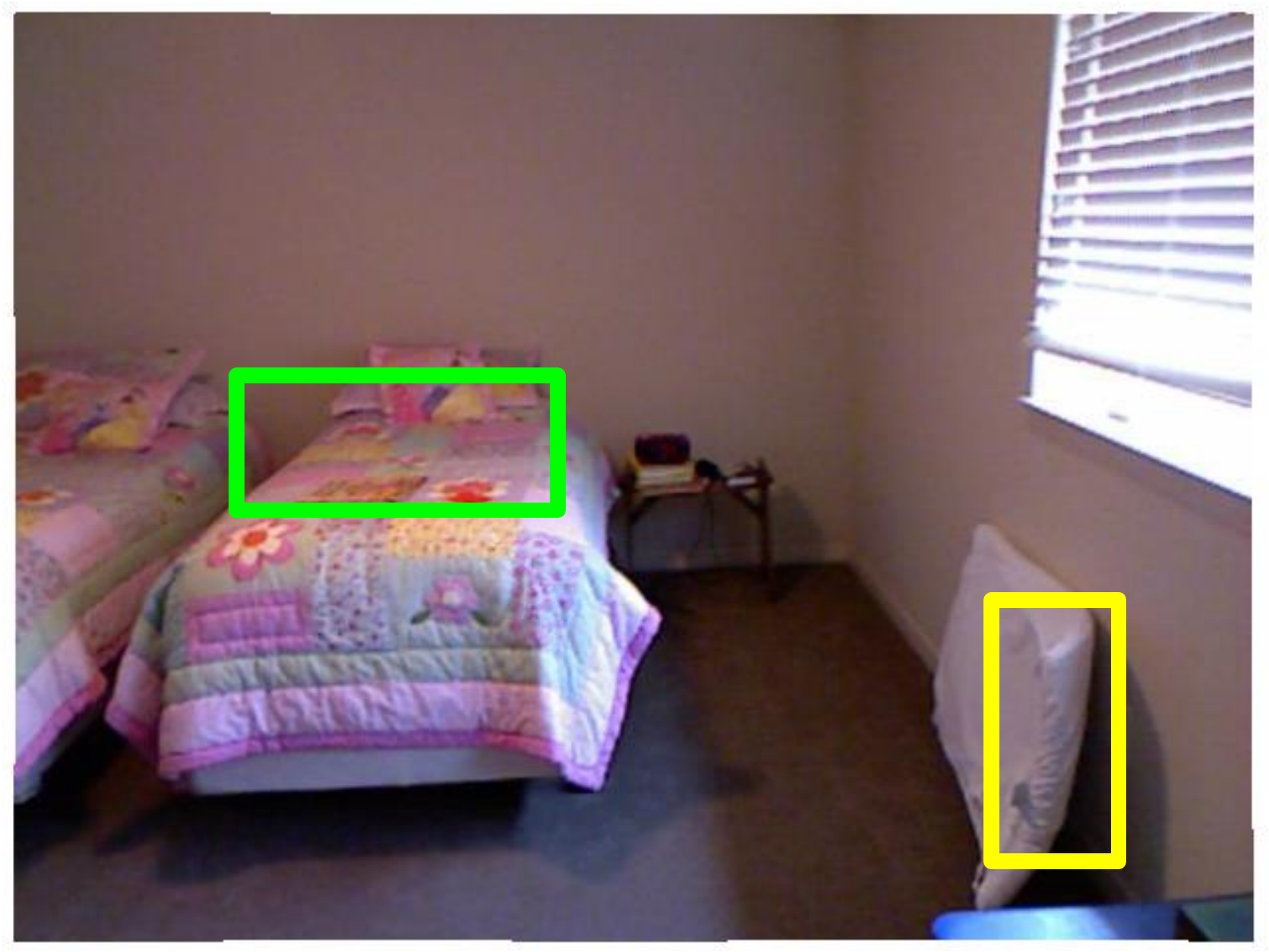} &
\includegraphics[height=.1\linewidth, width = .131\linewidth]{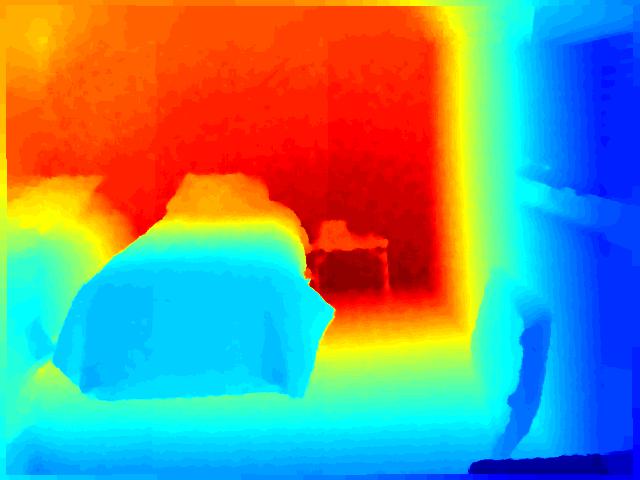} &
\includegraphics[height=.1\linewidth, width = .131\linewidth]{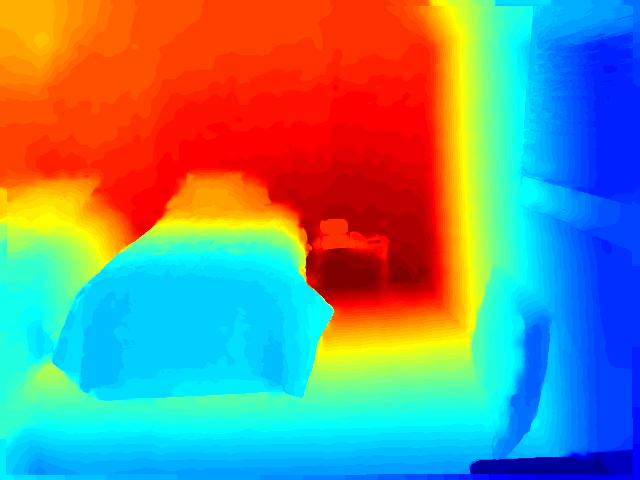} &
\includegraphics[height=.1\linewidth, width = .131\linewidth]{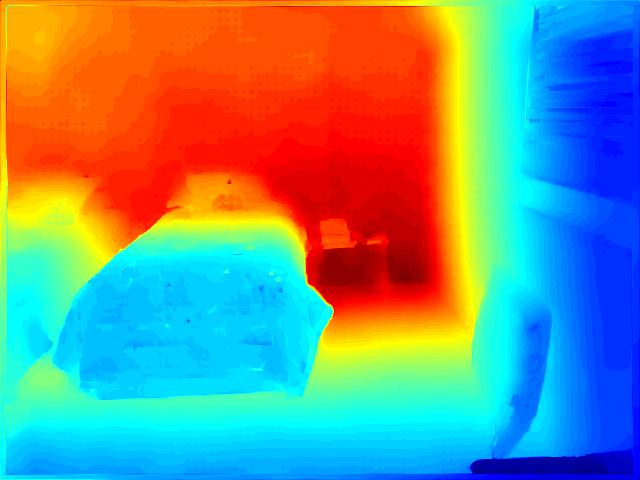} &
\includegraphics[height=.1\linewidth, width = .131\linewidth]{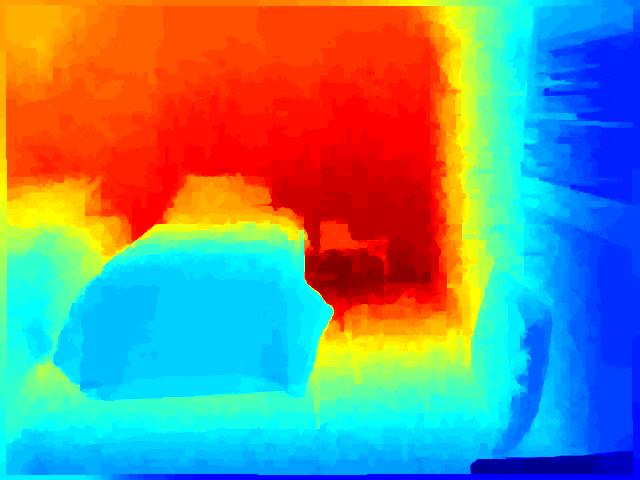} &
\includegraphics[height=.1\linewidth, width = .131\linewidth]{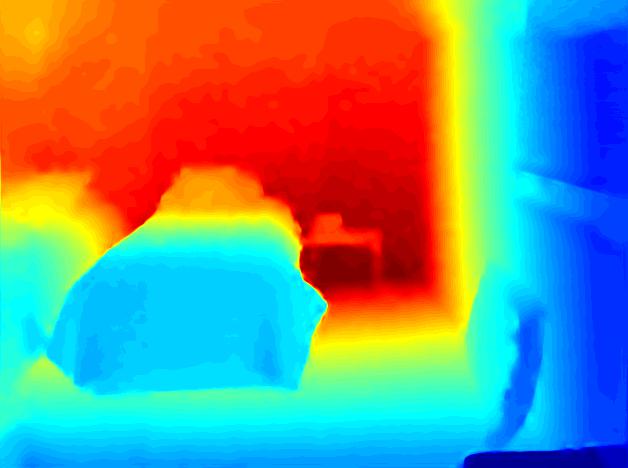} &
\includegraphics[height=.1\linewidth, width = .131\linewidth]{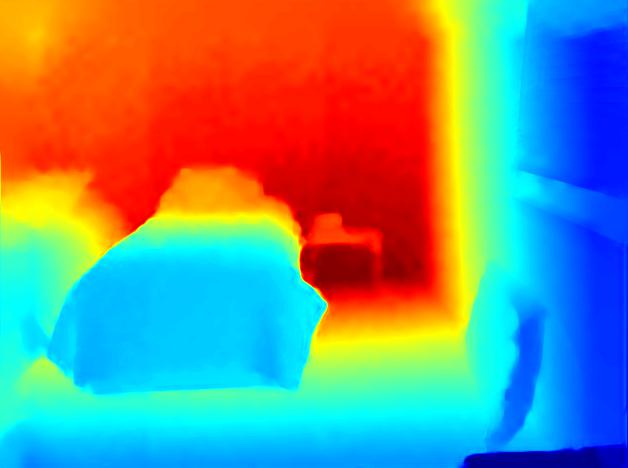} & \\

\includegraphics[height=.04\linewidth, width = .131\linewidth]{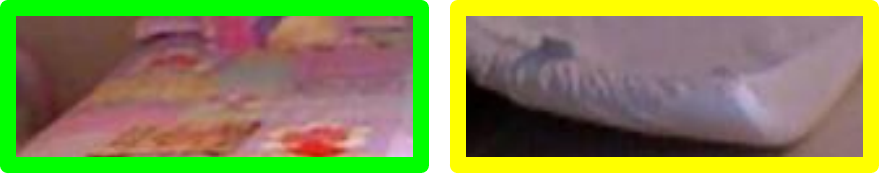} &
\includegraphics[height=.04\linewidth, width = .131\linewidth]{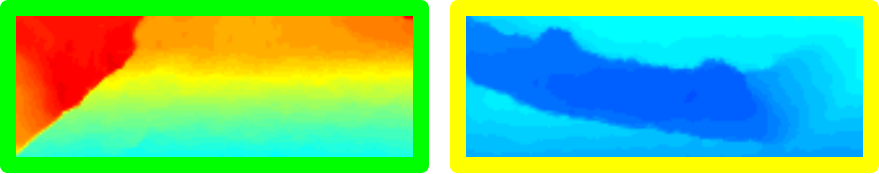} &
\includegraphics[height=.04\linewidth, width = .131\linewidth]{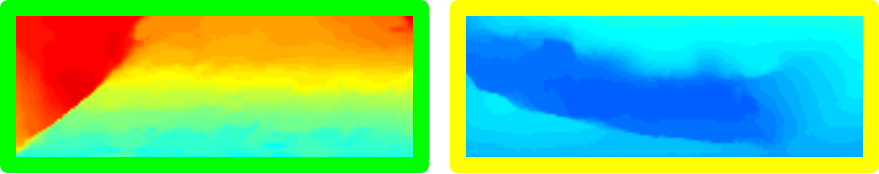} &
\includegraphics[height=.04\linewidth, width = .131\linewidth]{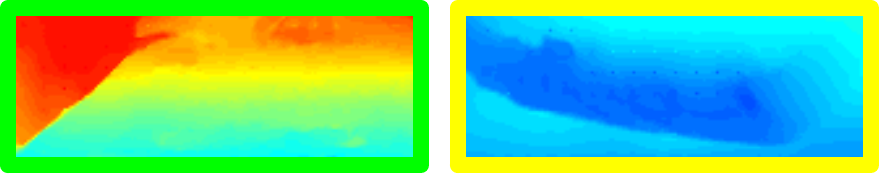} &
\includegraphics[height=.04\linewidth, width = .131\linewidth]{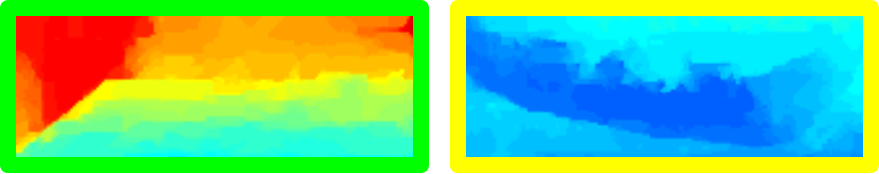} &
\includegraphics[height=.04\linewidth, width = .131\linewidth]{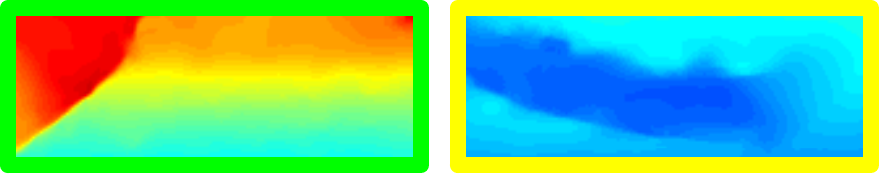} &
\includegraphics[height=.04\linewidth, width = .131\linewidth]{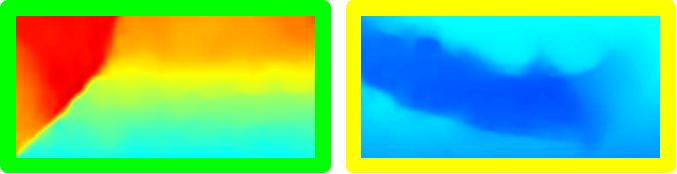} & \\

{ (a) Guidance }& { (b) GT } & { (c) JBU \cite{JBU-TOG-2007}} & { (d) TGV \cite{TGV-ICCV-2013}} & { (e) Park \cite{Park-ICCV-2011}} & { (f) DJF\cite{DJF-ECCV-2016}} & { (g) Ours}\\

{}&{ RMSE}&{ 4.27}&{ 5.62} & { 4.72} & { 3.07} & { \textbf{2.93}} & \\

\includegraphics[height=.1\linewidth, width = .131\linewidth]{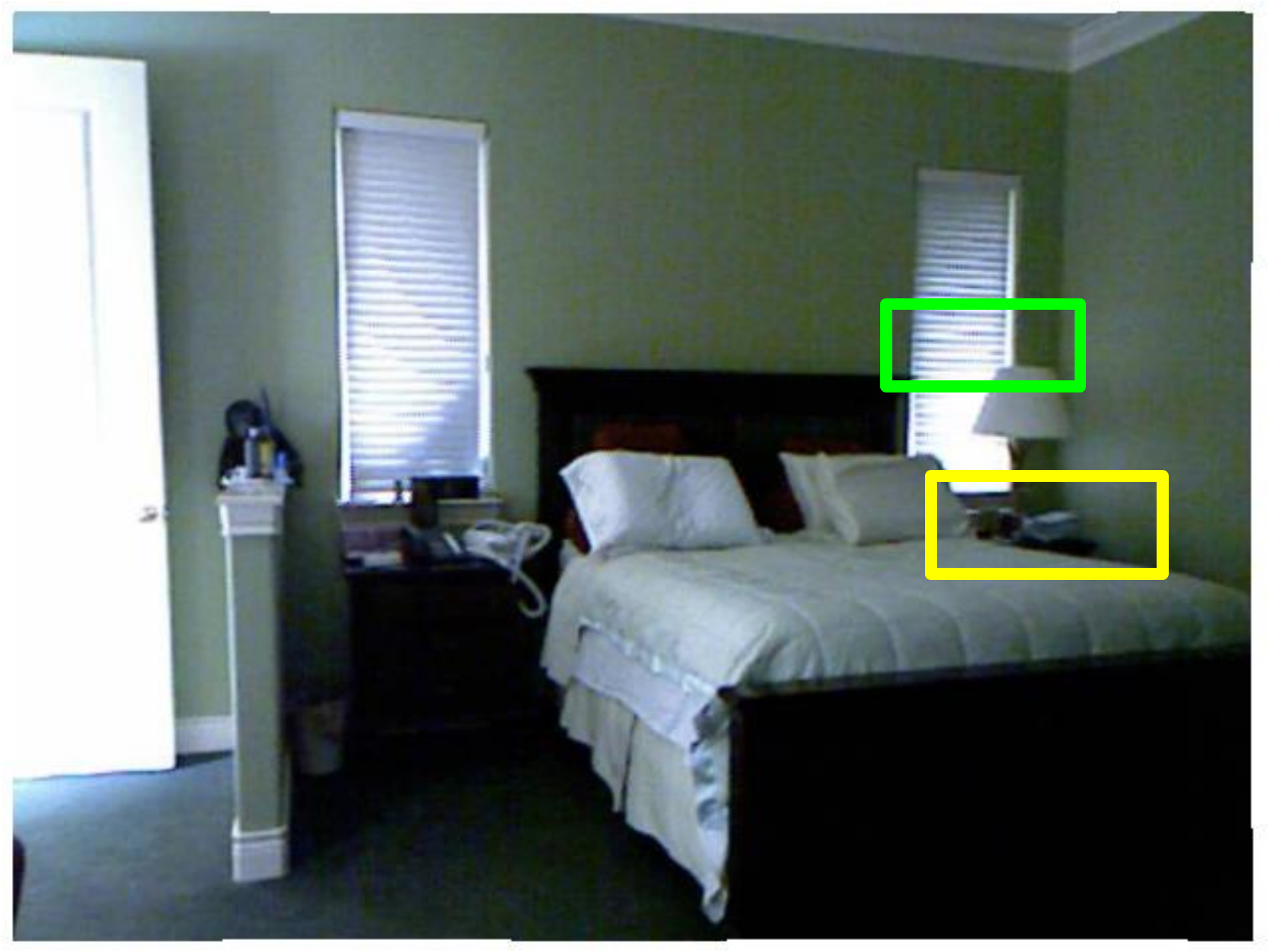} &
\includegraphics[height=.1\linewidth, width = .131\linewidth]{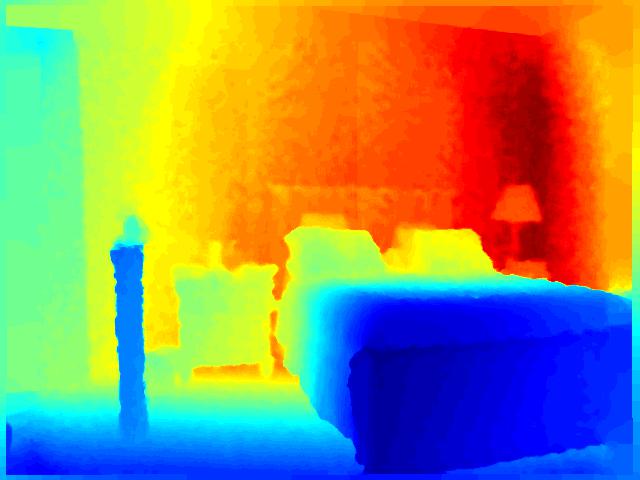} &
\includegraphics[height=.1\linewidth, width = .131\linewidth]{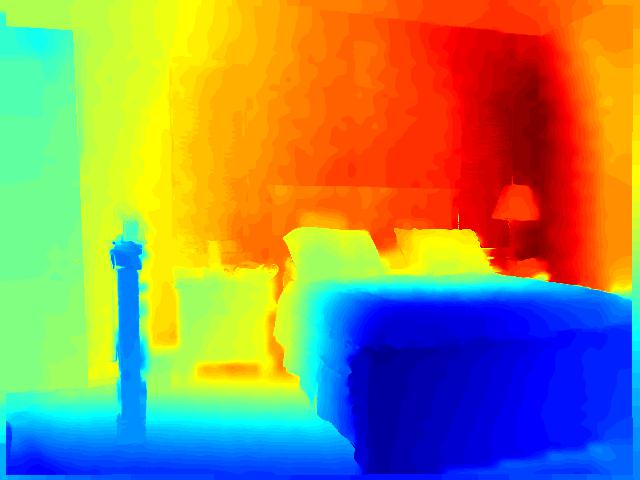} &
\includegraphics[height=.1\linewidth, width = .131\linewidth]{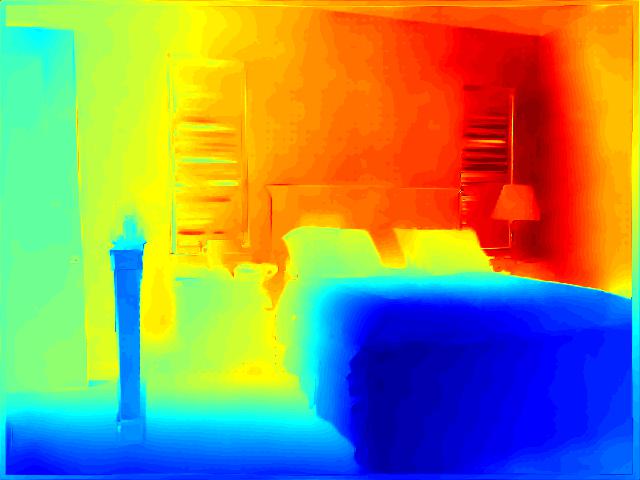} &
\includegraphics[height=.1\linewidth, width = .131\linewidth]{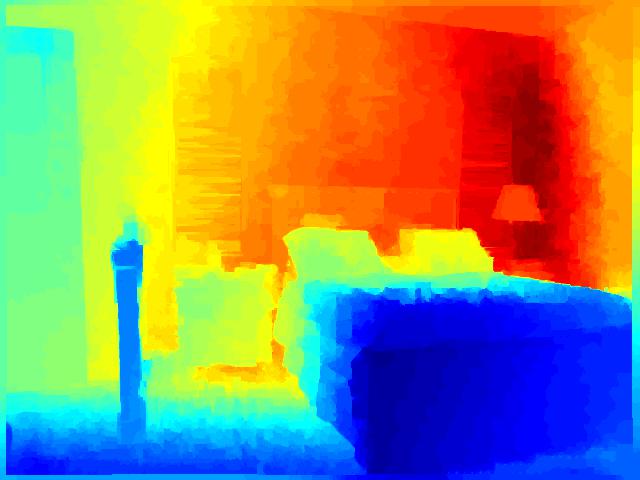} &
\includegraphics[height=.1\linewidth, width = .131\linewidth]{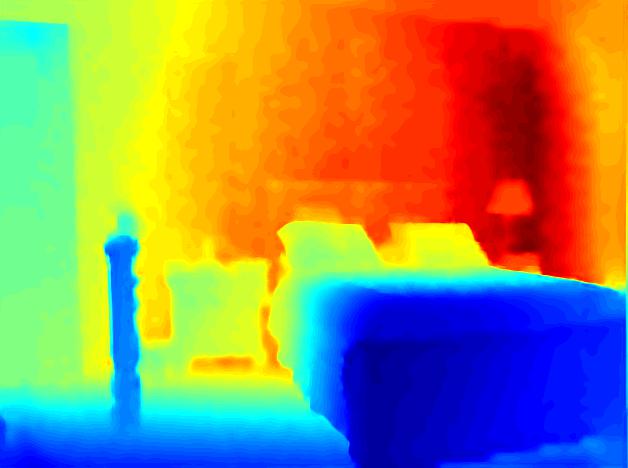} &
\includegraphics[height=.1\linewidth, width = .131\linewidth]{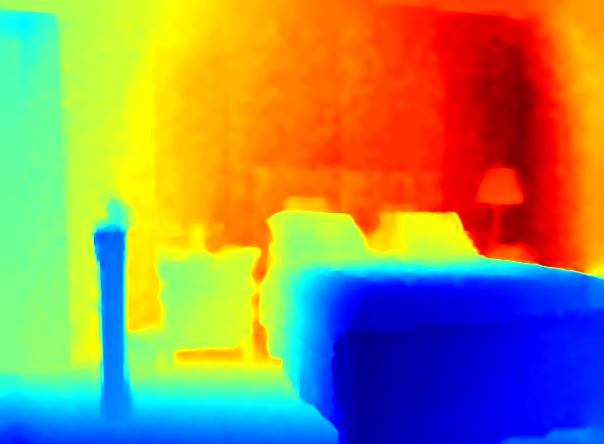} & \\

\includegraphics[height=.04\linewidth, width = .131\linewidth]{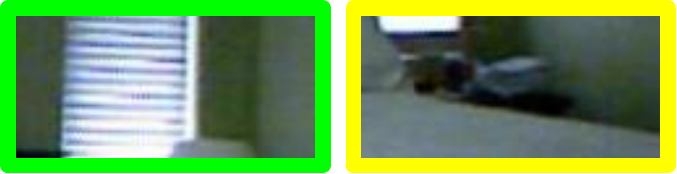} &
\includegraphics[height=.04\linewidth, width = .131\linewidth]{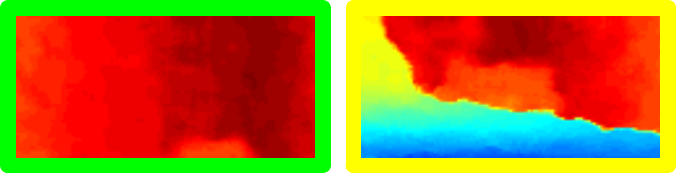} &
\includegraphics[height=.04\linewidth, width = .131\linewidth]{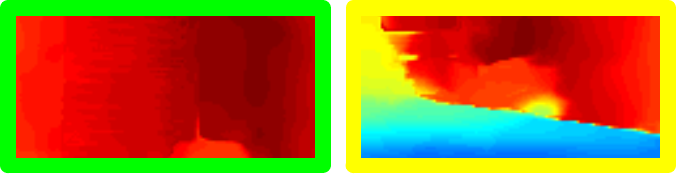} &
\includegraphics[height=.04\linewidth, width = .131\linewidth]{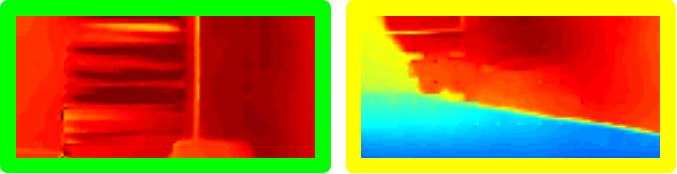} &
\includegraphics[height=.04\linewidth, width = .131\linewidth]{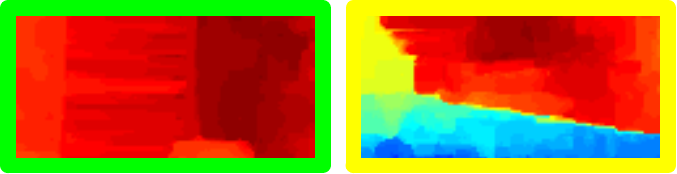} &
\includegraphics[height=.04\linewidth, width = .131\linewidth]{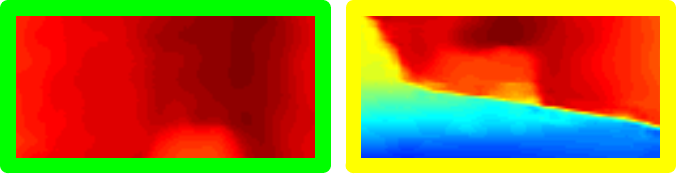} &
\includegraphics[height=.04\linewidth, width = .131\linewidth]{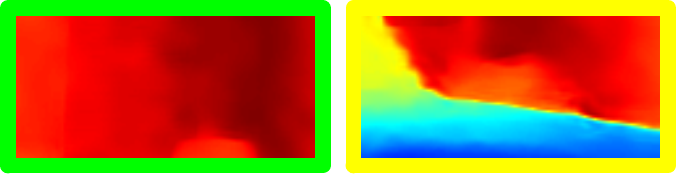} &\\

{ (a) Guidance }& { (b) GT } & { (c) JBU \cite{JBU-TOG-2007}} & { (d) TGV \cite{TGV-ICCV-2013}} & { (e) Park \cite{Park-ICCV-2011}} & { (f) DJF\cite{DJF-ECCV-2016}} & { (g) Ours}\\

{}&{ RMSE}&{ 4.98}&{ 7.24} & { 6.24} & { 3.80} & { \textbf{3.68}} & \\

\includegraphics[height=.1\linewidth, width = .131\linewidth]{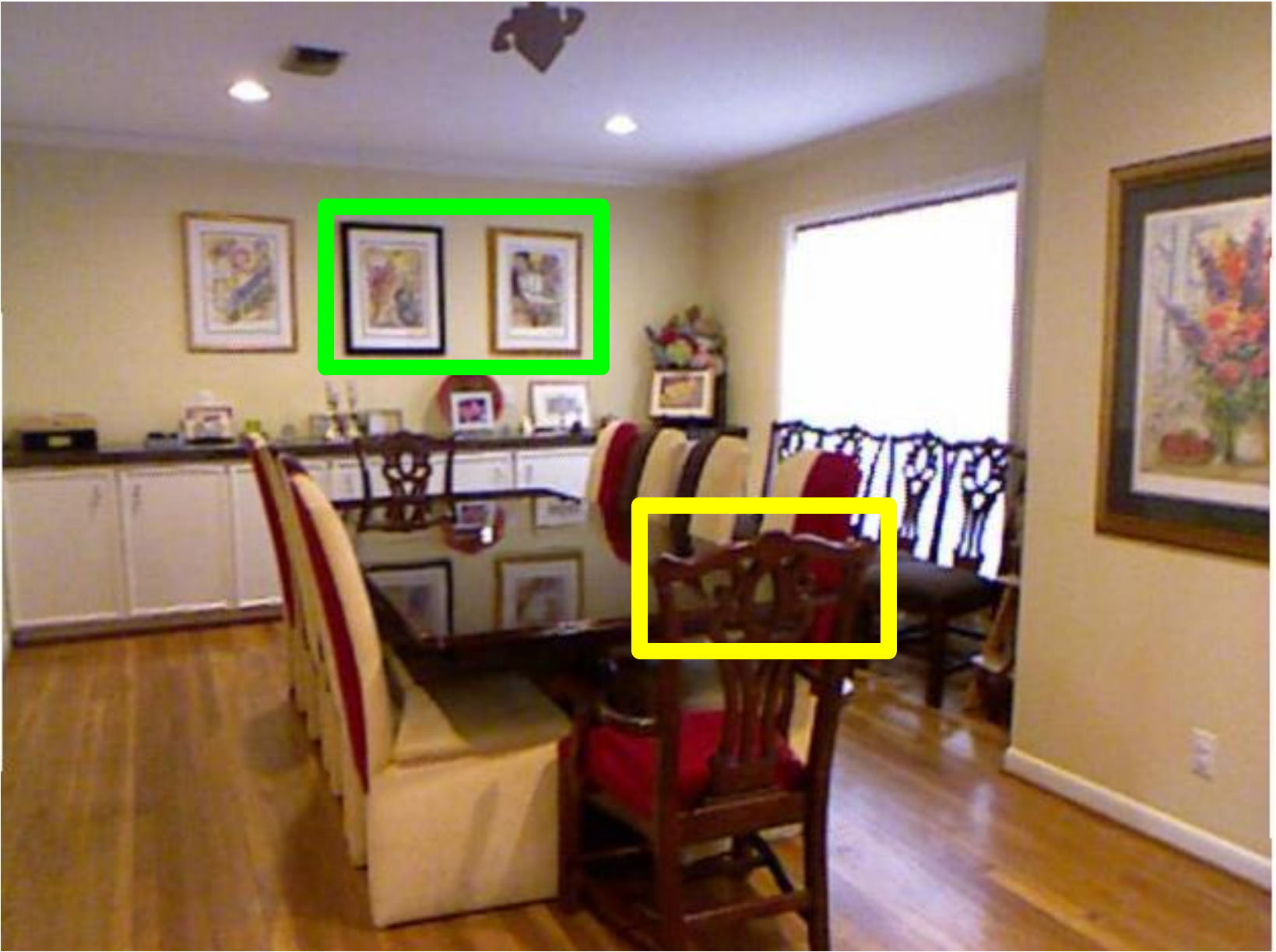} &
\includegraphics[height=.1\linewidth, width = .131\linewidth]{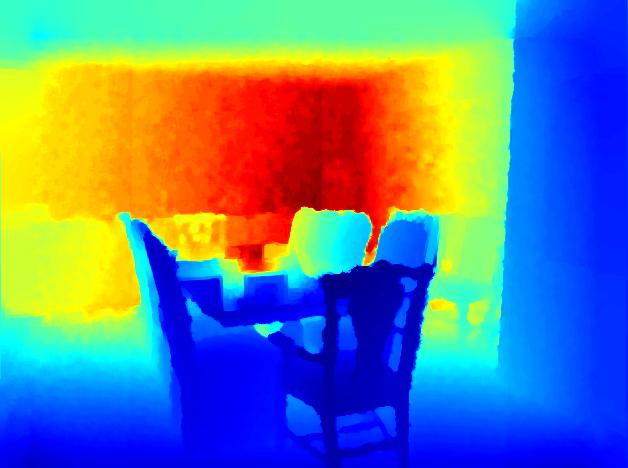} &
\includegraphics[height=.1\linewidth, width = .131\linewidth]{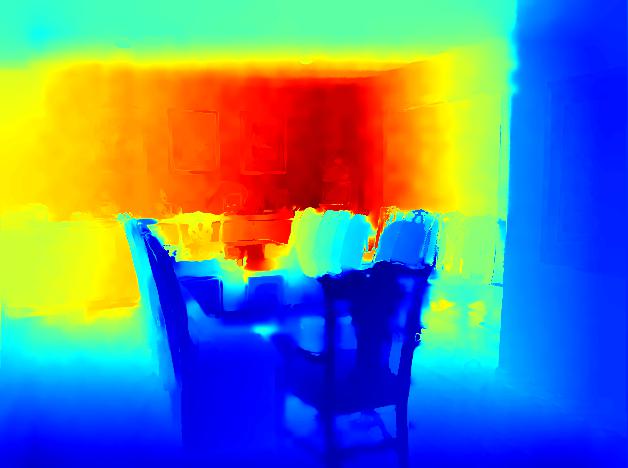} &
\includegraphics[height=.1\linewidth, width = .131\linewidth]{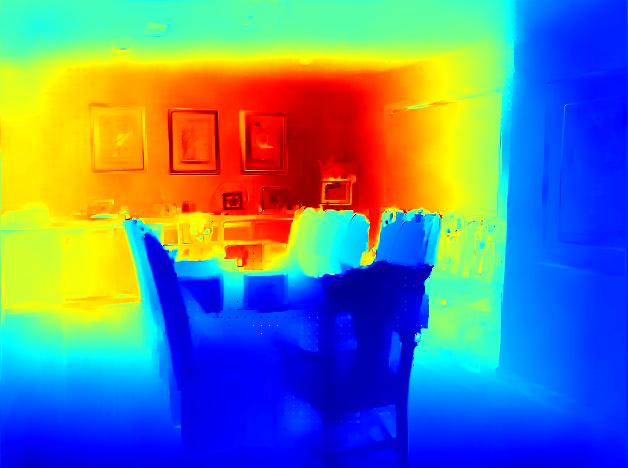} &
\includegraphics[height=.1\linewidth, width = .131\linewidth]{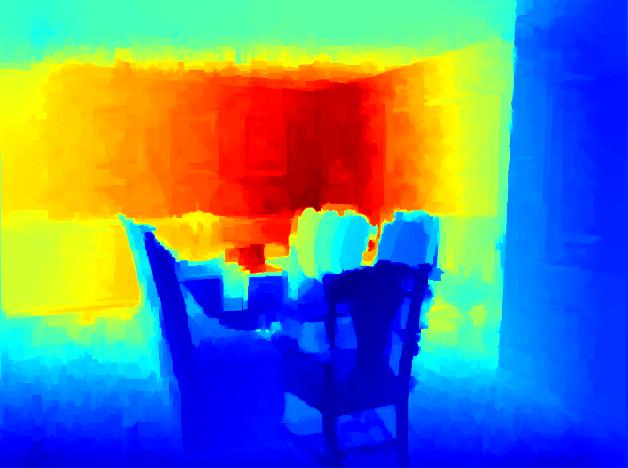} &
\includegraphics[height=.1\linewidth, width = .131\linewidth]{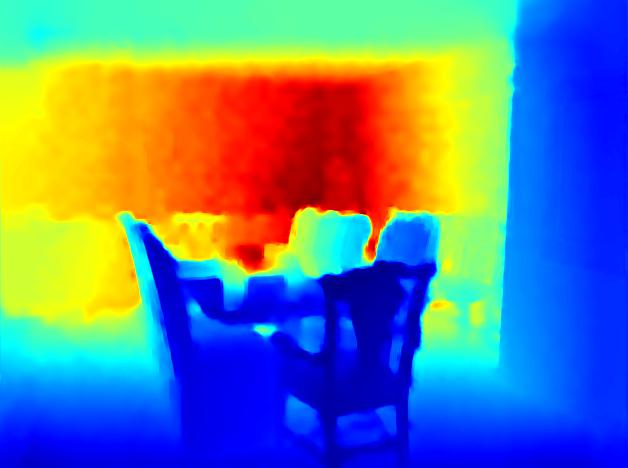} &
\includegraphics[height=.1\linewidth, width = .131\linewidth]{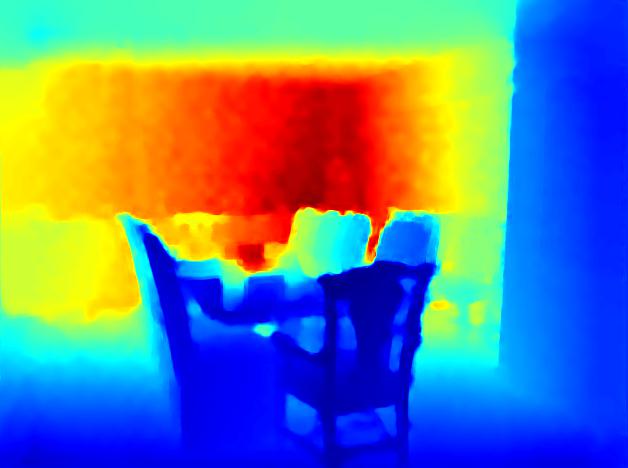} & \\

\includegraphics[height=.05\linewidth, width = .131\linewidth]{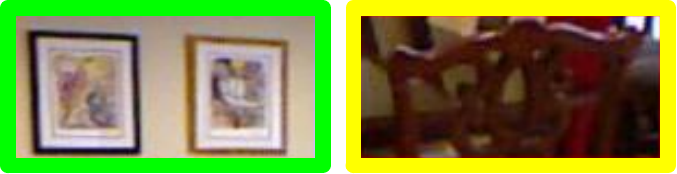} &
\includegraphics[height=.05\linewidth, width = .131\linewidth]{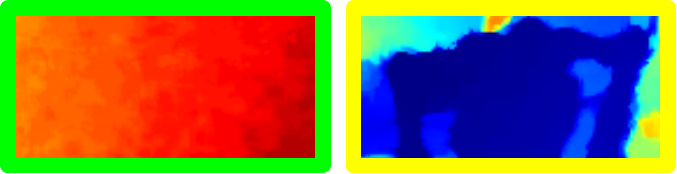} &
\includegraphics[height=.05\linewidth, width = .131\linewidth]{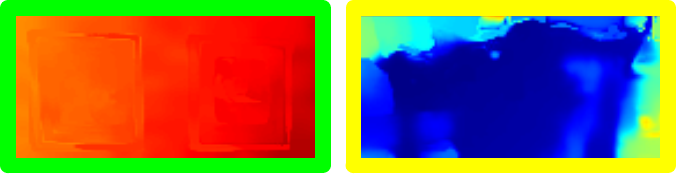} &
\includegraphics[height=.05\linewidth, width = .131\linewidth]{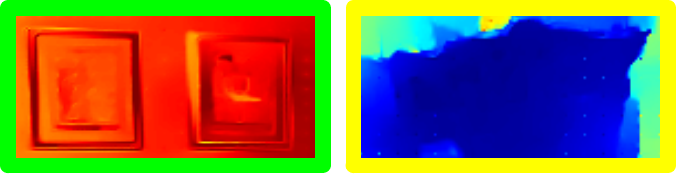} &
\includegraphics[height=.05\linewidth, width = .131\linewidth]{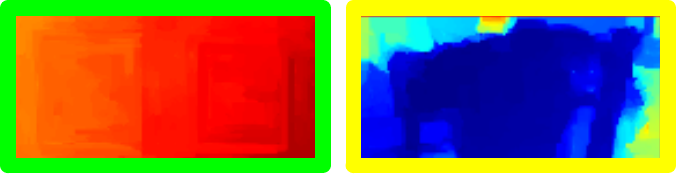} &
\includegraphics[height=.05\linewidth, width = .131\linewidth]{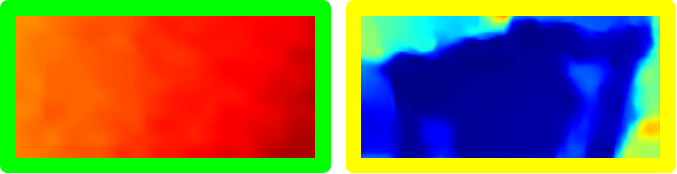} &
\includegraphics[height=.05\linewidth, width = .131\linewidth]{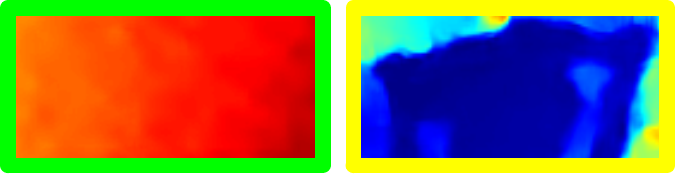} & \\

{ (a) Guidance }& { (b) GT } & { (c) JBU \cite{JBU-TOG-2007}} & { (d) TGV \cite{TGV-ICCV-2013}} & { (e) Park \cite{Park-ICCV-2011}} & { (f) DJF\cite{DJF-ECCV-2016}} & { (g) Ours}\\

{}&{ RMSE}&{ 11.84}&{ 16.05} & { 12.67} & { 7.98} & { \textbf{7.57}} & \\

\includegraphics[height=.1\linewidth, width = .131\linewidth]{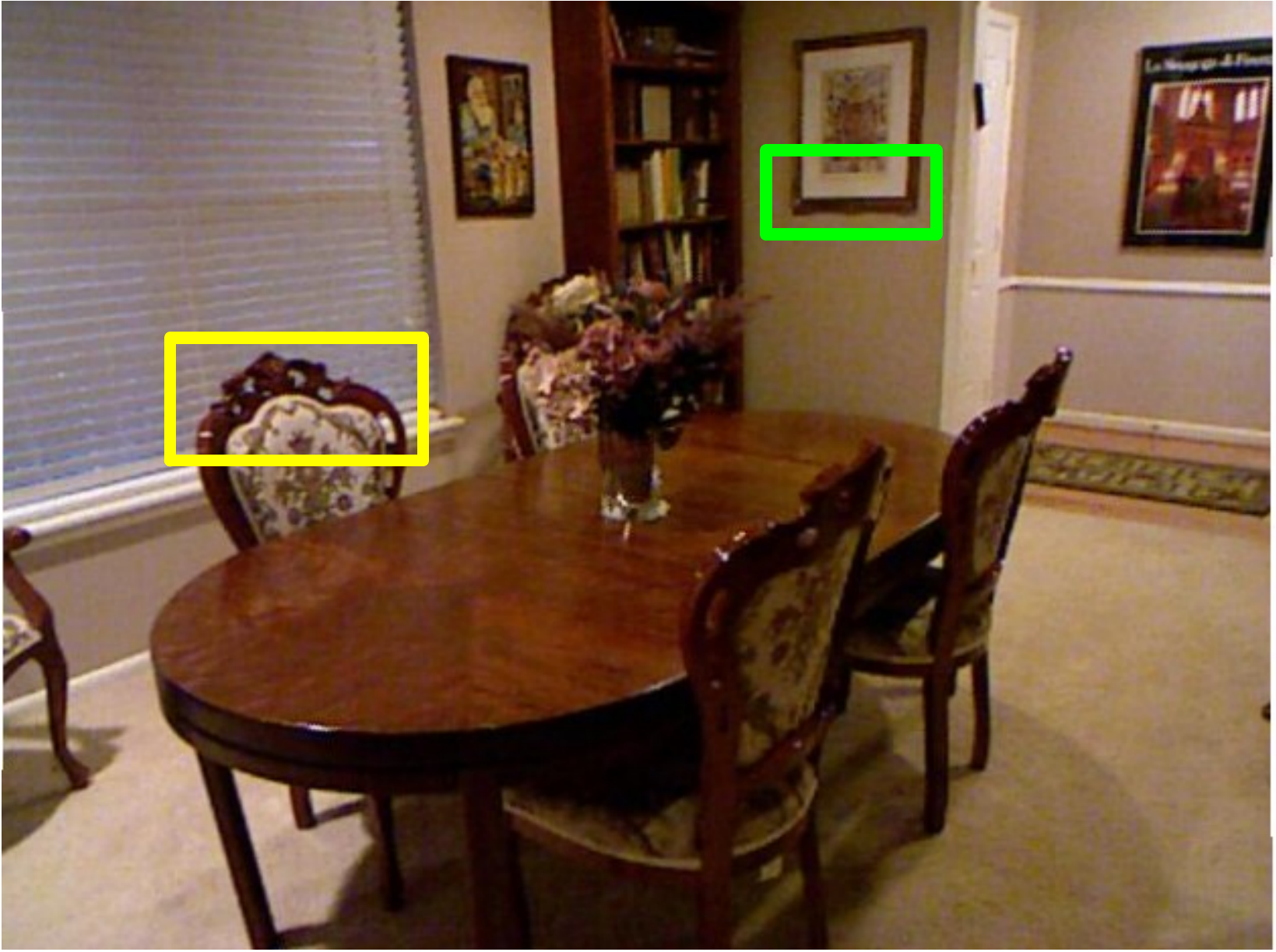} &
\includegraphics[height=.1\linewidth, width = .131\linewidth]{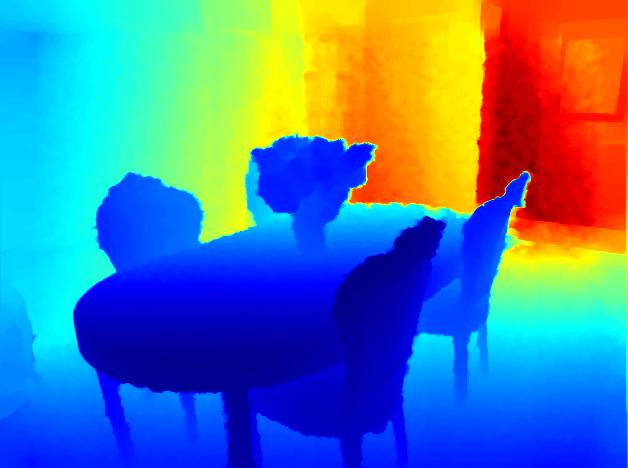} &
\includegraphics[height=.1\linewidth, width = .131\linewidth]{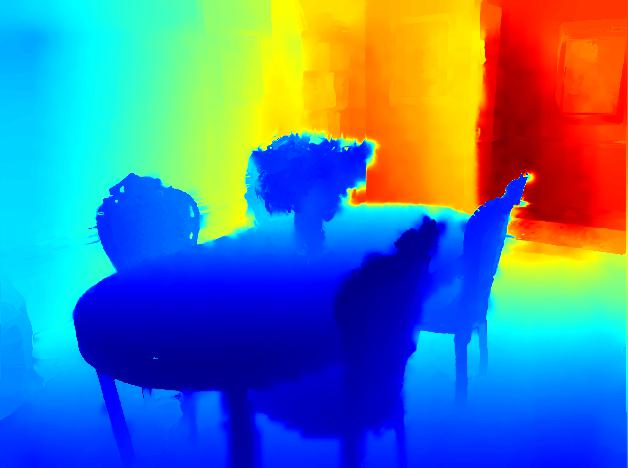} &
\includegraphics[height=.1\linewidth, width = .131\linewidth]{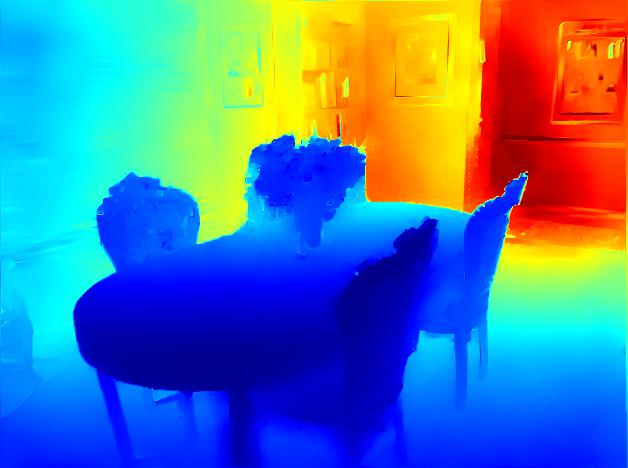} &
\includegraphics[height=.1\linewidth, width = .131\linewidth]{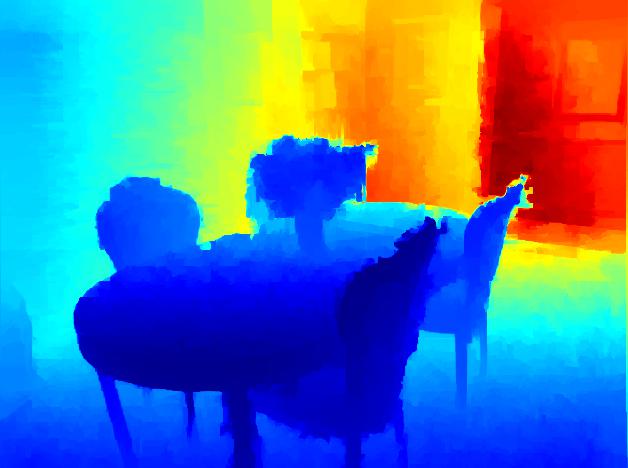} &
\includegraphics[height=.1\linewidth, width = .131\linewidth]{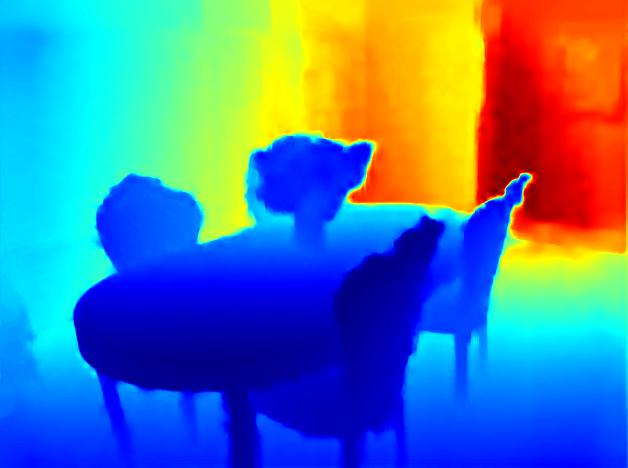} &
\includegraphics[height=.1\linewidth, width = .131\linewidth]{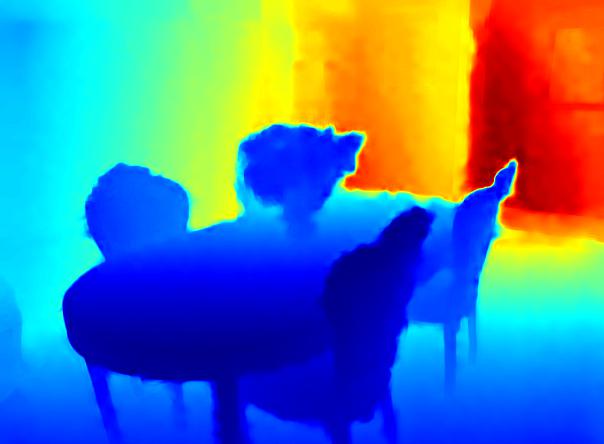} & \\

\includegraphics[height=.05\linewidth, width = .131\linewidth]{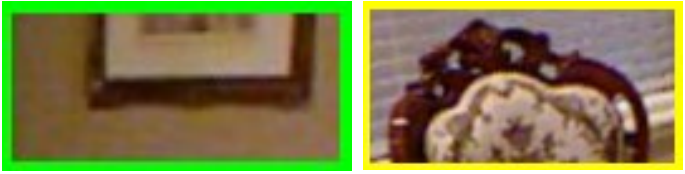} &
\includegraphics[height=.05\linewidth, width = .131\linewidth]{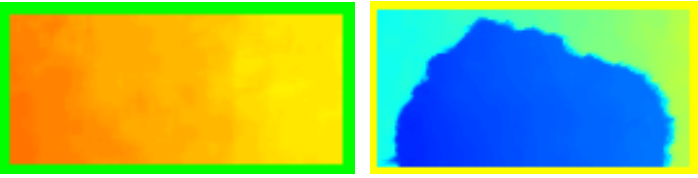} &
\includegraphics[height=.05\linewidth, width = .131\linewidth]{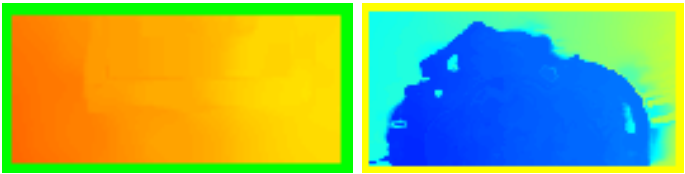} &
\includegraphics[height=.05\linewidth, width = .131\linewidth]{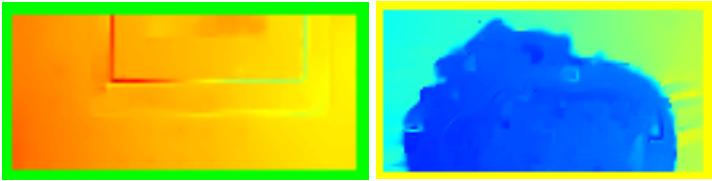} &
\includegraphics[height=.05\linewidth, width = .131\linewidth]{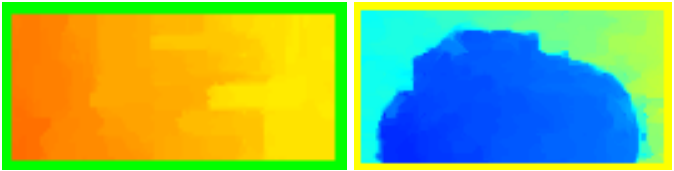} &
\includegraphics[height=.05\linewidth, width = .131\linewidth]{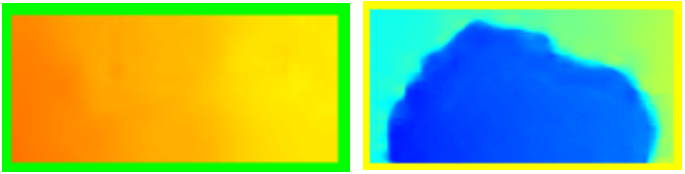} &
\includegraphics[height=.05\linewidth, width = .131\linewidth]{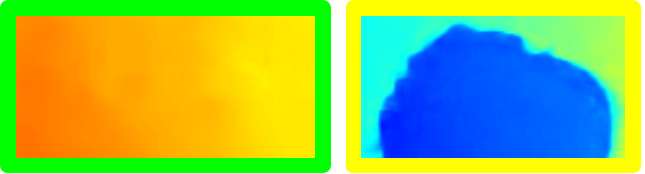} \\

{ (a) Guidance }& { (b) GT } & { (c) JBU \cite{JBU-TOG-2007}} & { (d) TGV \cite{TGV-ICCV-2013}} & { (e) Park \cite{Park-ICCV-2011}} & { (f) DJF\cite{DJF-ECCV-2016}} & { (g) Ours}\\

{}&{ RMSE}&{ 9.47}&{ 16.23} & { 14.18} & { 7.98} & { \textbf{7.74}} & \\

\end{tabular}
\caption{
\textbf{Qualitative comparisons on depth upsampling}. Comparisons against existing depth upsampling algorithms for a scaling factor of 8$\times$. The numbers (in centimeter) are the RMSE metric comparing against the GT in (b).}
\label{fig:Qualitative}
\end{figure*}

\section{Experimental Results}
\label{sec:exp}

In this section, we demonstrate the effectiveness and applicability of our approach through a broad range of joint image filtering tasks, including joint image upsampling, texture-structure separation, and cross-modality image restoration.
%
%MH: I add the following lines.
The source code and datasets will be made available to the public.
More results can be found at \url{http://vllab1.ucmerced.edu/~yli62/DJF_residual/}.

%\vspace{1em}
{\flushleft \textbf{Network training.}}
To train our network, we randomly collect 160,000 training patch pairs of $32\times32$ pixels from 1,000 RGB and depth images in the NYU v2 dataset~\cite{NYU-ECCV-2012}.
Images in the NYU dataset are absolute depth maps captured in complicated indoor scenarios.
We train two models for two different tasks:
(1) joint image upsampling and (2) noise reduction.
For the upsampling task, we obtain each low-quality target image from downsampling the ground-truth image (with scale factors of 4$\times$, 8$\times$, 16$\times$) using
the nearest neighbor interpolation.
For the noise reduction task, we generate the low-quality target image by adding Gaussian noise to each of the ground-truth depth maps with zero mean and variance of 1e-3.
We use the MatConvNet toolbox~\cite{Matconvnet-ACMMM-2015}
to train our joint filters.
%
%We set the learning rate of the first two layers as 1e-3 and the third layer as 1e-4.
%

%\vspace{1em}
{\flushleft \textbf{Testing.}}
Using RGB/depth data for training, our model takes a 1-channel target image (depth map) and a 3-channel guidance image (RGB) as inputs.
However, the trained model can be applied to other data types in addition to RGB/depth
images with simple modifications.
For the multi-channel target images, we apply the trained model independently for each channel.
For the single-channel guidance images, we replicate it three times to create the 3-channel guidance image.

\subsection{Depth map upsampling}

\begin{table*}[t]
%\vspace{-2mm}
\caption{
\textbf{Run-time performance comparisons.}
Average run-time of depth map upsampling algorithms on images of size $640\times480$ pixels.
}
\label{table:time}
\centering
\begin{tabular}{lcccccccccc}
\toprule \\
~&MRF~\cite{MRF-NIPS-2005} & GF~\cite{He-PAMI-2013} & JBU~\cite{JBU-TOG-2007} & TGV~\cite{TGV-ICCV-2013} & Park~\cite{Park-ICCV-2011} &Ham~\cite{Ham-CVPR-2015} & DMSG~\cite{Tai-2016-depth} & FBS~\cite{Barron-2016-solver} & {\revyj{Ours (CPU)}} & Ours (GPU)\\
\midrule
% JB: Use two separate columns to show CPU/GPU
Time (s)&0.76 & 0.08 & 5.64 & 68.21 & 45.79 &8.62 & 0.71&0.34 & \revyj{1.31} & 0.07 \\
\bottomrule
\end{tabular}
\end{table*}

%\vspace{1em}
{\flushleft \textbf{Datasets.}}
We present quantitative performance evaluation on joint depth upsampling using three benchmark datasets where the corresponding high-resolution RGB images are available:
\begin{itemize}
\item Middlebury dataset~\cite{Midd1-CVPR-2007,Midd2-CVPR-2007}: We collect 30 images from 2001-2006 datasets with the missing depth values provided by Lu et al.~\cite{Lu-CVPR-2014}.
%\item Lu~\cite{Lu-CVPR-2014}: This dataset contains six depth maps captured with the ASUS Xtion Pro camera.
\item NYU v2 dataset~\cite{NYU-ECCV-2012}: As we use the 1,000 images in this dataset for training, we use the rest of 449 images for testing.
\item SUN RGB-D~\cite{Song-CVPR-2015}: We use a random subset of 2,000 high-quality RGB/depth image pairs from the 3,784 pairs captured by the Kinect v2 sensor.
These images are captured from a variety of complicated indoor scenes.
\end{itemize}

\revyj{Note that the data in~\cite{NYU-ECCV-2012,Song-CVPR-2015} are \emph{absolute} depth maps representing the physical distances in meters to the observer.
However, the data in~\cite{Midd1-CVPR-2007,Midd2-CVPR-2007} are \emph{relative} depth maps (disparity), which measure the distance between two corresponding points in a scene under two different views.
Each disparity value denotes the number of shifted pixels.
%
%It is able to represent the depth information because the object closer to the observer would appear to jump about their position more than the object further away.
}

\begin{figure*}[t]
\centering
\begin{tabular}{c@{\hspace{0.005\linewidth}}c@{\hspace{0.005\linewidth}}c@{\hspace{0.005\linewidth}}c@{\hspace{0.005\linewidth}}c@{\hspace{0.005\linewidth}}c@{\hspace{0.005\linewidth}}c}

\includegraphics[height = .12\linewidth, width = .15\linewidth]{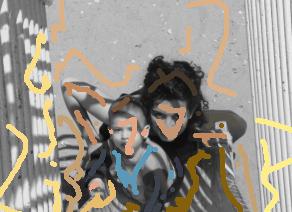} &
\includegraphics[height = .12\linewidth, width = .15\linewidth]{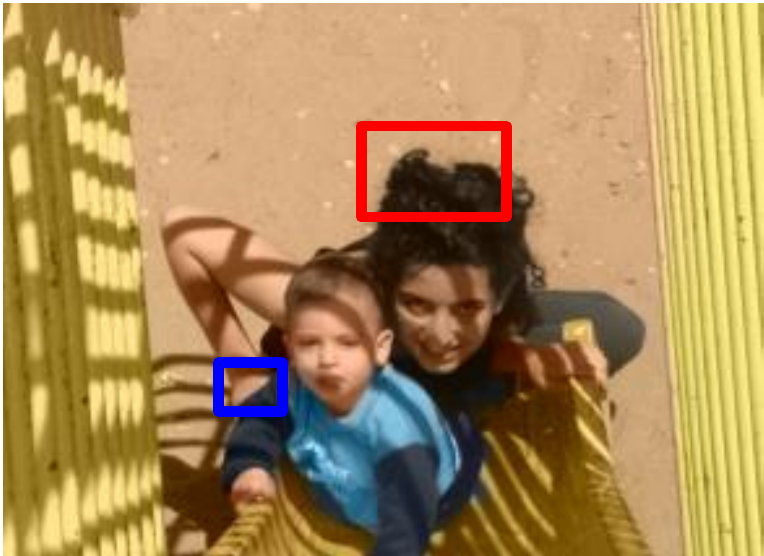} &

\includegraphics[height = .12\linewidth, width = .15\linewidth]{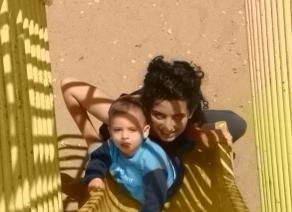} &
\includegraphics[height = .12\linewidth, width = .15\linewidth]{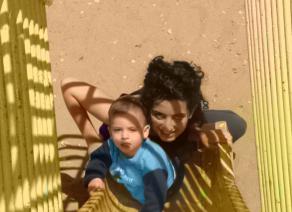} &
\includegraphics[height = .12\linewidth, width = .15\linewidth]{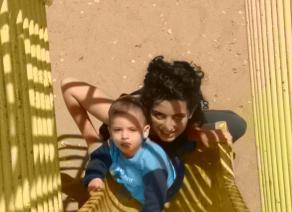} &
\includegraphics[height = .12\linewidth, width = .15\linewidth]{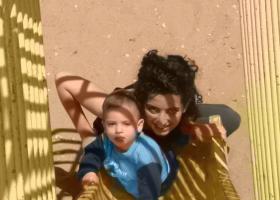} & \\

\includegraphics[height = .08\linewidth, width = .15\linewidth]{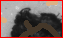} &
\includegraphics[height = .08\linewidth, width = .15\linewidth]{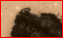} &

\includegraphics[height = .08\linewidth, width = .15\linewidth]{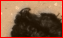} &
\includegraphics[height = .08\linewidth, width = .15\linewidth]{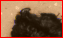} &
\includegraphics[height = .08\linewidth, width = .15\linewidth]{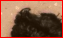} &
\includegraphics[height = .08\linewidth, width = .15\linewidth]{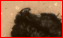} &\\

\includegraphics[height = .08\linewidth, width = .15\linewidth]{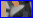} &
\includegraphics[height = .08\linewidth, width = .15\linewidth]{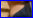} &

\includegraphics[height = .08\linewidth, width = .15\linewidth]{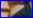} &
\includegraphics[height = .08\linewidth, width = .15\linewidth]{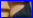} &
\includegraphics[height = .08\linewidth, width = .15\linewidth]{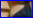} &
\includegraphics[height = .08\linewidth, width = .15\linewidth]{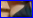} &\\

{(a) Scribbles}& {(b) Levin \cite{Levin-TOG-2004} } & {(c) GF \cite{He-PAMI-2013} }& {(d) Ham \cite{Ham-CVPR-2015} }& { (e) DJF~\cite{DJF-ECCV-2016}} & {(f) Ours} \\

{} & {RMSE} & {5.94} & {6.28}& {5.57} & {\textbf{5.38}} \\

% cat

\includegraphics[width = .15\linewidth]{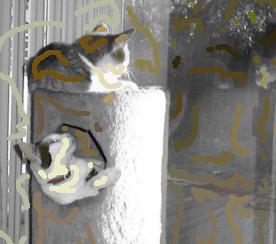} &
\includegraphics[width = .15\linewidth]{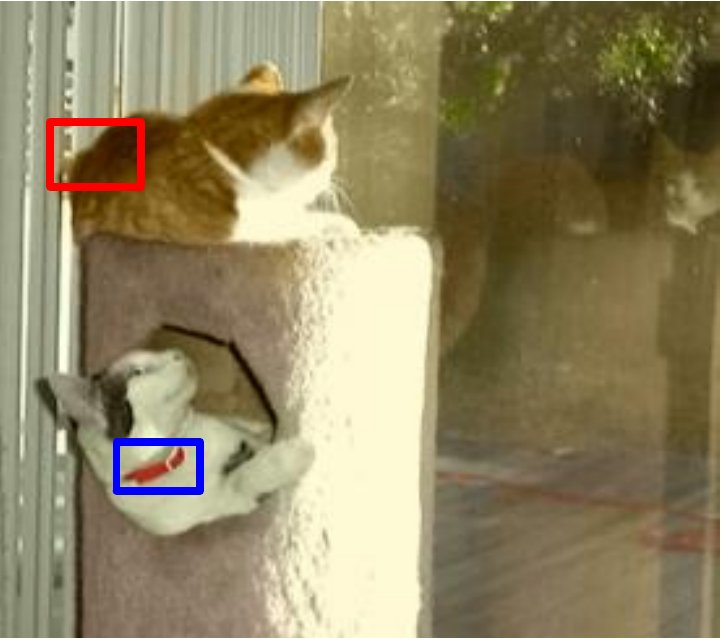} &
\includegraphics[width = .15\linewidth]{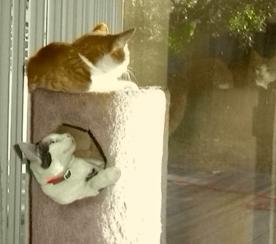} &
\includegraphics[width = .15\linewidth]{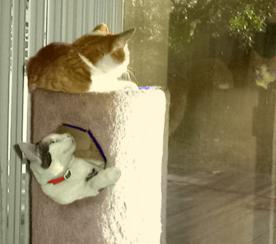} &
\includegraphics[width = .15\linewidth]{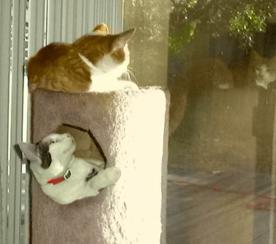} &
\includegraphics[width = .15\linewidth]{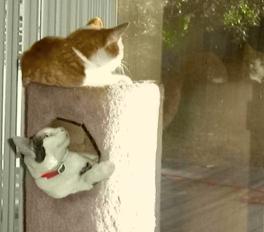} & \\

\includegraphics[width = .15\linewidth]{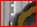} &
\includegraphics[width = .15\linewidth]{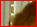} &
\includegraphics[width = .15\linewidth]{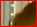} &
\includegraphics[width = .15\linewidth]{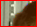} &
\includegraphics[width = .15\linewidth]{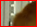} &
\includegraphics[width = .15\linewidth]{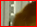} & \\

\includegraphics[width = .15\linewidth]{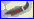} &
\includegraphics[width = .15\linewidth]{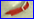} &
\includegraphics[width = .15\linewidth]{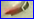} &
\includegraphics[width = .15\linewidth]{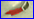} &
\includegraphics[width = .15\linewidth]{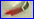} &
\includegraphics[width = .15\linewidth]{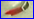} & \\

{(a) Scribbles}& {(b) Levin \cite{Levin-TOG-2004} } & {(c) GF \cite{He-PAMI-2013} }& {(d) Ham \cite{Ham-CVPR-2015} }& { (e) DJF~\cite{DJF-ECCV-2016}} & {(f) Ours} \\

{} & {RMSE} & {5.94} & {6.28}& {5.57} & {\textbf{5.38}} \\

{Time (s)} & {8.20} & {\textbf{1.50}} & {28.8}& {2.80} & {2.82} \\

\end{tabular}

\caption{
\textbf{Colorization upsampling.}
Joint image upsampling applied to colorization.
We also list the runtime for the colorization upsampling process for each method.
The close-up areas show that our joint upsampling results (f) have fewer color bleeding artifacts when compared with other competing algorithms (c-e). Our visual results (f) are comparable with the results computed using the full resolution image in (b).
%
%The numbers in the first row are the running time.
%
The RMSE metric comparing against the GT in (b) are presented. The average RMSE over all test images are shown in Table~\ref{table:quan_color}.
}
\label{fig:colorization}
\end{figure*}

%\vspace{1em}
{\flushleft \textbf{Evaluated methods.}}
We compare our model against several state-of-the-art joint image filters for depth map upsampling.
The JBU~\cite{JBU-TOG-2007}, GF~\cite{He-PAMI-2013}, Ham~\cite{Ham-CVPR-2015} and FBS~\cite{Barron-2016-solver} methods are generic joint image upsampling.
On the other hand, the MRF~\cite{MRF-NIPS-2005}, TGV~\cite{TGV-ICCV-2013}, Park~\cite{Park-ICCV-2011} and DMSG~\cite{Tai-2016-depth}, algorithms are
designed specifically for image-guided depth upsampling.
Using the experimental protocols for evaluating the joint depth upsampling algorithms~\cite{Park-ICCV-2011,TGV-ICCV-2013,Ham-CVPR-2015}, we obtain the low-resolution target image from the ground-truth depth map using the nearest-neighbor downsampling method.

{\flushleft \textbf{Quantitative comparisons.}}
Table~\ref{table:depth} shows the quantitative results in terms of the root mean squared errors (RMSE).
For other methods, we use the default parameters in the original implementations.
\revyj{The proposed algorithm performs well against 
the state-of-the-art methods across all three datasets.
The extensive evaluations on absolute depth datasets~\cite{NYU-ECCV-2012,Song-CVPR-2015} demonstrate the effectiveness of our algorithm in handling 
complicated real-world indoor scenes.
}
Furthermore, we compare the average run-time of different methods on the NYU v2 dataset in Table~\ref{table:time}.
We carry out all the experiments on the same machine with an Intel i7 3.6GHz CPU and 16GB RAM.
\revyj{We report the running time of our model in either CPU or GPU mode (GTX 745).}
%
% Since our deep learning model is easy to run on the GPU (GTX 745), we also report its faster run-time.}
%
Among all the evaluated methods, the proposed algorithm is efficient while delivering high-quality upsampling results.

The concurrent DMSG method by Tai et al.~\cite{Tai-2016-depth} outperforms
the proposed algorithm on the Middlebury dataset.
This can be attributed to several reasons.
First, Tai et al.~\cite{Tai-2016-depth} leverage multi-scale guidance data while we
use only single scale signals.
The multi-scale design requires more network parameters to learn.
For example, the model size of the upsampling model (8$\times$) in~\cite{Tai-2016-depth} is 1,822 KB compared to our model size of 526 KB.
Second, the model in~\cite{Tai-2016-depth} is trained on a small collection of relative depth maps (82 images)~\cite{Midd1-CVPR-2007,Midd2-CVPR-2007}.
In contrast, our model is trained on a large dataset (1000 images) of absolute depth maps~\cite{NYU-ECCV-2012}.
\revyjnew{For fair comparisons using absolute depth maps, we re-train the model of~\cite{Tai-2016-depth} with the same dataset~\cite{NYU-ECCV-2012} based on our own implementation.}
%When evaluating the model of~\cite{Tai-2016-depth} on absolute depth datasets~\cite{NYU-ECCV-2012,Song-CVPR-2015}, we find that it does not generalize well due to different characteristics of absolute and relative depth maps.
%
\revyjnew{Table~\ref{table:depth}
shows that the performance of both~\cite{Tai-2016-depth} and our previous work DJF~\cite{DJF-ECCV-2016} on absolute depth datasets~\cite{NYU-ECCV-2012,Song-CVPR-2015} achieve similar performance. While the method in~\cite{Tai-2016-depth} also uses the similar strategy of predicting residuals, we demonstrate that the proposed algorithm achieves improved results with fewer parameters, suggesting the practical applicability of our model to real-world applications.}
Another important difference is that the model in~\cite{Tai-2016-depth} is designed only for depth upsampling.
Our approach, on the other hand, can be applied to generic joint image filtering tasks.

\begin{figure*}[t]
\centering

\begin{tabular}{c@{\hspace{0.005\linewidth}}c@{\hspace{0.005\linewidth}}c@{\hspace{0.005\linewidth}}c@{\hspace{0.005\linewidth}}c@{\hspace{0.005\linewidth}}c@{\hspace{0.005\linewidth}}c}

\includegraphics[width = .185\linewidth]{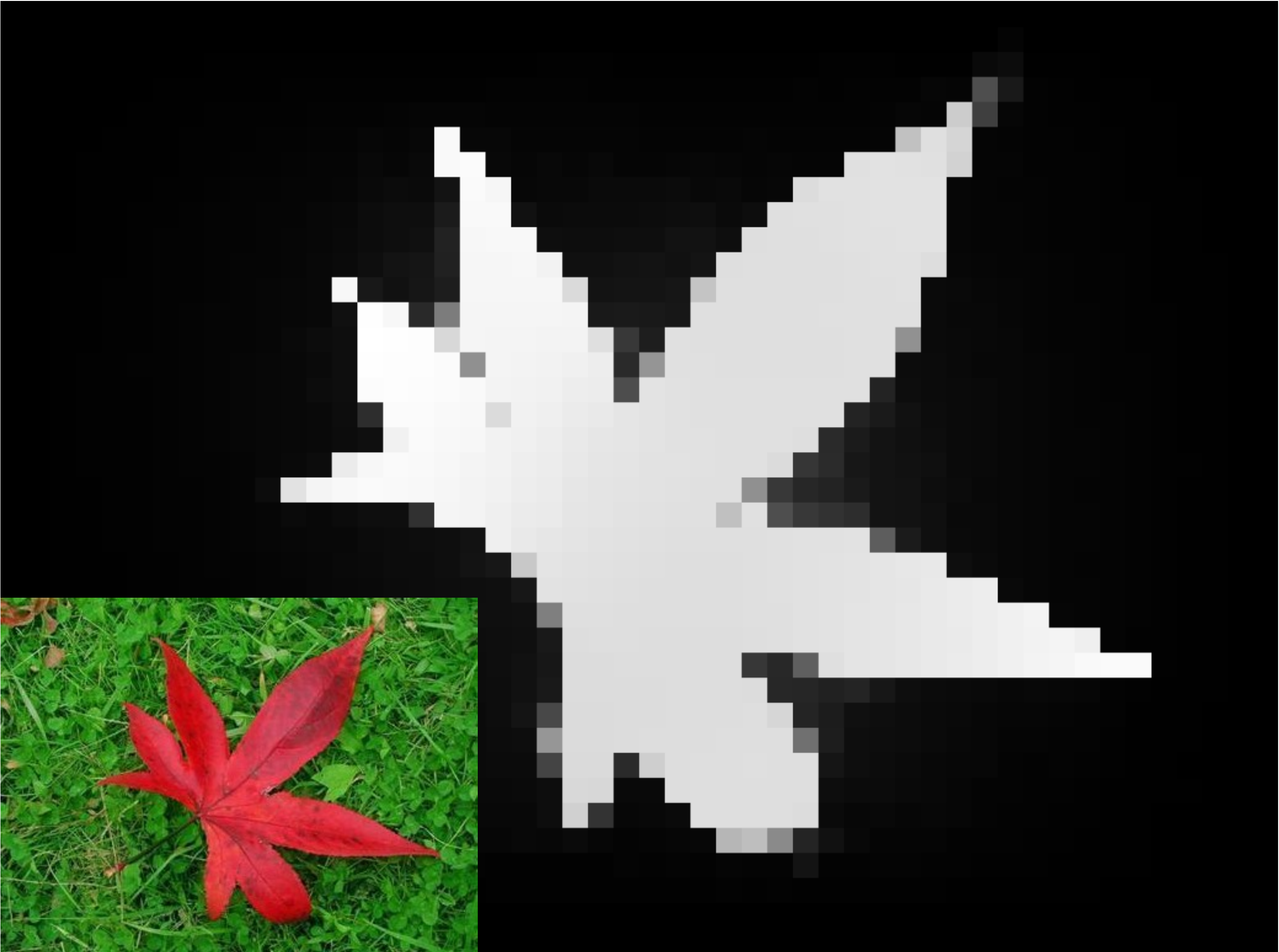} &

\includegraphics[width = .185\linewidth]{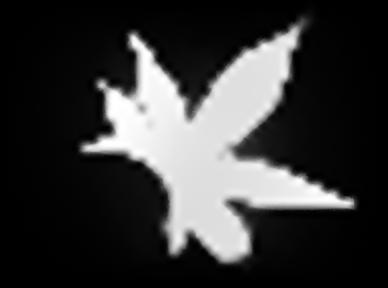} &

\includegraphics[width = .185\linewidth]{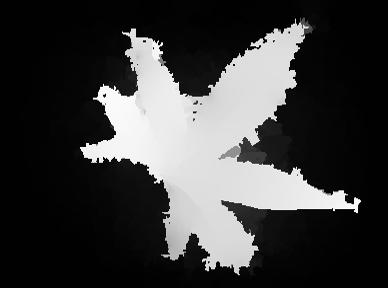} &
\includegraphics[width = .185\linewidth]{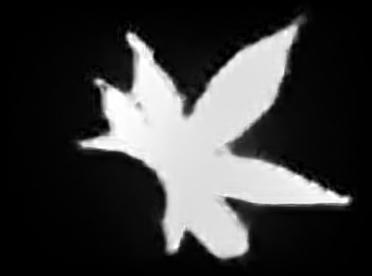} &
\includegraphics[width = .185\linewidth]{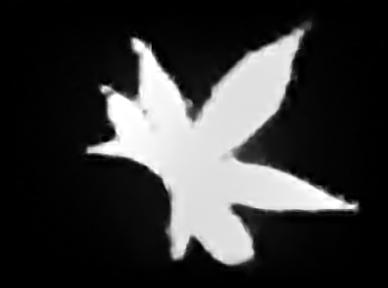} &\\

{(a) Low-res saliency~\cite{Manifold-2013-saliency}} & {(b) GF~\cite{He-PAMI-2013}} & {(c) Ham~\cite{Ham-CVPR-2015}} & { (d) DJF~\cite{DJF-ECCV-2016}} & {(e) Ours} \\
{F-measure}&{0.722}&{0.716}&{0.737}&{\textbf{0.748}} \\

\includegraphics[width = .185\linewidth]{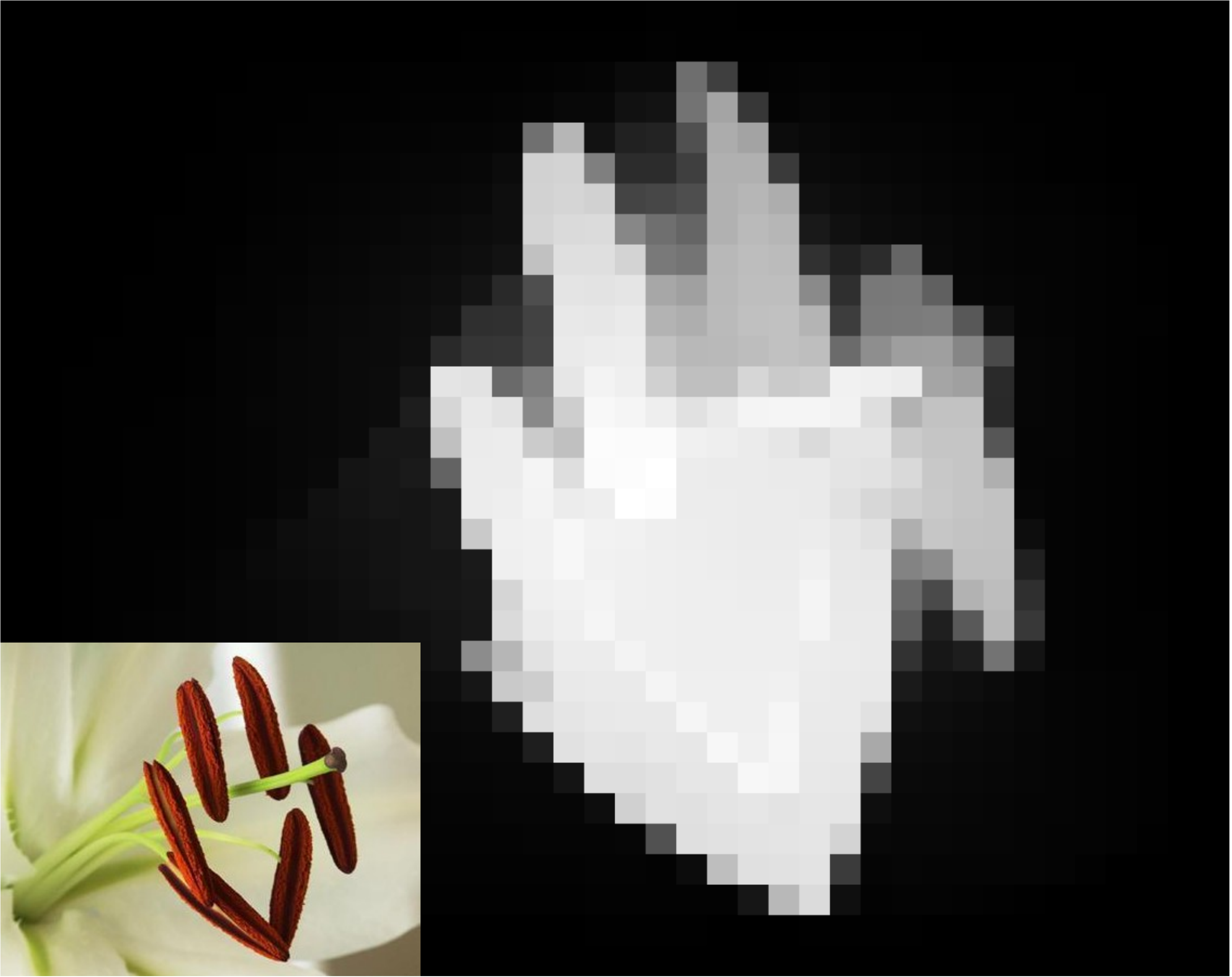} &

\includegraphics[width = .185\linewidth]{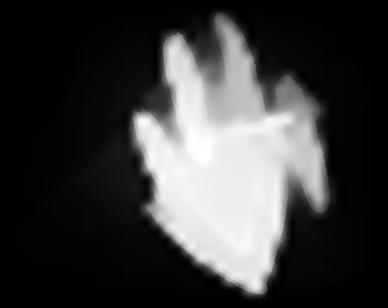} &

\includegraphics[width = .185\linewidth]{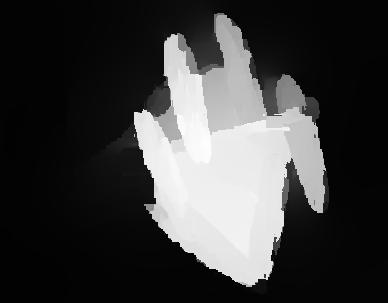} &

\includegraphics[width = .185\linewidth]{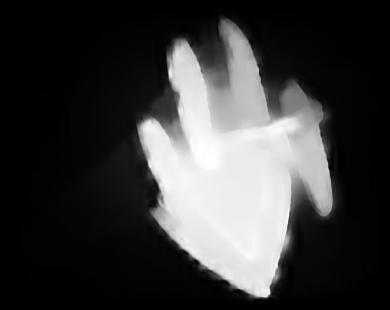} &
\includegraphics[width = .185\linewidth]{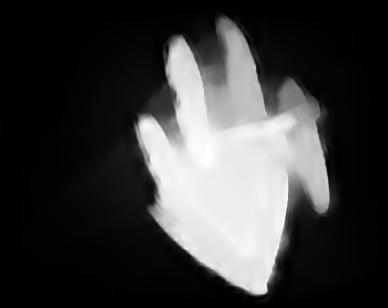} &\\

{(a) Low-res saliency~\cite{Manifold-2013-saliency}} & {(b) GF~\cite{He-PAMI-2013}} & {(c) Ham~\cite{Ham-CVPR-2015}} & { (d) DJF~\cite{DJF-ECCV-2016}} & {(e) Ours} \\
{F-measure}&{0.758}&{0.747}&{0.772}&{\textbf{0.779}} \\
\end{tabular}

\caption{
\textbf{Saliency map upsampling.}
Visual comparisons of saliency map upsampling results (10$\times$). (a) Low-res saliency map obtained from the downsampled RGB image (inset: guidance image). The numbers are the F-measure metric comparing against the GT. The average F-measure over all test images are shown in Table~\ref{table:quan_color}.
}
\label{fig:saliency}
\end{figure*}

\begin{figure*}[t!]
\centering
\begin{tabular}{c@{\hspace{0.005\linewidth}}c@{\hspace{0.005\linewidth}}c@{\hspace{0.005\linewidth}}c@{\hspace{0.005\linewidth}}c@{\hspace{0.005\linewidth}}c@{\hspace{0.005\linewidth}}c@{\hspace{0.005\linewidth}}c@{\hspace{0.005\linewidth}}c@{\hspace{0.005\linewidth}}c@{\hspace{0.005\linewidth}}c@{\hspace{0.005\linewidth}}c@{\hspace{0.005\linewidth}}c}

\includegraphics[width = .15\linewidth, height=.087\linewidth]{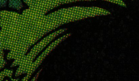} &
\includegraphics[width = .15\linewidth, height=.087\linewidth]{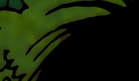} &
\includegraphics[width = .15\linewidth, height=.087\linewidth]{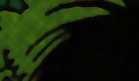} &

\includegraphics[width = .15\linewidth, height=.087\linewidth]{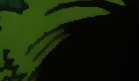} &

\includegraphics[width = .15\linewidth, height=.087\linewidth]{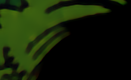} & 
\includegraphics[width = .15\linewidth, height=.087\linewidth]{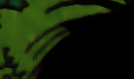}  \\

{(a) Input } & {(b)  Kopf~\cite{Kopf-TOG-2012} } & {(c) RGF~\cite{Rolling-ECCV-2014}} & {(d) Xu~\cite{Xu-TOG-2012}} &{(e) DJF~\cite{DJF-ECCV-2016} } & {(f) Ours}\\

{} & {RMSE} & {7.14} & {7.30} & {7.07}  & {\textbf{6.91}}\\

\includegraphics[width = .15\linewidth, height=.07\linewidth]{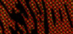} &
\includegraphics[width = .15\linewidth, height=.07\linewidth]{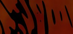} &

\includegraphics[width = .15\linewidth, height=.07\linewidth]{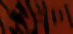} &

\includegraphics[width = .15\linewidth, height=.07\linewidth]{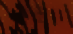} &

\includegraphics[width = .15\linewidth, height=.07\linewidth]{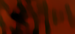} &
\includegraphics[width = .15\linewidth, height=.07\linewidth]{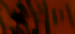}  \\

{(a) Input } & {(b)  Kopf~\cite{Kopf-TOG-2012} } & {(c) RGF~\cite{Rolling-ECCV-2014}} & {(d) Xu~\cite{Xu-TOG-2012}} &{(e) DJF~\cite{DJF-ECCV-2016} } & {(f) Ours}\\

{} & {RMSE} & {16.88} & {\textbf{16.62}} & {17.86}  & {17.24}\\

\end{tabular}
\caption{\revyjnew{\textbf{Inverse halftoning}. For each method, we carefully select the parameter for the optimal results. (c) $\sigma_s=2, \sigma_r=0.05, iter=4$. (d) $\lambda=0.005, \sigma=1$. \revyjnewminor{(e)-(f) top: $iter=2$, bottom: $iter=3$.} Since there exists no GT result, we regard the result of~\cite{Kopf-TOG-2012} in (b) as the GT because it is an algorithm specifically designed for reconstructing halftoned images. The numbers are the RMSE metric comparing against the result in (b). }
}
\label{fig:half1}
\end{figure*}

{\flushleft \textbf{Effects of skip connection.}}
We validate the contribution of the introduced skip connection by comparing
the DJF~\cite{DJF-ECCV-2016} method and proposed algorithm (bottom two rows of Table~\ref{table:depth}).
In Section~\ref{sec:dis}, we show that it is difficult to gain further improvement
by simply modifying network parameters, such as the filter size, filter number, and network depth.
However, with the skip connection, the proposed algorithm obtains
significant performance improvement.
The performance gain can be explained by that using skip connection
alleviates the issues that the network only learns the appearance of the target input images,
and helps the network focus on learning the residuals instead.

{\flushleft \textbf{Effects of training modality.}}
To validate the effect of training with different modalities, we
compare our model with a variant that is trained with RGB/flow data (denoted as Ours-flow).
We randomly select 1,000 RGB/flow image pairs from the Sintel dataset~\cite{Sintel-ECCV-2012} and collect 80,000 training patch pairs of 32$\times$32 pixels.
We use either x-component or y-component of the optical flow as our target image.
During the testing phase, we apply the trained model independently for each channel of the target image.
Although the model Ours-flow is trained with the RGB/flow data for optical flow upsampling, Ours-flow performs favorably on the task of depth upsampling against our final model (Ours) trained with the RGB/depth data, as shown in Table~\ref{table:depth}.

{\flushleft \textbf{Visual comparisons.}}
We show four examples for qualitative comparisons in Figure~\ref{fig:Qualitative}.
It is worth noticing that the proposed joint filter selectively transfers salient structures in the guidance image while avoiding texture-copying artifacts (see the green boxes).
The GF~\cite{He-PAMI-2013} method does not recover the degraded boundary well under a large upsampling factor (e.g., 8$\times$).
The JBU~\cite{JBU-TOG-2007}, TGV~\cite{TGV-ICCV-2013} and Park~\cite{Park-ICCV-2011} approaches are agnostic to structural consistency between the target and guidance images, and thus transfer erroneous details.
In contrast, the results of our algorithm are smoother, sharper and more accurate with respect to the ground truth.

\begin{table}[t]
%\vspace{-2mm}
\caption{Quantitative comparisons of different upsampling methods on difference solution maps.}
%\vspace{-1mm}
\label{table:quan_color}
\centering
\begin{tabular}{lccccc}
\toprule
~& ~Bicubic~ & ~GF \cite{He-PAMI-2013}~ &~Ham \cite{Ham-CVPR-2015}~~&~DJF~\cite{DJF-ECCV-2016}~~&~Ours~\\
\midrule
RMSE & 6.01 & 5.74 & 6.31 & 5.48 & 5.40 \\
F-measure & 0.759 & 0.766 & 0.763 & 0.778 & 0.781 \\
\bottomrule
\end{tabular}
%\vspace{-2mm}
\end{table}

\subsection{Joint image upsampling}
Numerous computational photography applications require obtaining
a solution map (e.g., chromaticity, saliency, disparity, labels) over the pixel grid.
However, it is often time-consuming or memory-intensive to compute
the high-resolution solution maps directly.
An alternative is to first obtain a low-res solution map over the downsampled pixel grids
and then upsample the low-resolution solution map back to the original resolution with a joint image upsampling algorithm.
Such a pipeline requires the upsampling method to restore well
image degradation caused by downsampling and avoid the inconsistency issues.
In what follows, we demonstrate the use of the learned joint image filters for colorization and saliency as examples.
\revyj{Note that in the following applications we use the \emph{same} model trained with RGB/depth data and evaluate on other image modalities without retraining the network using data in the new domains. % We do not retrain the network with any new data in the following applications.
}

\begin{figure*}[t]
\centering

\begin{tabular}{c@{\hspace{0.005\linewidth}}c@{\hspace{0.005\linewidth}}c@{\hspace{0.005\linewidth}}c@{\hspace{0.005\linewidth}}c@{\hspace{0.005\linewidth}}c@{\hspace{0.005\linewidth}}c}
\includegraphics[height = .16\linewidth, width = .24\linewidth]{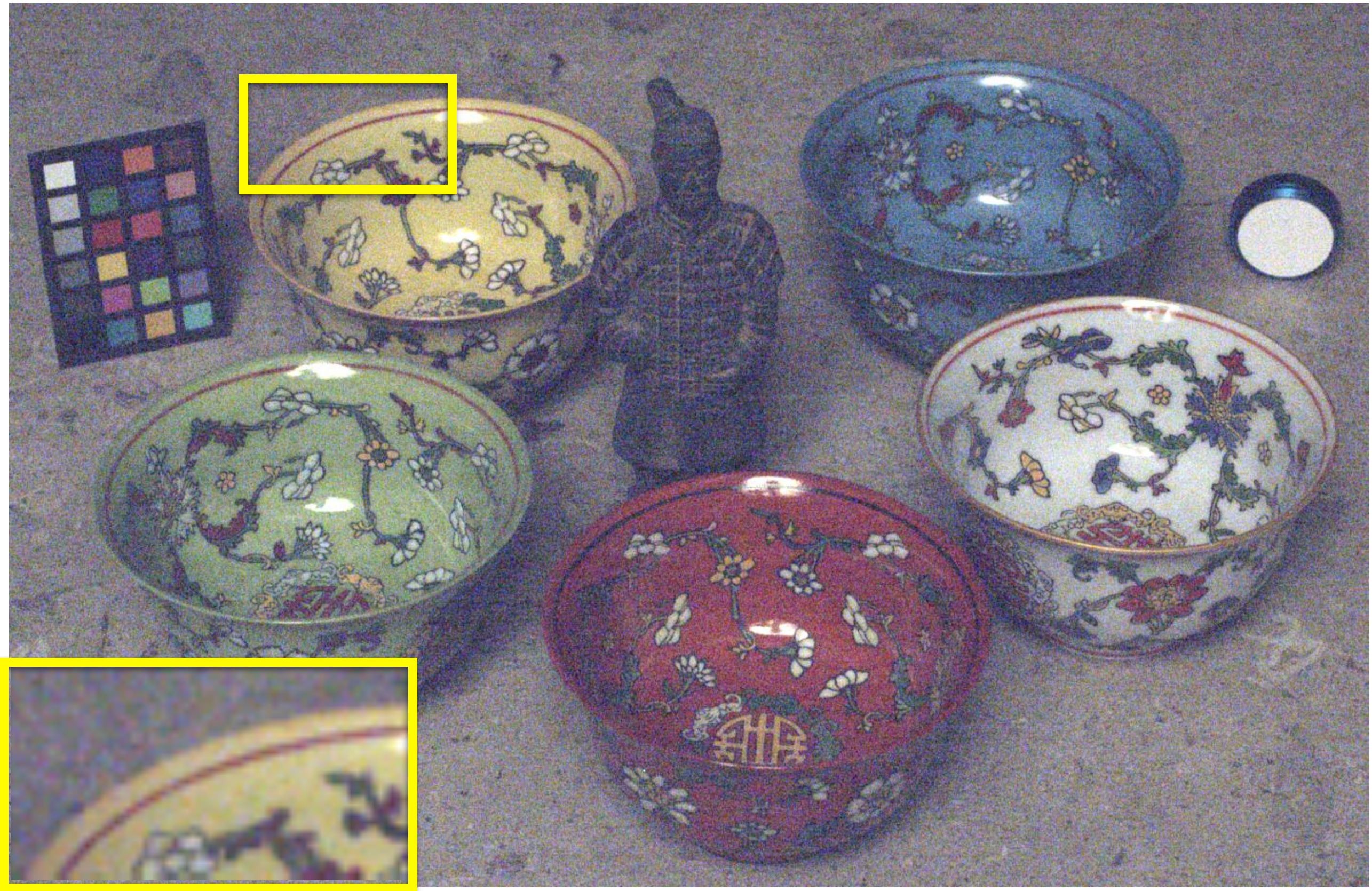} &
\includegraphics[height = .16\linewidth, width = .24\linewidth]{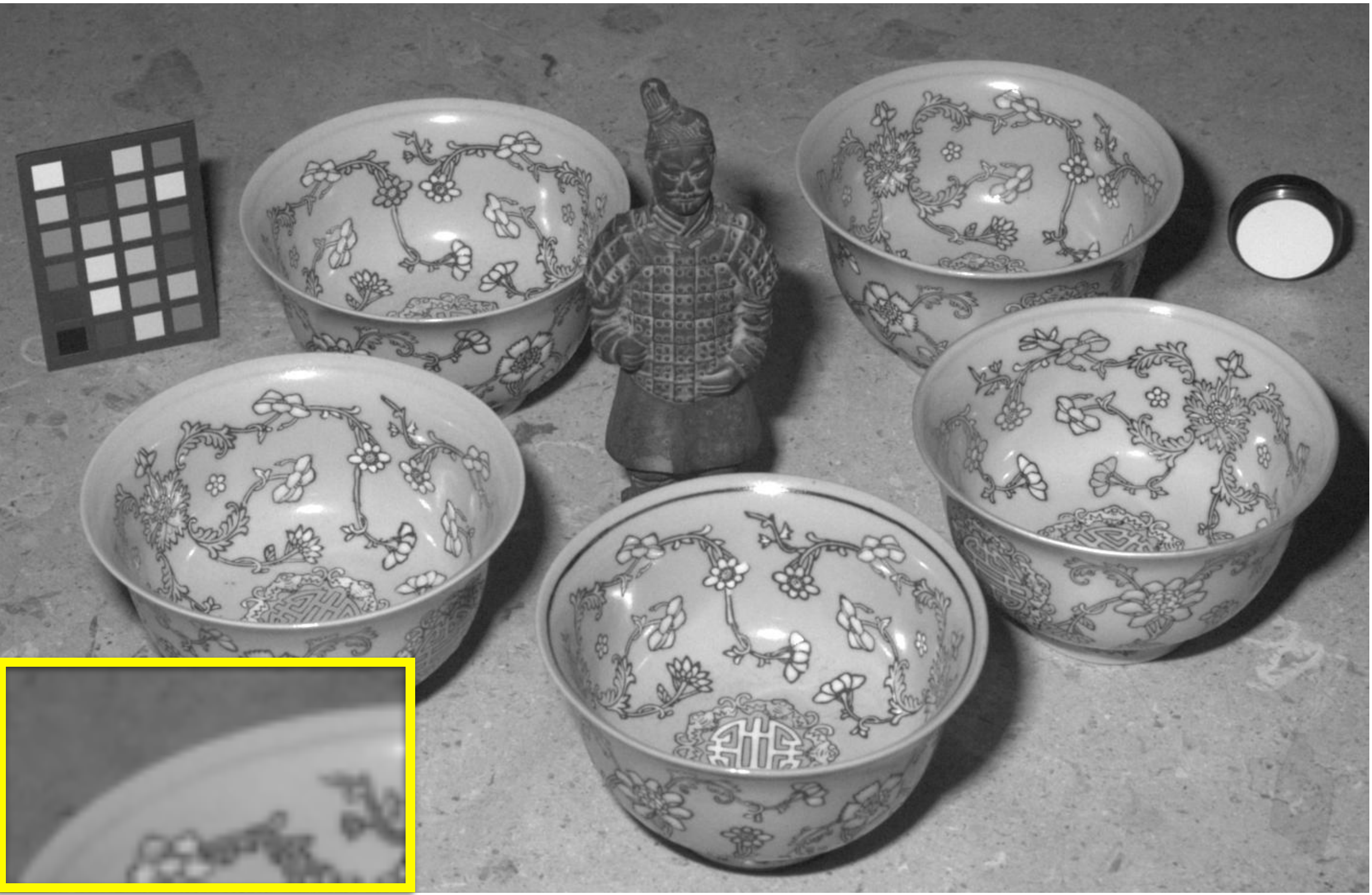} & 

\hspace{1pt}\vrule\hspace{1pt}

\includegraphics[height = .16\linewidth, width = .24\linewidth]{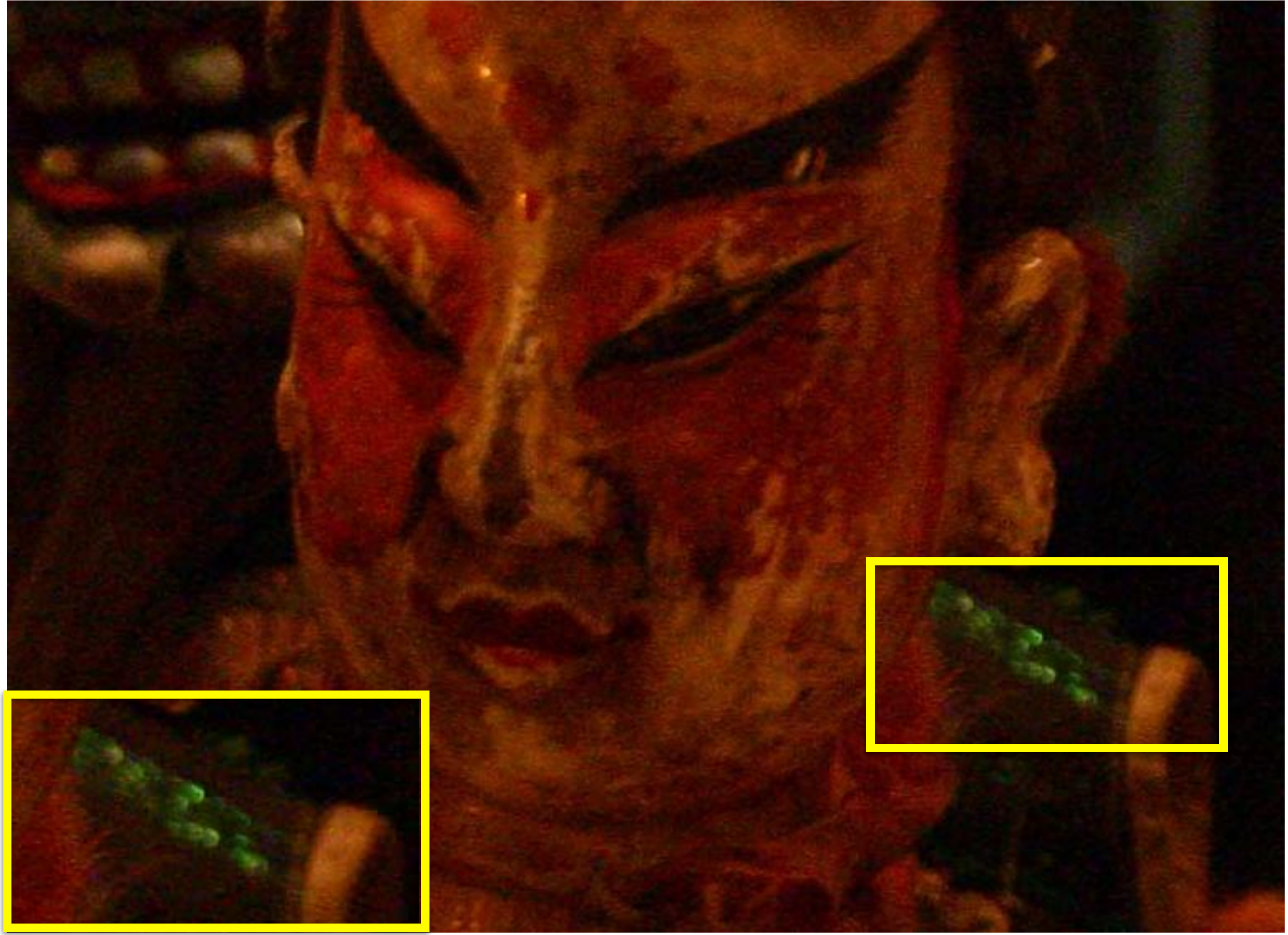} &
\includegraphics[height = .16\linewidth, width = .24\linewidth]{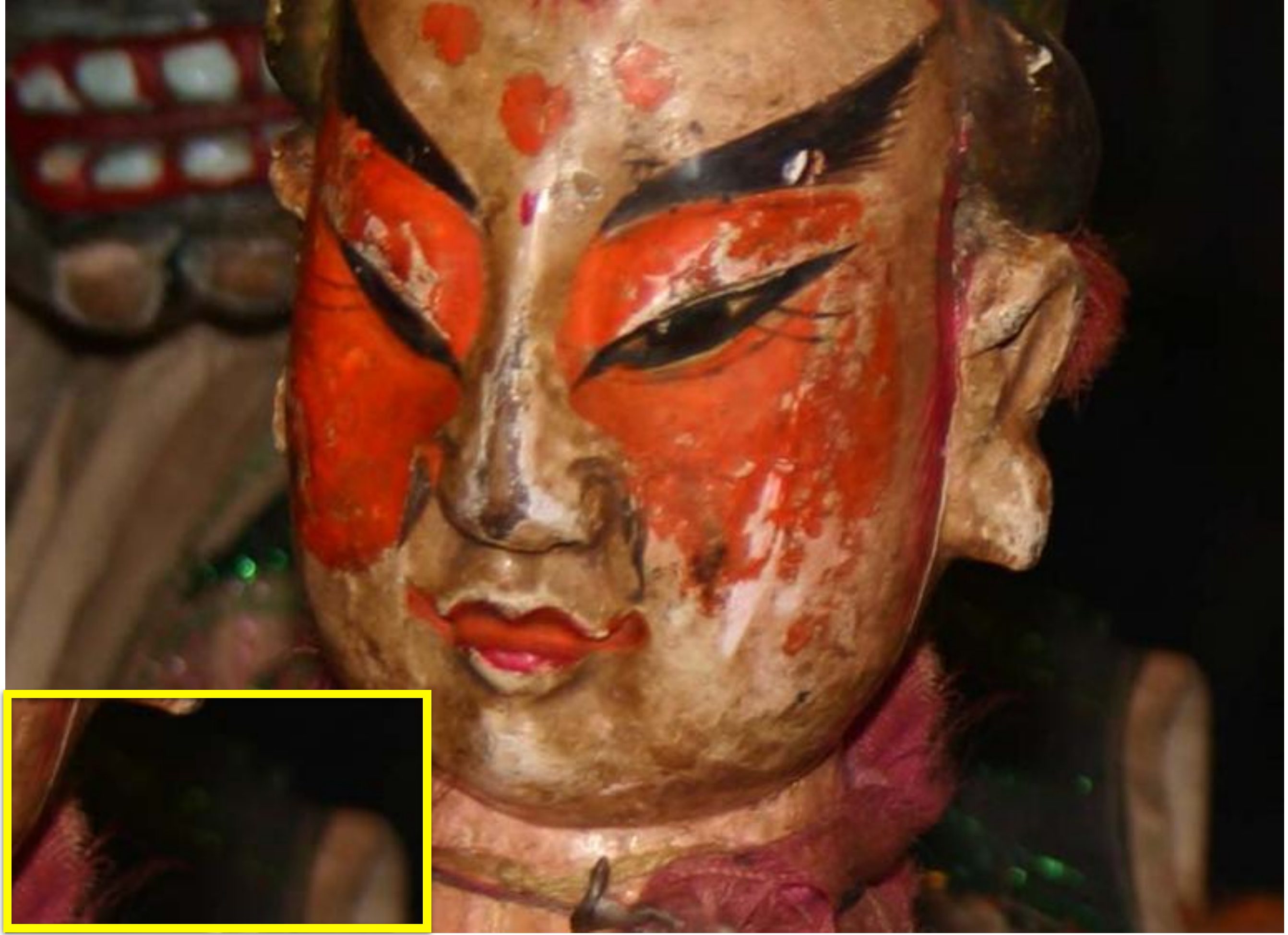} & \\

{Noisy RGB}& {Guided NIR} & {Noisy Non-Flash}& {Guided Flash}  \\

\includegraphics[height = .16\linewidth, width = .24\linewidth]{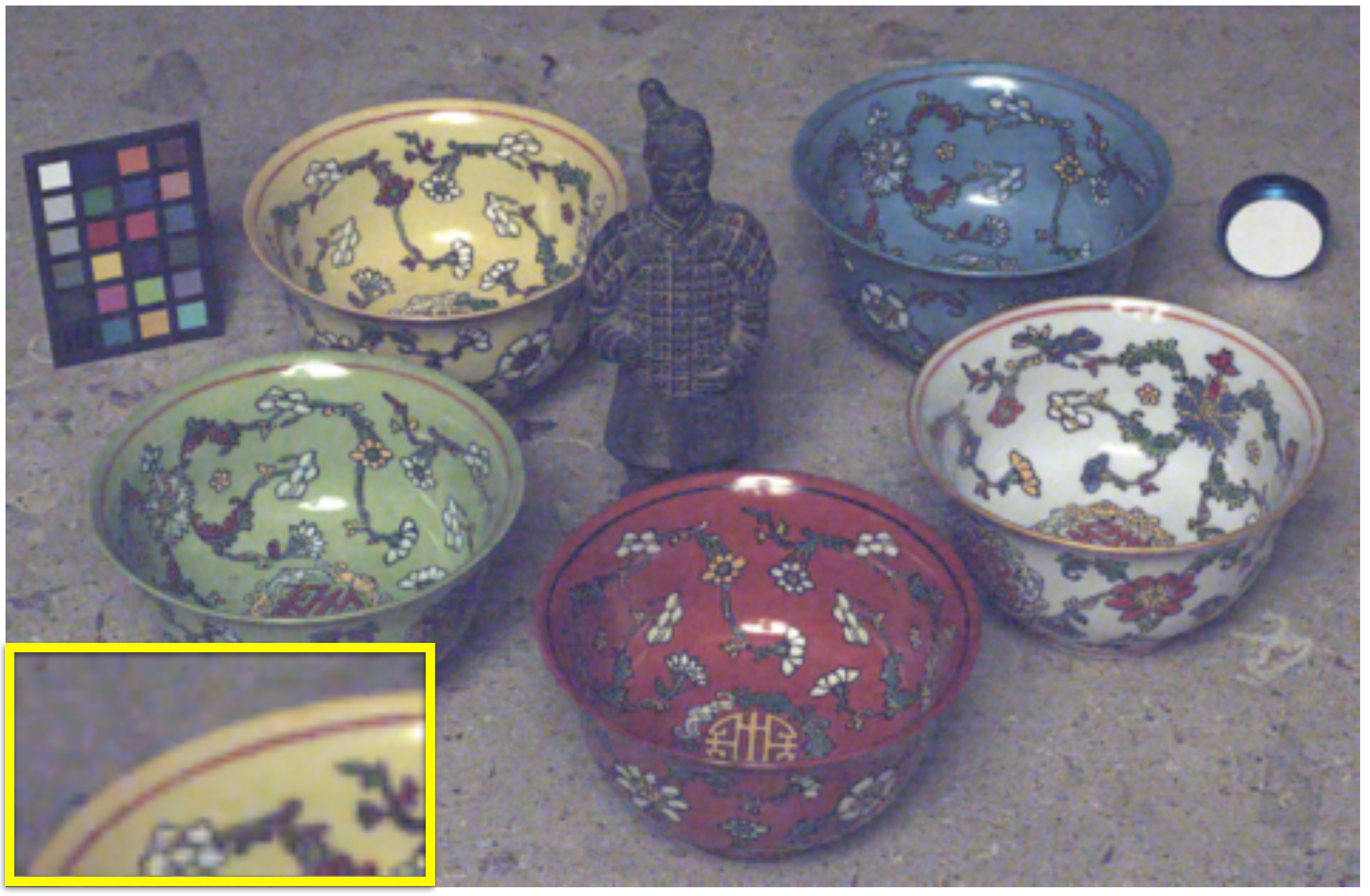} &
\includegraphics[height = .16\linewidth, width = .24\linewidth]{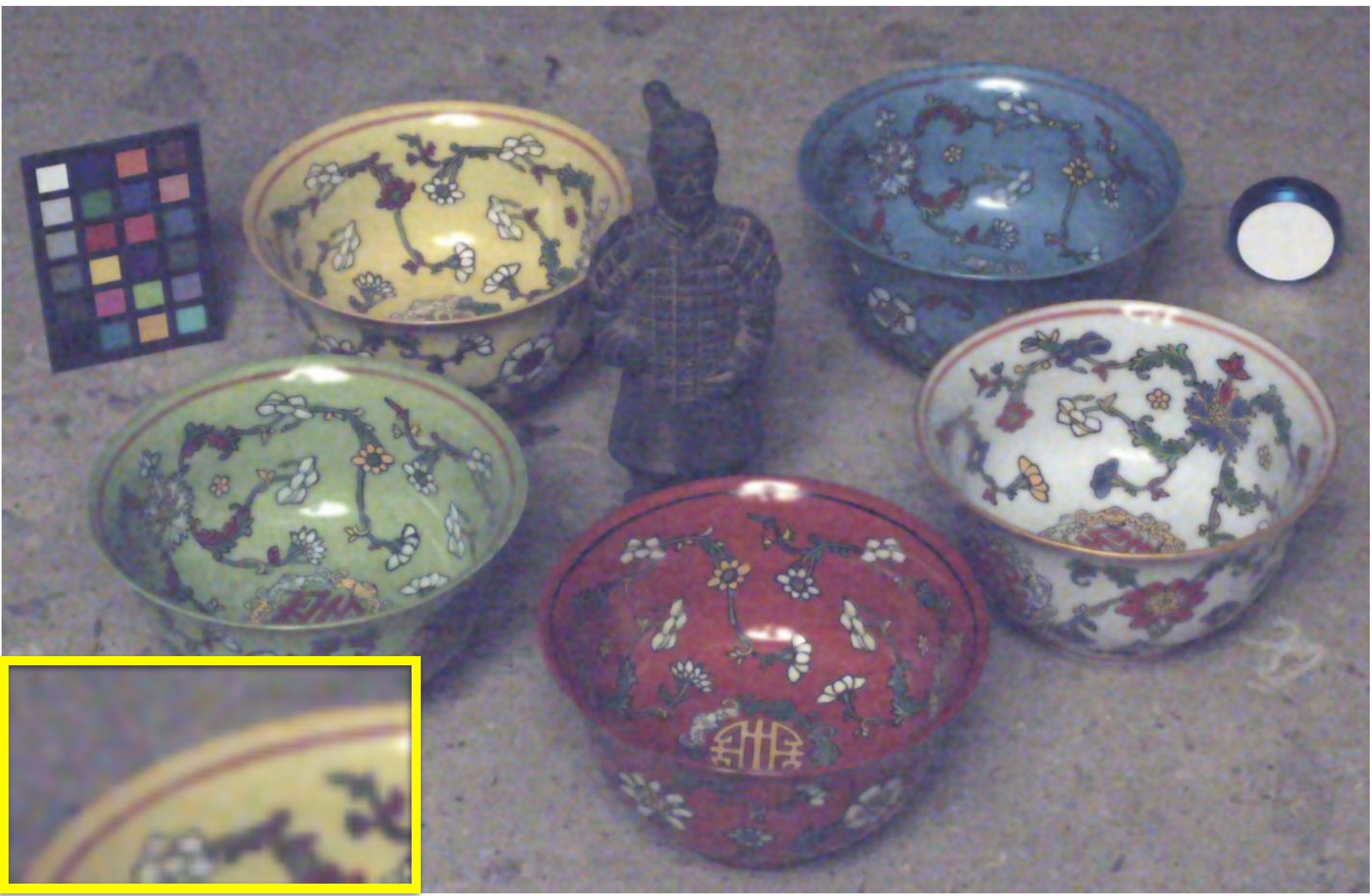} & 

\hspace{1pt}\vrule\hspace{1pt}

\includegraphics[height = .16\linewidth, width = .24\linewidth]{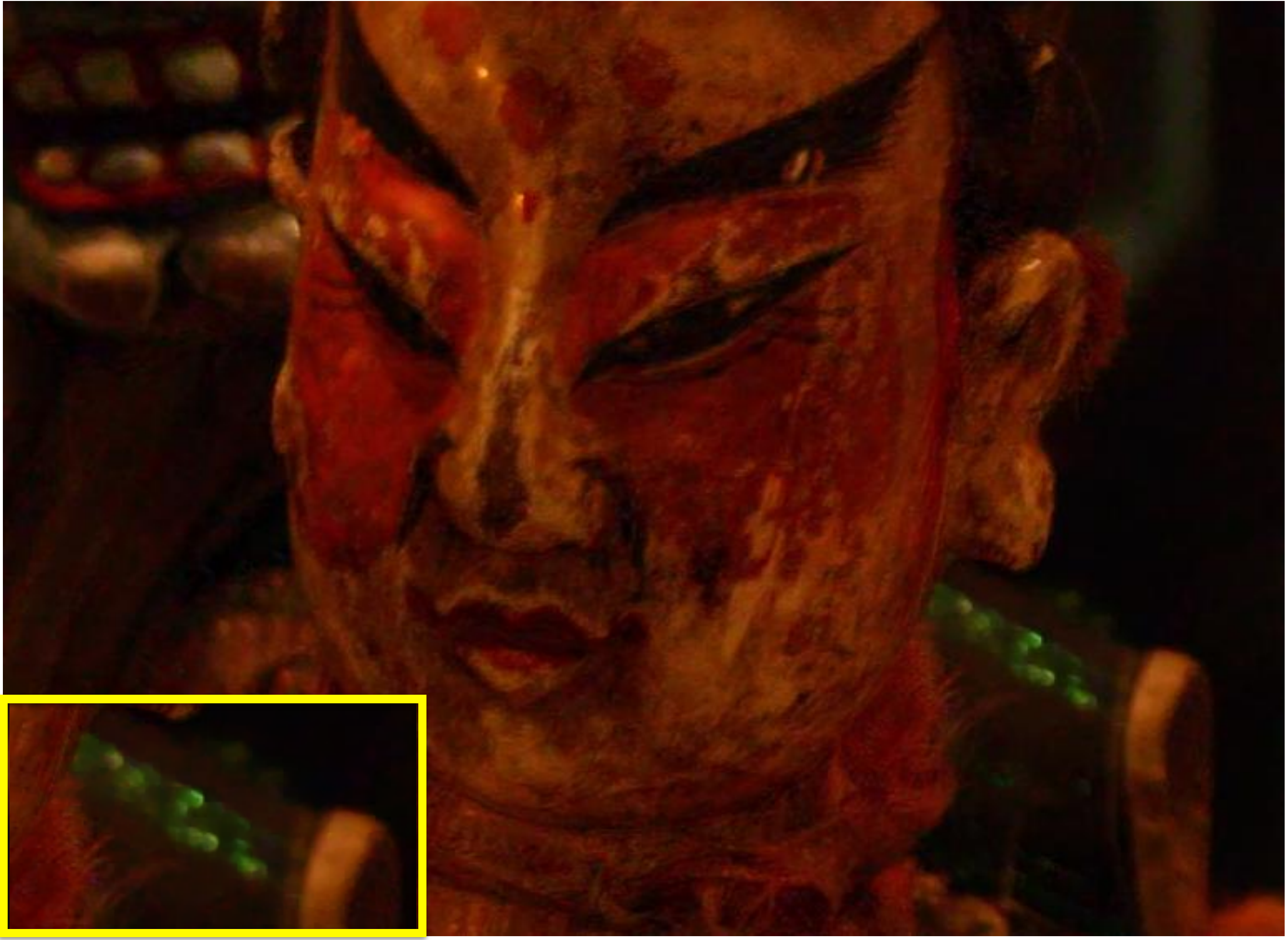} &
\includegraphics[height = .16\linewidth, width = .24\linewidth]{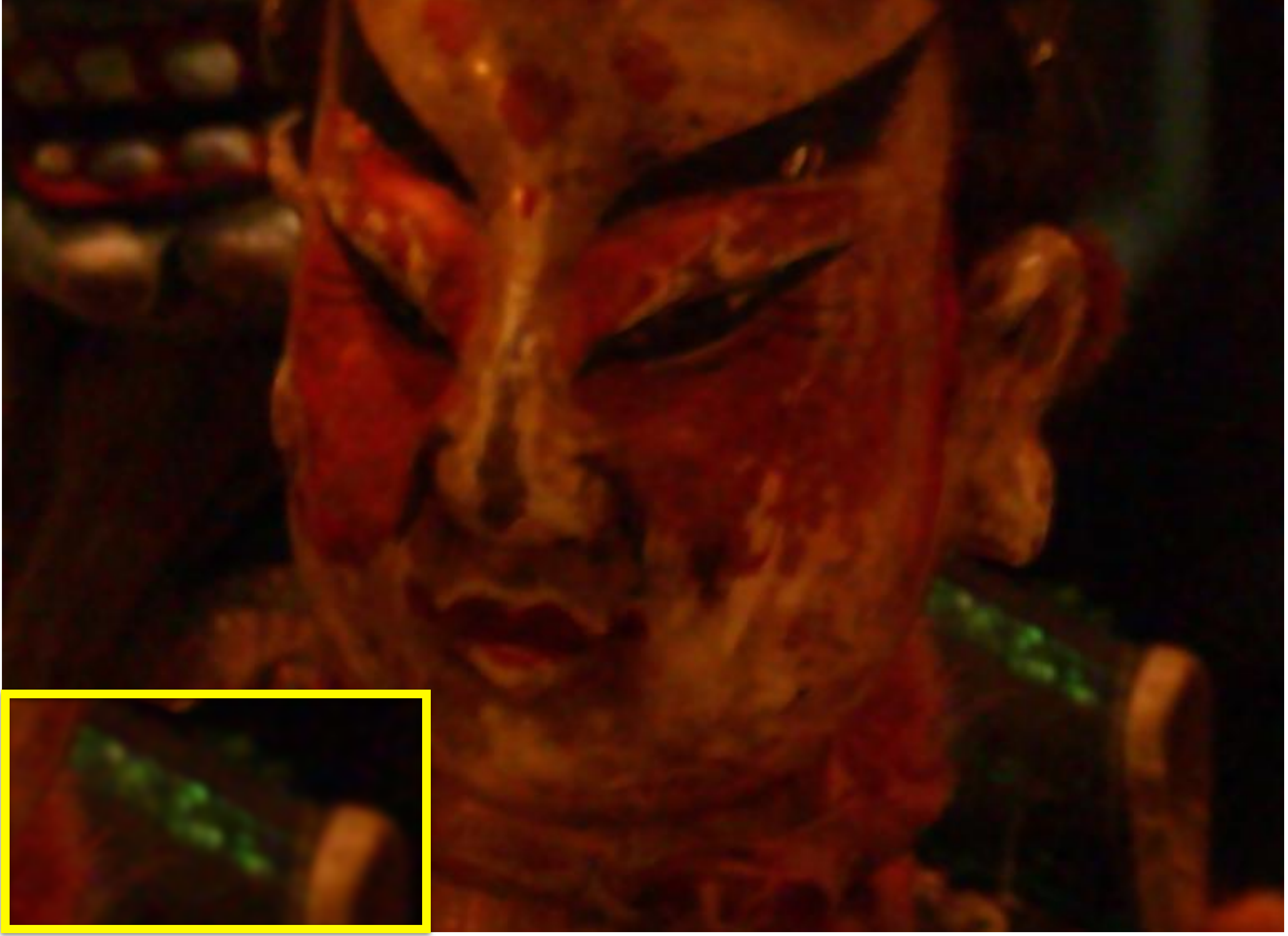} & \\

{Restoration \cite{NIR-ICCV-2013}}& {GF~\cite{He-PAMI-2013}, 7.81} &{Restoration \cite{NIR-ICCV-2013}}& {GF~\cite{He-PAMI-2013}, 8.96} \\

\includegraphics[height = .16\linewidth, width = .24\linewidth]{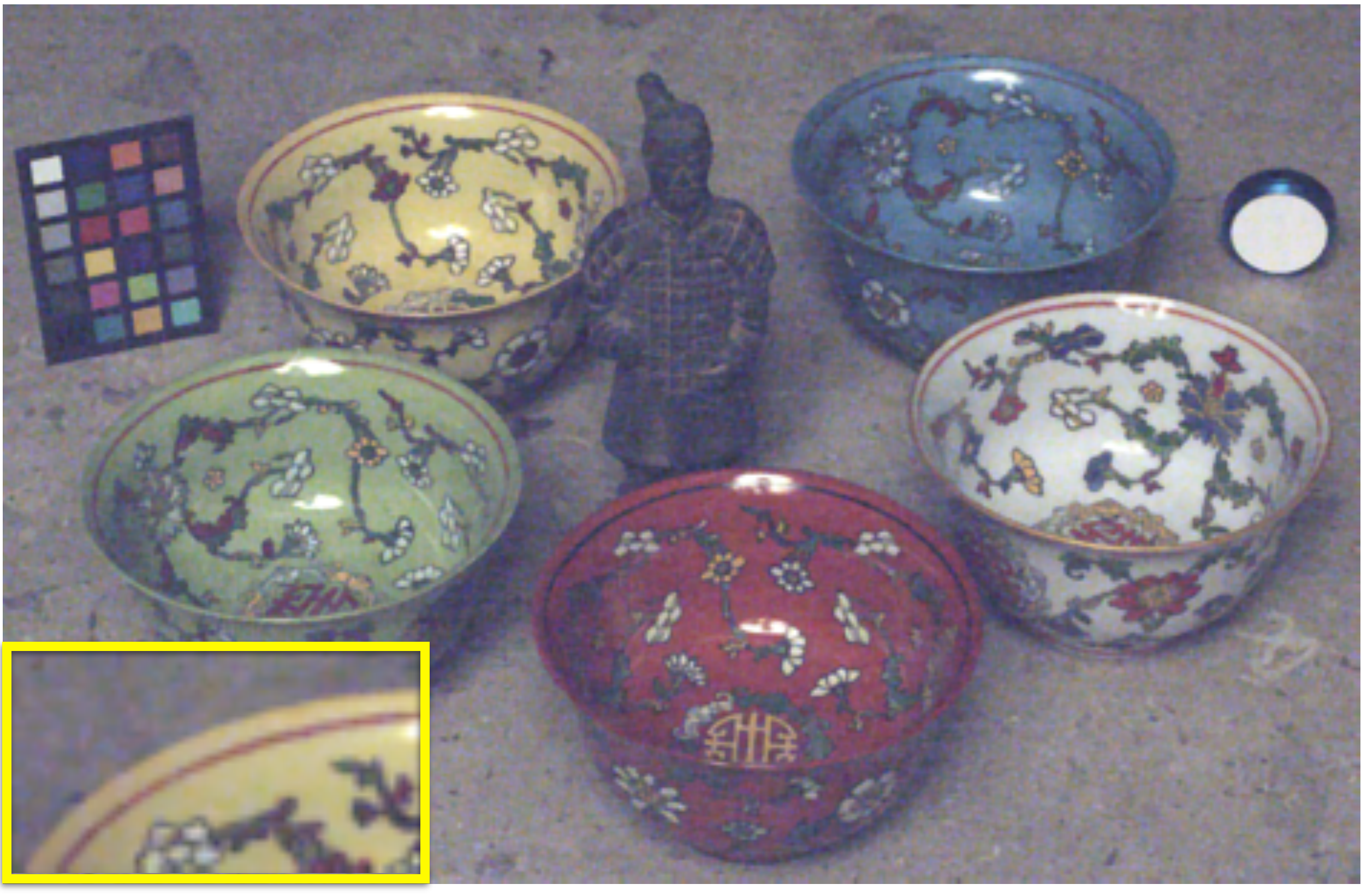} &
\includegraphics[height = .16\linewidth, width = .24\linewidth]{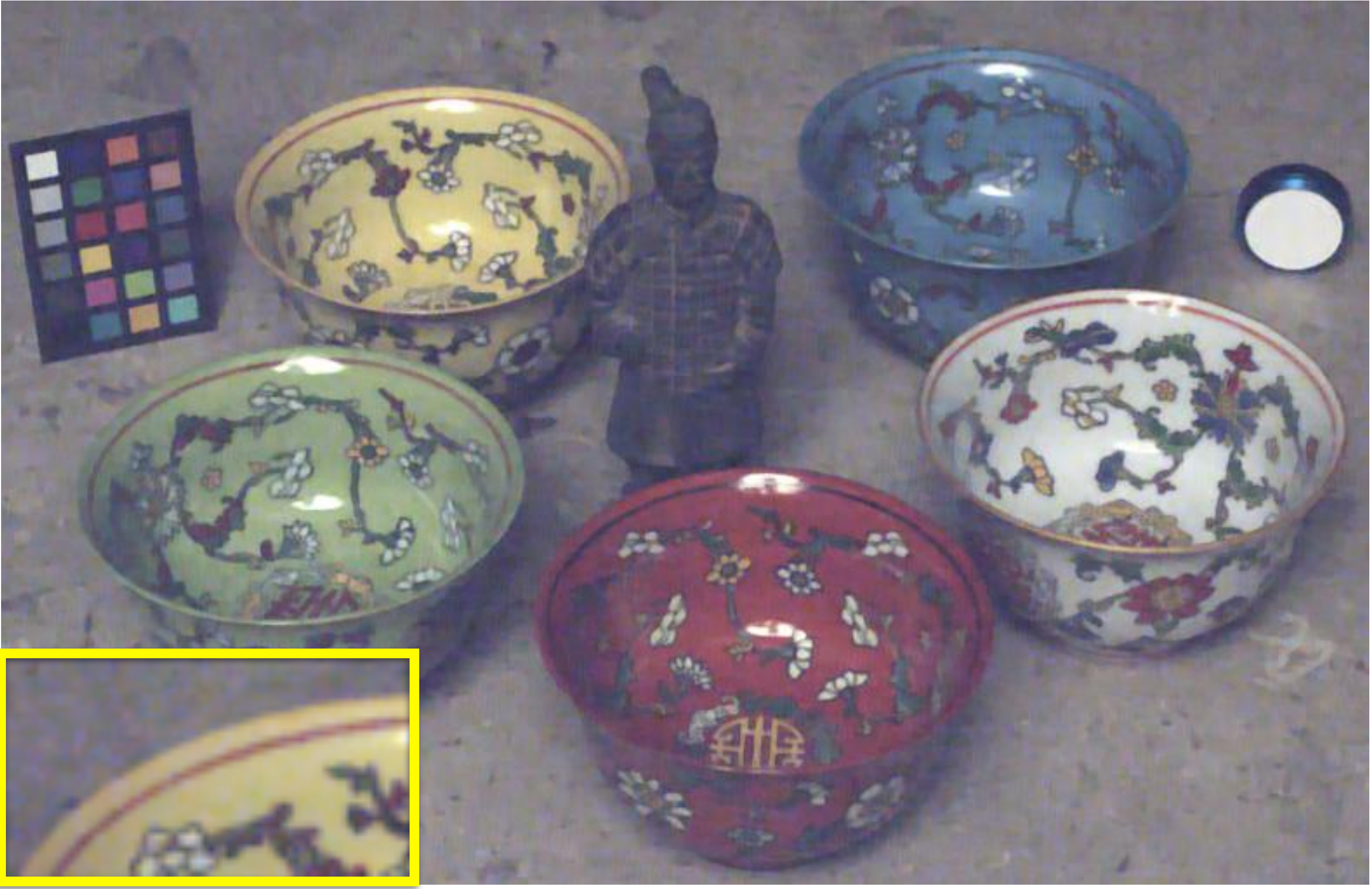} & 

\hspace{1pt}\vrule\hspace{1pt}

\includegraphics[height = .16\linewidth, width = .24\linewidth]{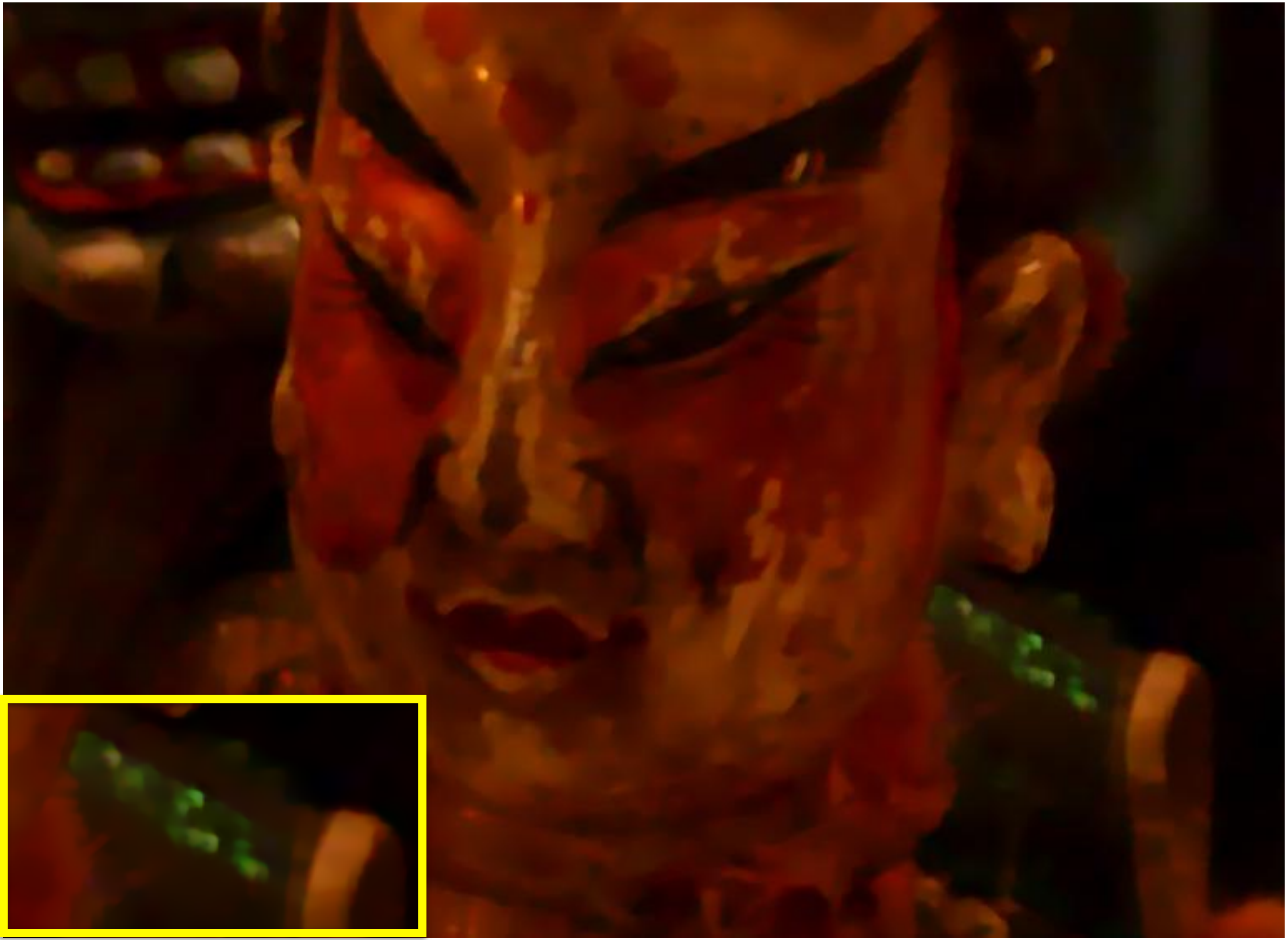} &
\includegraphics[height = .16\linewidth, width = .24\linewidth]{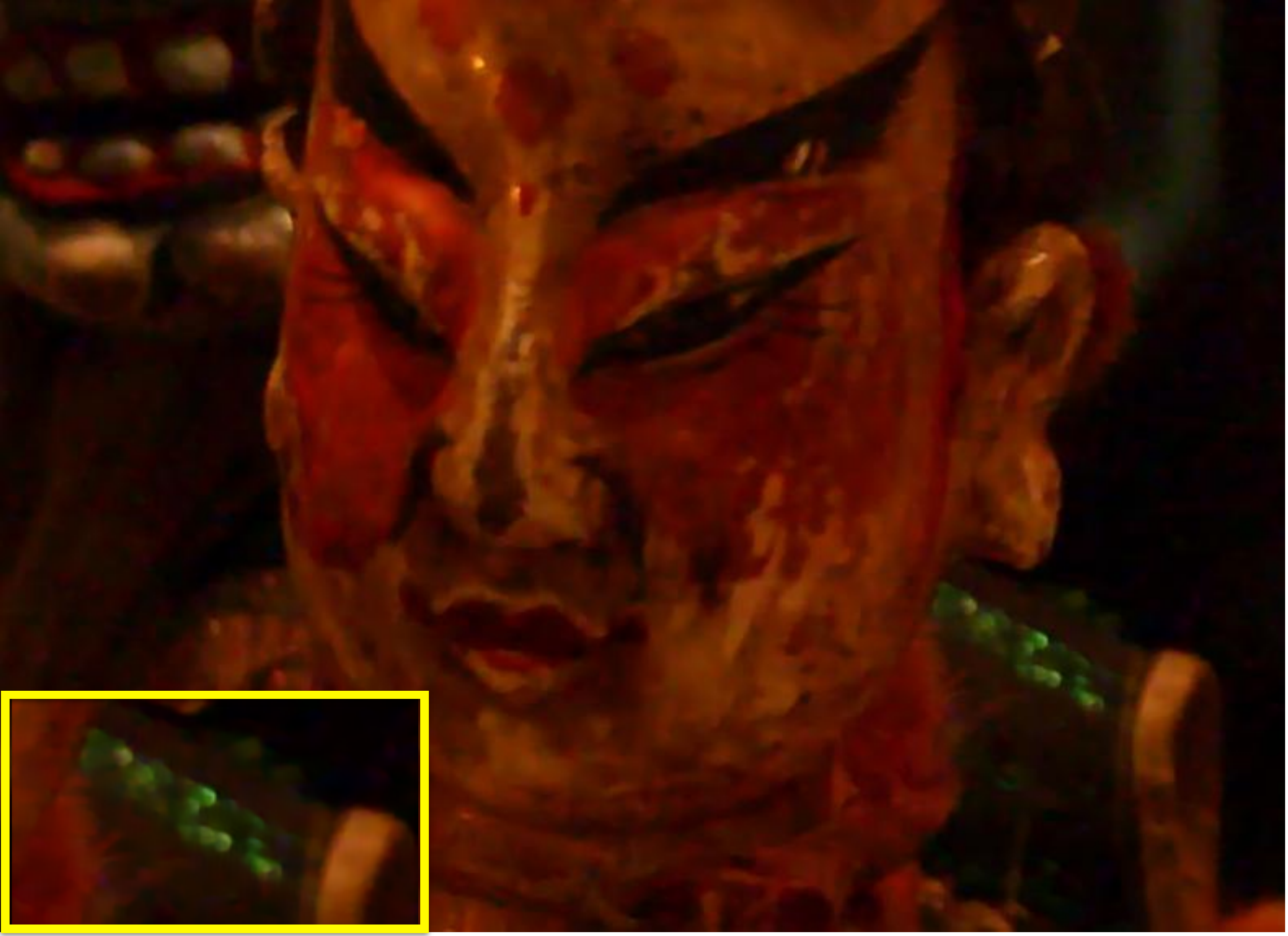} & \\

{DJF~\cite{DJF-ECCV-2016}, 7.44}& {Ours, \textbf{7.19}} & {DJF~\cite{DJF-ECCV-2016}, 8.49}& {Ours, \textbf{8.22}}\\

\end{tabular}

\caption{\revyjnew{\textbf{Cross-modality filtering for noise reduction.}
Left: Results of noise reduction using RGB/NIR image pairs (Target: RGB, Guidance: NIR). Right: Results of noise reduction using flash/non-flash image pairs (Target: Non-Flash, Guidance: Flash). The numbers are the RMSE metric comparing against the result of~\cite{NIR-ICCV-2013}.} 
}
\label{fig:NIR}
\end{figure*}

For the colorization task, we first compute the chromaticity map on the downsampled (4$\times$) image using the user-specified color scribbles \cite{Levin-TOG-2004}.
We then use the original high-resolution intensity image as the guidance image to jointly upsample the low-resolution chromaticity map.
Figure~\ref{fig:colorization} shows that our model is able to achieve visually pleasing results with fewer color bleeding artifacts and efficiently.
Our results are visually similar to the direct solutions on the high-resolution intensity images (Figure~\ref{fig:colorization}(b)).
The quantitative comparisons are presented in the first row of Table~\ref{table:quan_color}.
We use the direct solution of \cite{Levin-TOG-2004} on the high-resolution image as ground truth and compute the RMSE over seven test images in \cite{Levin-TOG-2004}.
Table~\ref{table:quan_color} shows that our method performs well with the lowest error.
Note that our pipeline (low-res result + joint upsampling) is nearly three times faster (2.82 seconds) than directly running the colorization algorithm~\cite{Levin-TOG-2004} on the original pixel grid to obtain the high-resolution result (8.20 seconds). \revyjnew{Note that for fair comparisons, all run-time results are obtained based on the CPU mode.}

%For saliency detection, we first compute the saliency map on the downsampled (10$\times$) image using Manifold et al.~\cite{Manifold-2013-saliency}.
For saliency detection, we first compute the saliency map on the downsampled (10$\times$) image using the manifold method by Yang et al. ~\cite{Manifold-2013-saliency}.
We then use the original high-resolution intensity image as guidance to upsample the low-resolution saliency map.
%We show that the proposed method generates sharper edges when compared with other alternatives.
Figure~\ref{fig:saliency} shows the saliency detection results by the state-of-the-art methods and proposed algorithm.
Overall, the proposed algorithm generates sharper edges than other alternatives.
In addition, we present quantitative evaluation using the ASD benchmark dataset~\cite{ASD-2009-frequency} which consists of 1,000 images
with manually labeled ground truth.
Table~\ref{table:quan_color} shows the comparison between different upsampling methods and our approach in terms of F-measure~\cite{mai-2014-comparing}.
The experimental results demonstrate that the proposed algorithm
performs favorably against the state-of-the-art methods.

\subsection{Structure-texture separation}
We apply our model trained for noise reduction to the task of structure-texture separation.
%
%Here we use the target image itself as the guidance to remove small-scale textures.
\revyjnewminor{Here we use the target image itself as the guidance. We adopt a similar strategy as in the rolling guidance filter (RGF)~\cite{Rolling-ECCV-2014} to remove small-scale textures, i.e., using the output of the previous iteration as the input of the current iteration.}

We use the inverse halftoning task as an example.
A halftoned image is generated by the reprographic technique that simulates continuous tone imagery using various dot patterns~\cite{Kopf-TOG-2012}, as shown in Figure~\ref{fig:half1}(a).
The goal of inverse halftoning is to remove these dots while preserving the main structures.
\revyjnew{We compare our results with those from the RGF~\cite{Rolling-ECCV-2014}, Xu~\cite{Xu-TOG-2012}, DJF~\cite{DJF-ECCV-2016} and the method by Kopf~\cite{Kopf-TOG-2012} for halftoned images reconstruction.
Since there exists no ground truth data, we use the results from Kopf~\cite{Kopf-TOG-2012} as the \revyjnewminor{pseudo} ground truth as it is specifically designed for reconstructing halftoned images and achieves the best visual quality.
For~\cite{Rolling-ECCV-2014, Xu-TOG-2012}, we carefully select the parameters (listed in Figure~\ref{fig:half1}) for the optimal results by considering both removing the dot patterns and keeping the sharp edges intact.
\revyjnewminor{We use the same high-resolution test images from~\cite{Kopf-TOG-2012} and present two zoomed-in patch examples in Figure~\ref{fig:half1} for illustration, where one (top) is with small-scale dots and another one (bottom) is with large-scale dots.}
\revyjnewminor{For the DJF~\cite{DJF-ECCV-2016} and proposed method, we show the results of running two iterations in the first row and three iterations in the second row of Figure~\ref{fig:half1}(e)-(f).}
Our model achieves better results on removing small-scale dots but worse results on removing large-scale dots compared with the methods in~\cite{Rolling-ECCV-2014, Xu-TOG-2012}. 
%While it is more flexible for~\cite{Rolling-ECCV-2014, Xu-TOG-2012} to adjust parameters for different kinds of inputs, our model (trained on RGB/depth data only) is not expected to always achieve the best performance but able to generalize well for comparable results on the inverse halftoning task.
%
However, in order to get the best results, both~\cite{Rolling-ECCV-2014, Xu-TOG-2012} require to manually select optimal parameters for different inputs.   
Our model (trained on RGB/depth only) is not expected to consistently achieve the best performance but able to generalize well for comparable results on the inverse halftoning task without tuning parameters.
}

%Figure~\ref{fig:smooth} shows another application of our filter for image smoothing.
%
%The goal here is to remove insignificant details while retaining and sharpening salient edges.
%
%We compare our algorithm with the RGF~\cite{Rolling-ECCV-2014}, L0~\cite{L0-TOG-2011}, Xu~\cite{Xu-TOG-2012}, as well as Ham~\cite{Ham-CVPR-2015} methods, and select the corresponding parameters for the optimal results.
%
%The proposed algorithm achieves comparable results against all the other methods.
%
%By removing details located in petals and the background, the proposed algorithm also well preserves the silhouette of the green bush on the right while other methods over-smooth these regions.

\subsection{Cross-modality filtering for noise reduction}

Here, we demonstrate that our model can handle various visual domains through two noise reduction applications using RGB/NIR and flash/non-flash image pairs.
Figure~\ref{fig:NIR} (left) show sample results on joint image denoising with the NIR guidance image.
The filtering results by our method are comparable to those of the state-of-the-art technique~\cite{NIR-ICCV-2013}.
For flash/non-flash image pairs, we aim to merge the ambient qualities of the no-flash image with the high-frequency details of the flash image.
Guided by a flash image, the filtering result of our method is comparable to that of \cite{NIR-ICCV-2013}, as shown in Figure~\ref{fig:NIR} (right).
%

%%--------------------------------------------------------------------------------------------------------------------------------
%%--------------------------------------------------------------------------------------------------------------------------------
%%--------------Discussion------------------------------------------------------------------------------------------------------
%%--------------------------------------------------------------------------------------------------------------------------------
%%-------------------------------------------------------------------------------------------------------------------------------

\section{Discussions}
\label{sec:dis}

In this section, we first analyze the effects of the performance under different hyper-parameter settings using the network architecture in Figure~\ref{fig:framework}.
Then, we discuss several limitations of the proposed algorithm.
% As suggested in~\cite{SRCNN-PAMI-2015} that the number of layers does not play a significant role for low-level tasks, we mainly vary the filter number $n_i$ and size $f_i$ (i=1, 2, 3) of each layer in each sub-network.
\revyj{To validate the design choices, we vary the filter number $n$, filter size $f$, and depth $d$ of each sub-network.}
We use the same training process as described in Section~\ref{sec:exp} and evaluate different models on the NYU v2 dataset~\cite{NYU-ECCV-2012} for 8$\times$ upsampling in terms of RMSE.

\begin{table}[t]
%\vspace{-2mm}
\caption{Quantitative results (RMSE in centimeters for 8$\times$) of using different filter \textbf{numbers} in each sub-network. We apply the same parameters to three sub-networks. Top: without the skip connection, Bottom: with the skip connection.}
\label{table:param1}
\centering
\begin{tabular}{cccc}
\toprule

$~~n_1=256$ & $~~n_1=128$~~ & ~~$n_1=96$~~ & ~~$n_1=64$~~ \\
$~~n_2=128,$ & $~~n_2=64$~~ & ~~$n_2=48$~~ & ~~$n_2=32$~~ \\
\midrule
6.40 & 6.44 & 6.32 & 6.35 \\
5.82 & 5.84 & 5.90 & 5.97 \\
\bottomrule
\end{tabular}
%\vspace{-2mm}
\end{table}

\subsection{Filter number}
We first analyze the effects of the number of filters ($n_1$, $n_2$) in first two layers of each sub-network.
The quantitative results are shown in Table~\ref{table:param1}.
% illustrates some interesting findings.
%
In the setting of without the skip connection (top row), we observe that larger filter number may not always result in performance improvements because it increases the difficulty of training the network.
The results suggest that the performance of such network design is somewhat saturated with the sufficient number of filters.
% Meanwhile,
% it also indicates that the current network design may already meet its saturation point of performance at a certain filter number.
%
In order to get further improvements, we need to adjust the network design or the learning objectives, rather than simply modifying hyper-parameters.

Such a hypothesis is supported by the setting of with the skip connection, where we add a skip connection to the entire network and reformulate the network as learning residual functions.
The bottom row of Table~\ref{table:param1} shows that the filter number do yield progressive improvements when it is increased.
This is in accordance with the observation in~\cite{resnet-2016-deep,VDSR-2016-accurate} where residual learning is more effective for training the network with larger capacity.
However, a larger network also slows down the training process and may only provide marginal performance improvements.
Consequently, the selected hyper-parameters of our method (shown in Figure~\ref{fig:framework}) strike a good balance between accuracy and computational efficiency.

Furthermore, we discuss the effects of the output channels ($n_3$) of $\mathrm{CNN_T}$ and $\mathrm{CNN_G}$ and show the results in Table~\ref{table:param2}.
Intuitively, using multi-dimensional features may improve the model capacity and therefore its performance.
However, our experimental results indicate that using multi-dimensional feature maps only slows down the training process without clear performance gain, for both without and with the skip connection settings.
Therefore, we set the output feature maps extracted from the target and guidance images as one single channel ($n_3=1$).
%
%The output of $\mathrm{CNN_T}$ and $\mathrm{CNN_G}$ can be viewed as a transformed pair of the original input target/guidance pair.

\subsection{Filter size}
We examine the network sensitivity to the spatial support of the filters.
With all the other experimental settings kept the same, we gradually increase the filter size $f_i$ (i=1, 2, 3) in different layers and show the corresponding performance in Table~\ref{table:param3}.

Starting from using small filter sizes ($f_1=5$, $f_2=1$, $f_3=3$), we observe a steady trend of improvements when increasing the filter sizes.
This is because smaller filters will restrict the network to focus on detailed local smooth regions that provide little information for restoration.
In contrast, a reasonably large filter size can cover richer structural cues that lead to better results.
However, when we further enlarge the filter size (e.g.., up to $f_1=11$, $f_2=3$, $f_3=7$), we do not see additional performance gain.
We attribute this to the increasing difficulty of network training because larger filter sizes indicate more number of parameters to be learned.
Consequently, we choose the filter size $f_1=9$, $f_2=1$, and $f_3=5$ as a good trade-off between the efficiency and performance.

\begin{table}[t]
%\vspace{-2mm}
\caption{Quantitative results (RMSE in centimeters for 8$\times$) of using different filter \textbf{numbers} in the 3rd layer of $\mathrm{CNN_T}$ and $\mathrm{CNN_G}$. Top: without the skip connection, Bottom: with the skip connection.}
%\vspace{-1mm}
\label{table:param2}
\centering
\begin{tabular}{cccc}
\toprule

$~n_3=1$~ & $~n_3=16$~ & $~n_3=32$ & $~n_3=64$ \\
\midrule
6.20 & 6.40 & 6.24 & 6.34 \\
5.86 & 6.11 & 5.93 & 6.02 \\
\bottomrule
\end{tabular}
%\vspace{-2mm}
\end{table}

\begin{table}[t]
%\vspace{-2mm}
\caption{Quantitative results (RMSE in centimeters for 8$\times$) of using different filter \textbf{sizes} in each sub-network. Top: without the skip connection, Bottom: w/ the skip connection.}
%\vspace{-1mm}
\label{table:param3}
\centering
\begin{tabular}{ccccc}
\toprule

$~f_1=11$~ & $~f_1=9$~ & $~f_1=9$ & $~f_1=7$ & $~f_1=5$\\
$~f_2=3$~ & $~f_2=3$~ & $~f_2=1~$ & $~f_2=1$ & $~f_2=1$\\
$~f_3=7$~ & $~f_3=7$~ & $~f_3=5~$ & $~f_3=5$ & $~f_3=3$\\
\midrule
6.28 & 6.40 & 6.20 & 6.47 & 6.62\\
5.93 & 6.05 & 5.86 & 6.06 & 6.24\\
\bottomrule
\end{tabular}
%\vspace{-2mm}
\end{table}

\subsection{Network depth}

\revyj{
As suggested in~\cite{SRCNN-PAMI-2015} that the number of layers does not play a significant role in non-residual based models for low-level tasks, we focus on evaluating the residual-based model (with the skip connection) with different network depth.
}
\revyj{
First, we analyze whether using one generic but deeper residual-based 
$\mathrm{CNN_F\_R}$ network can improve the performance.
We gradually increase the depth from 3 to 8 and show the results in Table~\ref{table:cnnfdepth}.
Overall, the performance of the $\mathrm{CNN_F\_R}$ network 
improves with a deeper network.
However, the performance quickly reaches the point of diminishing returns after $d$ is larger than $4$.
%
% of increasing the network depth indeed getting better but improved more and more slowly, especially after $d=5$.
%
% It is possible to keep enlarging the network to even deeper levels.
%
% However as analyzed in Section~\ref{sec:design}, such a design rational is originally not reasonable, without extracting effective features from the target and guidance image first.
%
% In contrast, our model shows much more obvious improvements through using three subnetworks.
}

\revyj{
Next, we evaluate our model (three subnetworks) by increasing the network depth.
We simultaneously increase the depth $d$ of each subnetwork from 2 to 5 and show the corresponding results in Table~\ref{table:networkdepth}.
We observe that equipped with the skip connection a deeper network generally leads to better performance.
This is in accordance with the observation in~\cite{VDSR-2016-accurate} where a 20-layer deep residual net is used for image super-resolution.
However, in our case with three subnetworks, the deeper network also induces fast 
growth of model size as well as longer training time.
We find the performance improvement is incremental when $d$ is varied from $3$ to $5$.
Thus, we set $d$ to $3$ as a trade-off between model size and performance.
}

\begin{table}[t]
%\vspace{-2mm}
\caption{\revyj{Quantitative evaluation (RMSE in centimeters for 8$\times$) when using residual-based $\mathrm{CNN_F\_R}$ only under different network depth $d$.}}
%\vspace{-1mm}
\label{table:cnnfdepth}
\centering
\begin{tabular}{ccccccc}
\toprule

$~d=3$ & $~d=4$ & $~d=5$& $~d=6$ & $~d=7$ & $~d=8$ & ~Ours\\
\midrule
6.31 & 6.25 & 6.22 & 6.20 & 6.17 & 6.16 & 5.86\\
\bottomrule
\end{tabular}
%\vspace{-2mm}
\end{table}

\begin{table}[t]
%\vspace{-2mm}
\caption{\revyj{Quantitative evaluation of our model by increasing the number of layers (the depth $d$) used in each subnetwork. %The architecture of each subnetwork is shown as $f_1$-$f_2$-... where $f_i$ is the filter size of the $i$-th layer.
}}
%\vspace{-1mm}
\label{table:networkdepth}
\centering
\begin{tabular}{ccccc}
\toprule

$$~ & $~d=2 $~ & $~d=3$ & $~d=4$ & $~d=5$\\
%$$~ & ~9-5 ~ & ~9-1-5 & ~9-5-1-5 & ~9-5-5-1-5\\
\midrule
RMSE~/~cm & 5.99 & 5.86 & 5.77 & 5.73 \\
Model size~/~MB & 0.48 & 0.53 & 5.0 & 11.4 \\
\bottomrule
\end{tabular}
%\vspace{-2mm}
\end{table}

\subsection{Merging layer}

%HERE
\revyj{
As shown in Figure~\ref{fig:framework}, the $\mathrm{CNN_T}$ and $\mathrm{CNN_G}$ are merged at the output (third) layer.
Here we further analyze the effect of merging $\mathrm{CNN_T}$ and $\mathrm{CNN_G}$ at different layers.
We fix the whole network depth as 6 and analyze different combinations of network depth of $\mathrm{CNN_{T}}$, $\mathrm{CNN_{G}}$ and $\mathrm{CNN_{F}}$.
We gradually increase the depth of $\mathrm{CNN_{T}}$ and $\mathrm{CNN_{G}}$ while decreasing the depth of $\mathrm{CNN_{F}}$ (in order to maintain the overall network depth).
For example, $0/0-6$ (Table~\ref{table:networkmerge}) indicates that we directly stack the target and guidance image and applying a 6-layer $\mathrm{CNN_F}$ only.
The evaluation results of different models are shown in Table~\ref{table:networkmerge}.
Overall, deeper target/guidance networks ($\mathrm{CNN_{T}}$ and $\mathrm{CNN_{G}}$) result in sizable performance improvement resulting from effective feature extractions.
However, as the $\mathrm{CNN_{F}}$ becomes shallower, the performance degrades again.
This indicates that neither the $\mathrm{CNN_{T}}$ ($\mathrm{CNN_{G}}$) nor the $\mathrm{CNN_{F}}$ should be too shallow.
Therefore, we chose the combination of $3/3-3$ for best performance.
}

\begin{table}[t]
%\vspace{-2mm}
\caption{\revyj{Quantitative evaluation of different combinations of network depth of $\mathrm{CNN_{T}}$ ($\mathrm{CNN_{G}}$) and $\mathrm{CNN_F}$.}}
%\vspace{-1mm}
\label{table:networkmerge}
\centering
\begin{tabular}{ccccc}
\toprule

$\mathrm{CNN_{T}}/\mathrm{CNN_{G}}-\mathrm{CNN_F}$ & ~0/0~--~6 & ~2/2~--~4 & ~3/3~--~3 & ~4/4~--~2 \\
\midrule
RMSE~/~cm & 6.13 & 5.95 & 5.86 & 6.03 \\
\bottomrule
\end{tabular}
%\vspace{-2mm}
\end{table}

\subsection{Limitations}
We note that in some images, our model fails to transfer small-scale details from the guidance map.
In such cases, our model incorrectly treats certain small-scale details as noise.
This can be explained by the fact that our training data is based on depth images that are mostly smooth and does not contain many spatial details.
%
% The depth map usually tends to be smooth and does not contain many details.

\begin{figure}[t]
\centering

\begin{tabular}{c@{\hspace{0.005\linewidth}}c@{\hspace{0.005\linewidth}}c@{\hspace{0.005\linewidth}}c@{\hspace{0.005\linewidth}}c@{\hspace{0.005\linewidth}}c@{\hspace{0.005\linewidth}}c}

\includegraphics[width = .315\linewidth]{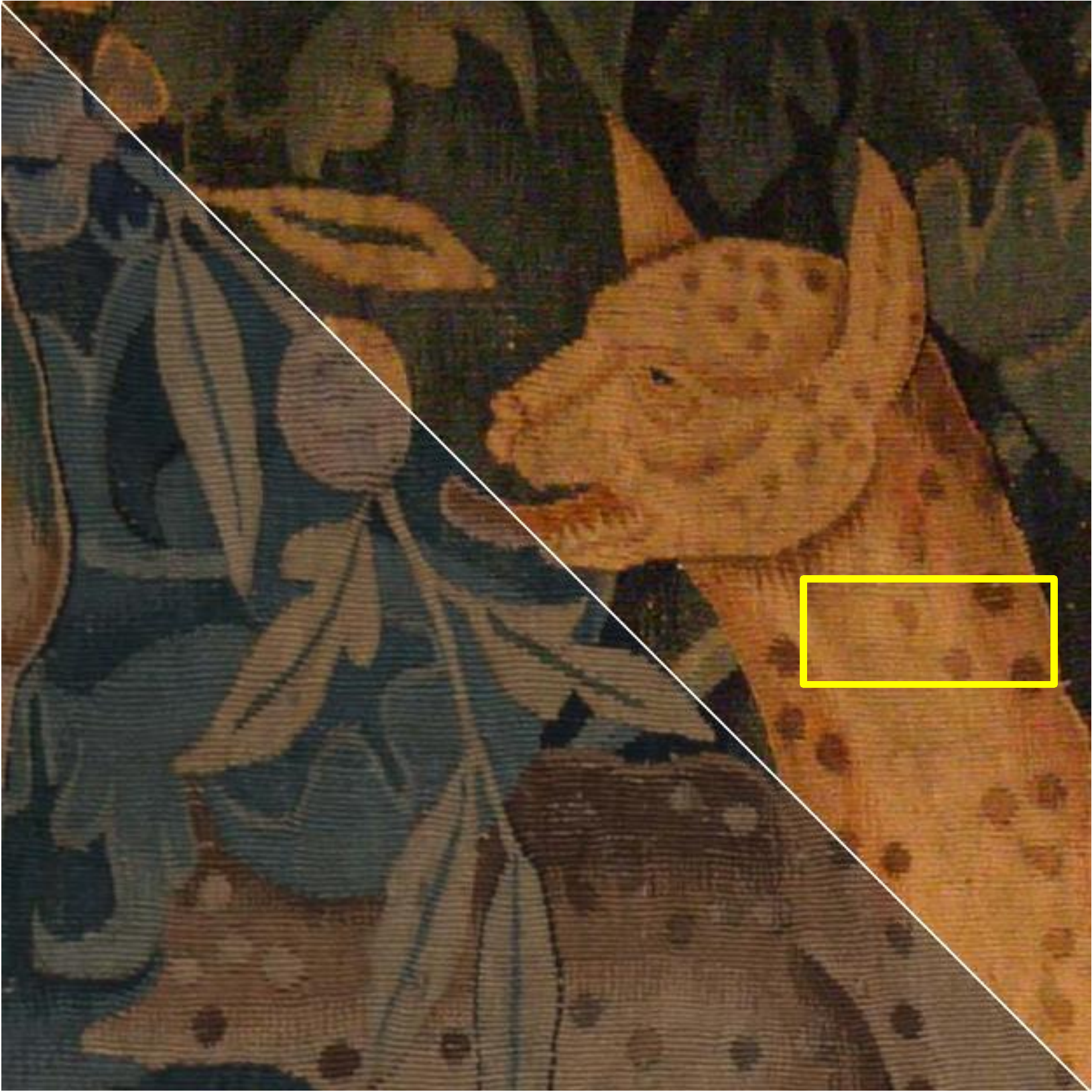} &
\includegraphics[width = .315\linewidth]{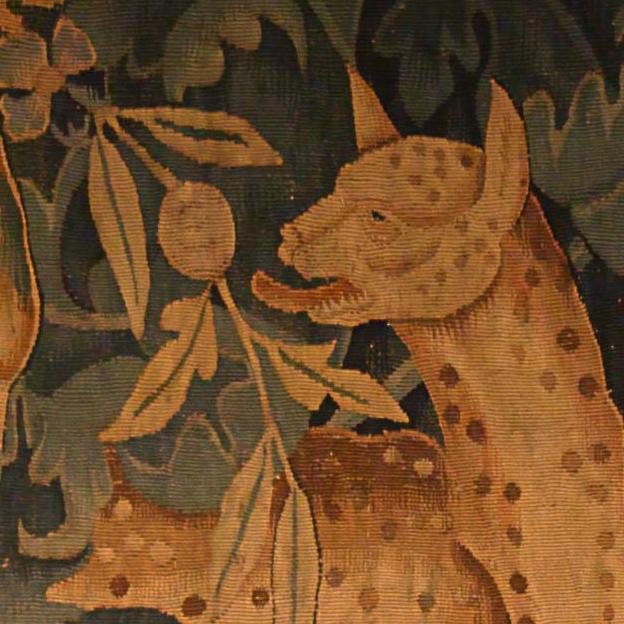} &
\includegraphics[width = .315\linewidth]{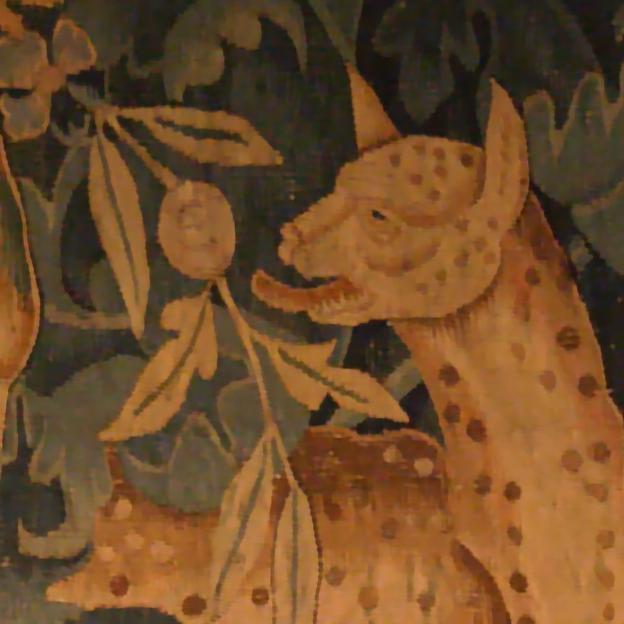} & \\

\includegraphics[height=.12\linewidth, width = .315\linewidth]{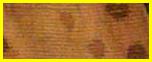} &
\includegraphics[height=.12\linewidth, width = .315\linewidth]{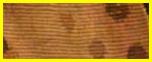} &
\includegraphics[height=.12\linewidth, width = .315\linewidth]{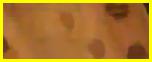} & \\

{(a) Input } & {(b) Georg et al.~\cite{Details-TOG-2004} } & {(c) Ours} \\

\end{tabular}

\caption{\textbf{Failure cases.} Detailed small-scale textures (yellow rectangle) in the guidance image are over-smoothed by our filter.}
\label{fig:Failure}

\end{figure}

Figure~\ref{fig:Failure} shows two examples of a flash/non-flash pair for noise reduction.
There are several spotty textures on the porcelain in the guided flash image that should have been preserved when filtering the noisy non-flash image.
Similarly, our method is not able to effectively transfer the small-scale strip textures on the carpet to the target image.
Compared with the method by Georg et al.~\cite{Details-TOG-2004} (Figure~\ref{fig:Failure}(b) and (d)) that is designed specifically for flash/non-flash images, our filter
treats these small-scale details as noise and tends to over-smooth the contents.
% We will use non-depth data to address the over-smoothing problem in our future work.
We will collect more training data from other domains (e.g., flash/non-flash) to address the over-smoothing problem in our future work.

%%--------------------------------------------------------------------------------------------------------------------------------
%%--------------------------------------------------------------------------------------------------------------------------------
%%--------------Conclusion------------------------------------------------------------------------------------------------------
%%--------------------------------------------------------------------------------------------------------------------------------
%%--------------------------------------------------------------------------------------------------------------------------------

\section{Conclusions}

In this paper, we present a learning-based approach for joint filtering based on convolutional neural networks.
Instead of relying only on the guidance image, we design two sub-networks $\mathrm{CNN_T}$ and $\mathrm{CNN_G}$ to extract informative features from both the target and guidance images.
These feature maps are then concatenated as inputs for the network $\mathrm{CNN_F}$ to selectively transfer salient structures from the guidance image to the target image while suppressing structures that are not consistent in both images.
While we train our network on one type of data (RGB/depth or RGB/flow), our model generalizes well on handling images in various modalities, e.g., RGB/NIR and flash/non-Flash image pairs.
We show that the proposed algorithm is computationally efficient and
performs favorably against
the state-of-the-art techniques on a wide variety of computer vision and computational photography applications, including cross-modal denoising, joint image upsampling,
and texture-structure separation.
%

\ifCLASSOPTIONcompsoc
%\section*{Acknowledgments}
%\else
%\section*{Acknowledgment}
%\fi

%This work is supported in part by the NSF CAREER Grant \#1149783, gifts from Adobe and Nvidia, and Office of Naval Research under grant N00014-16-1-2314.

\ifCLASSOPTIONcaptionsoff
\newpage
\fi

%\clearpage

\bibliographystyle{IEEEtran}
\bibliography{joint_filter}

\begin{thebibliography}{10}
\providecommand{\url}[1]{#1}
\csname url@samestyle\endcsname
\providecommand{\newblock}{\relax}
\providecommand{\bibinfo}[2]{#2}
\providecommand{\BIBentrySTDinterwordspacing}{\spaceskip=0pt\relax}
\providecommand{\BIBentryALTinterwordstretchfactor}{4}
\providecommand{\BIBentryALTinterwordspacing}{\spaceskip=\fontdimen2\font plus
\BIBentryALTinterwordstretchfactor\fontdimen3\font minus
  \fontdimen4\font\relax}
\providecommand{\BIBforeignlanguage}[2]{{%
\expandafter\ifx\csname l@#1\endcsname\relax
\typeout{** WARNING: IEEEtran.bst: No hyphenation pattern has been}%
\typeout{** loaded for the language `#1'. Using the pattern for}%
\typeout{** the default language instead.}%
\else
\language=\csname l@#1\endcsname
\fi
#2}}
\providecommand{\BIBdecl}{\relax}
\BIBdecl

\bibitem{Yang-CVPR-2007}
Q.~Yang, R.~Yang, J.~Davis, and D.~Nist{\'e}r, ``Spatial-depth super resolution
  for range images,'' in \emph{IEEE Conference on Computer Vision and Pattern
  Recognition}, 2007.

\bibitem{Park-ICCV-2011}
J.~Park, H.~Kim, Y.-W. Tai, M.~S. Brown, and I.~Kweon, ``High quality depth map
  upsampling for 3d-tof cameras,'' in \emph{IEEE International Conference on
  Computer Vision}, 2011.

\bibitem{TGV-ICCV-2013}
D.~Ferstl, C.~Reinbacher, R.~Ranftl, M.~R{\"u}ther, and H.~Bischof, ``Image
  guided depth upsampling using anisotropic total generalized variation,'' in
  \emph{IEEE International Conference on Computer Vision}, 2013.

\bibitem{JBU-TOG-2007}
J.~Kopf, M.~F. Cohen, D.~Lischinski, and M.~Uyttendaele, ``Joint bilateral
  upsampling,'' in \emph{ACM SIGGRAPH}, 2007.

\bibitem{NIR-ICCV-2013}
Q.~Yan, X.~Shen, L.~Xu, S.~Zhuo, X.~Zhang, L.~Shen, and J.~Jia, ``Cross-field
  joint image restoration via scale map,'' in \emph{IEEE International
  Conference on Computer Vision}, 2013.

\bibitem{He-PAMI-2013}
K.~He, J.~Sun, and X.~Tang, ``Guided image filtering,'' \emph{IEEE Transactions
  on Pattern Analysis and Machine Intelligence}, vol.~35, no.~6, pp.
  1397--1409, 2013.

\bibitem{Shen-ICCV-2015}
X.~Shen, C.~Zhou, L.~Xu, and J.~Jia, ``Mutual-structure for joint filtering,''
  in \emph{IEEE International Conference on Computer Vision}, 2015.

\bibitem{Xu-TOG-2012}
L.~Xu, Q.~Yan, Y.~Xia, and J.~Jia, ``Structure extraction from texture via
  relative total variation,'' \emph{ACM Transactions on Graphics}, vol.~31,
  no.~6, p. 139, 2012.

\bibitem{Rolling-ECCV-2014}
Q.~Zhang, X.~Shen, L.~Xu, and J.~Jia, ``Rolling guidance filter,'' in
  \emph{European Conference on Computer Vision}, 2014.

\bibitem{BF-ICCV-1998}
C.~Tomasi and R.~Manduchi, ``Bilateral filtering for gray and color images,''
  in \emph{IEEE International Conference on Computer Vision}, 1998.

\bibitem{Eisemann-TOG-2004}
E.~Eisemann and F.~Durand, ``Flash photography enhancement via intrinsic
  relighting,'' in \emph{ACM SIGGRAPH}, 2004.

\bibitem{Details-TOG-2004}
P.~Georg, A.~Maneesh, H.~Hugues, S.~Richard, C.~Michael, and T.~Kentaro,
  ``Digital photography with flash and no-flash image pairs,'' in \emph{ACM
  SIGGRAPH}, 2004.

\bibitem{Ham-CVPR-2015}
B.~Ham, M.~Cho, and J.~Ponce, ``Robust image filtering using joint static and
  dynamic guidance,'' in \emph{IEEE Conference on Computer Vision and Pattern
  Recognition}, 2015.

\bibitem{NYU-ECCV-2012}
P.~K. Nathan~Silberman, Derek~Hoiem and R.~Fergus, ``Indoor segmentation and
  support inference from rgbd images,'' in \emph{European Conference on
  Computer Vision}, 2012.

\bibitem{Song-CVPR-2015}
S.~Song, S.~P. Lichtenberg, and J.~Xiao, ``Sun rgb-d: A rgb-d scene
  understanding benchmark suite,'' in \emph{IEEE Conference on Computer Vision
  and Pattern Recognition}, 2015.

\bibitem{Midd1-CVPR-2007}
D.~Scharstein and C.~Pal, ``Learning conditional random fields for stereo.'' in
  \emph{IEEE Conference on Computer Vision and Pattern Recognition}, 2007.

\bibitem{Midd2-CVPR-2007}
H.~Hirschm{\"u}ller and D.~Scharstein, ``Evaluation of cost functions for
  stereo matching.'' in \emph{IEEE Conference on Computer Vision and Pattern
  Recognition}, 2007.

\bibitem{DJF-ECCV-2016}
Y.~Li, J.-B. Huang, N.~Ahuja, and M.-H. Yang, ``Deep joint image filtering,''
  in \emph{European Conference on Computer Vision}, 2016.

\bibitem{Tai-2016-depth}
T.-W. Hui, C.~C. Loy, and X.~Tang, ``Depth map super-resolution by deep
  multi-scale guidance,'' in \emph{European Conference on Computer Vision},
  2016.

\bibitem{Barron-2016-solver}
J.~T. Barron and B.~Poole, ``The fast bilateral solver,'' in \emph{European
  Conference on Computer Vision}, 2016.

\bibitem{JGU-CVPR-2013}
M.-Y. Liu, O.~Tuzel, and Y.~Taguchi, ``Joint geodesic upsampling of depth
  images,'' in \emph{IEEE Conference on Computer Vision and Pattern
  Recognition}, 2013.

\bibitem{MRF-NIPS-2005}
J.~Diebel and S.~Thrun, ``An application of markov random fields to range
  sensing,'' in \emph{Neural Information Processing Systems}, 2005.

\bibitem{Imagenet-NIPS-2012}
A.~Krizhevsky, I.~Sutskever, and G.~E. Hinton, ``Image{N}et classification with
  deep convolutional neural networks,'' in \emph{Neural Information Processing
  Systems}, 2012.

\bibitem{Jam-2016-HDF}
V.~Jampani, M.~Kiefel, and P.~V. Gehler, ``Learning sparse high dimensional
  filters: Image filtering, dense crfs and bilateral neural networks,'' in
  \emph{IEEE Conference on Computer Vision and Pattern Recognition}, 2016.

\bibitem{gharbi2017deep}
M.~Gharbi, J.~Chen, J.~T. Barron, S.~W. Hasinoff, and F.~Durand, ``Deep
  bilateral learning for real-time image enhancement,'' in \emph{ACM SIGGRAPH},
  2017.

\bibitem{chen-ICCV-2017}
Q.~Chen, J.~Xu, and V.~Koltun, ``Fast image processing with fully-convolutional
  networks,'' in \emph{IEEE International Conference on Computer Vision}, 2017.

\bibitem{gu2017learning}
S.~Gu, W.~Zuo, S.~Guo, Y.~Chen, C.~Chen, and L.~Zhang, ``Learning dynamic
  guidance for depth image enhancement,'' in \emph{IEEE Conference on Computer
  Vision and Pattern Recognition}, 2017.

\bibitem{Denoise-CVPR-2012}
H.~C. Burger, C.~J. Schuler, and S.~Harmeling, ``Image denoising: Can plain
  neural networks compete with bm3d?'' in \emph{IEEE Conference on Computer
  Vision and Pattern Recognition}, 2012.

\bibitem{Rain-ICCV-2013}
D.~Eigen, D.~Krishnan, and R.~Fergus, ``Restoring an image taken through a
  window covered with dirt or rain,'' in \emph{IEEE International Conference on
  Computer Vision}, 2013.

\bibitem{SRCNN-ECCV-2014}
C.~Dong, C.~C. Loy, K.~He, and X.~Tang, ``Learning a deep convolutional network
  for image super-resolution,'' in \emph{European Conference on Computer
  Vision}, 2014.

\bibitem{zhang-2016-learning}
J.~Zhang, J.~Pan, W.-S. Lai, R.~Lau, and M.-H. Yang, ``Learning fully
  convolutional networks for iterative non-blind deconvolution,'' in \emph{IEEE
  Conference on Computer Vision and Pattern Recognition}, 2017.

\bibitem{Flownet-ICCV-2015}
F.~Philipp, D.~Alexey, I.~Eddy, H.~Philip, H.~Caner, G.~Vladimir, V.~d.~S.
  Patrick, C.~Daniel, and B.~Thomas, ``Flow{N}et: Learning optical flow with
  convolutional networks,'' in \emph{IEEE International Conference on Computer
  Vision}, 2015.

\bibitem{Xu-ICML-2015}
L.~Xu, J.~Ren, Q.~Yan, R.~Liao, and J.~Jia, ``Deep edge-aware filters,'' in
  \emph{International Conference on Machine Learning}, 2015.

\bibitem{resnet-2016-deep}
K.~He, X.~Zhang, S.~Ren, and J.~Sun, ``Deep residual learning for image
  recognition,'' in \emph{IEEE Conference on Computer Vision and Pattern
  Recognition}, 2016.

\bibitem{huang2017densely}
G.~Huang, Z.~Liu, K.~Q. Weinberger, and L.~van~der Maaten, ``Densely connected
  convolutional networks,'' in \emph{IEEE Conference on Computer Vision and
  Pattern Recognition}, 2017.

\bibitem{VDSR-2016-accurate}
J.~Kim, J.~Kwon~Lee, and K.~Mu~Lee, ``Accurate image super-resolution using
  very deep convolutional networks,'' in \emph{IEEE Conference on Computer
  Vision and Pattern Recognition}, 2016.

\bibitem{lai2017deep}
W.-S. Lai, J.-B. Huang, N.~Ahuja, and M.-H. Yang, ``Deep laplacian pyramid
  networks for fast and accurate super-resolution,'' in \emph{IEEE Conference
  on Computer Vision and Pattern Recognition}, 2017.

\bibitem{SRCNN-PAMI-2015}
C.~Dong, C.~C. Loy, K.~He, and X.~Tang, ``Image super-resolution using deep
  convolutional networks,'' \emph{IEEE Transactions on Pattern Analysis and
  Machine Intelligence}, vol.~38, no.~2, pp. 295 -- 307, 2015.

\bibitem{Dollar-ICCV-2013}
P.~Doll{\'a}r and C.~L. Zitnick, ``Structured forests for fast edge
  detection,'' in \emph{IEEE International Conference on Computer Vision},
  2013.

\bibitem{Sintel-ECCV-2012}
D.~J. Butler, J.~Wulff, G.~B. Stanley, and M.~J. Black, ``A naturalistic open
  source movie for optical flow evaluation,'' in \emph{European Conference on
  Computer Vision}, 2012.

\bibitem{David-NIPS-2014}
E.~David, P.~Christian, and F.~Rob, ``Depth map prediction from a single image
  using a multi-scale deep network,'' in \emph{Neural Information Processing
  Systems}, 2014.

\bibitem{Matconvnet-ACMMM-2015}
V.~Andrea and L.~Karel, ``Mat{C}onv{N}et -- convolutional neural networks for
  matlab,'' in \emph{ACM Multimedia}, 2015.

\bibitem{Lu-CVPR-2014}
S.~Lu, X.~Ren, and F.~Liu, ``Depth enhancement via low-rank matrix
  completion,'' in \emph{IEEE Conference on Computer Vision and Pattern
  Recognition}, 2014.

\bibitem{Levin-TOG-2004}
A.~Levin, D.~Lischinski, and Y.~Weiss, ``Colorization using optimization,'' in
  \emph{ACM SIGGRAPH}, 2004.

\bibitem{Manifold-2013-saliency}
C.~Yang, L.~Zhang, H.~Lu, X.~Ruan, and M.-H. Yang, ``Saliency detection via
  graph-based manifold ranking,'' in \emph{IEEE Conference on Computer Vision
  and Pattern Recognition}, 2013.

\bibitem{Kopf-TOG-2012}
J.~Kopf and D.~Lischinski, ``Digital reconstruction of halftoned color
  comics,'' in \emph{ACM SIGGRAPH}, 2012.

\bibitem{ASD-2009-frequency}
R.~Achanta, S.~Hemami, F.~Estrada, and S.~Susstrunk, ``Frequency-tuned salient
  region detection,'' in \emph{IEEE Conference on Computer Vision and Pattern
  Recognition}, 2009.

\bibitem{mai-2014-comparing}
L.~Mai and F.~Liu, ``Comparing salient object detection results without ground
  truth,'' in \emph{European Conference on Computer Vision}, 2014.

\end{thebibliography}

\vspace{-.5em}
\begin{IEEEbiography}[{\includegraphics[width=1in,height=1.25in,clip,keepaspectratio]{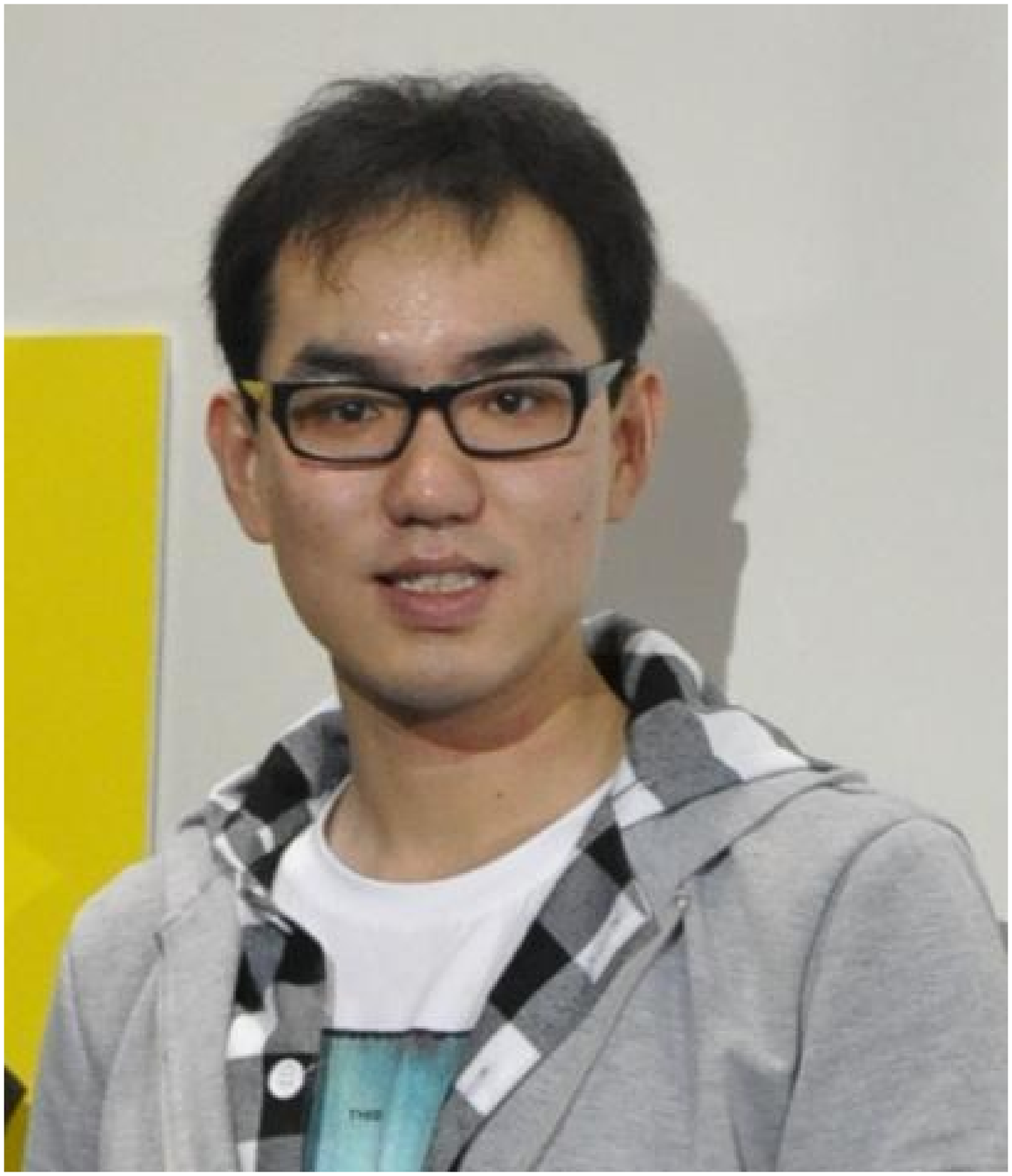}}]{Yijun
Li} is a Ph.D. student of Electrical
Engineering and Computer Science at
University of California, Merced. He
received the M.S. degree from Shanghai Jiao Tong University and B.S. degree from Zhejiang University, in 2015 and 2012 respectively.
His research interests lie in the computer vision and machine learning, including image generation, synthesis and low-level vision.
\end{IEEEbiography}

\vspace{-1.5em}
\begin{IEEEbiography}[{\includegraphics[width=1in,height=1.25in,clip,keepaspectratio]{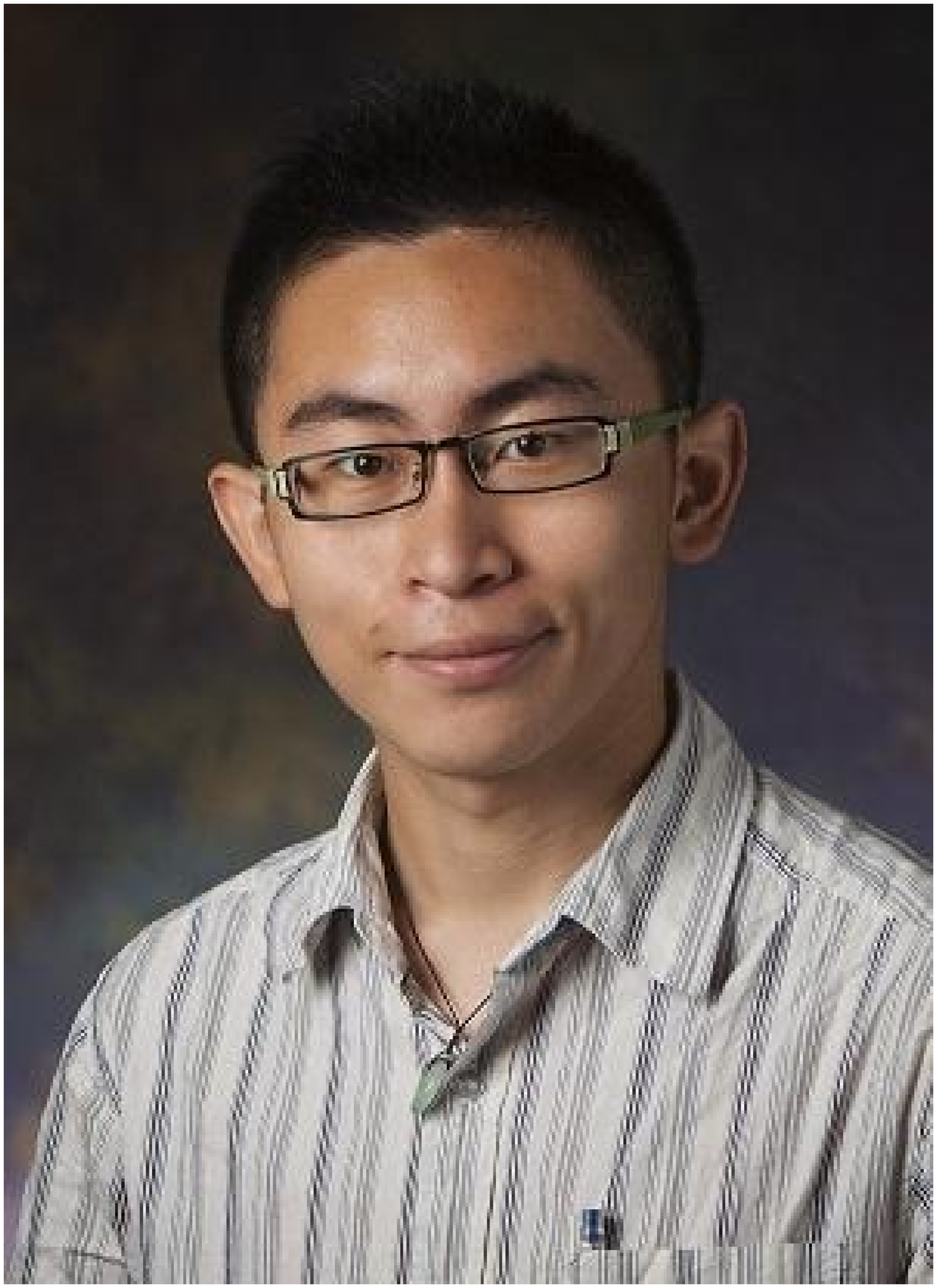}}]
{Jia-Bin
Huang} is an assistant professor in the Bradley Department of Electrical and Computer Engineering at Virginia Tech. He received the B.S. degree in Electronics Engineering from National Chiao-Tung University, Hsinchu, Taiwan and his Ph.D. degree in the Department of Electrical and Computer Engineering at University of Illinois, Urbana-Champaign in 2016. He is a member of the IEEE.
\end{IEEEbiography}

\vspace{-1.5em}
\begin{IEEEbiography}[{\includegraphics[width=1in,height=1.25in,clip,keepaspectratio]{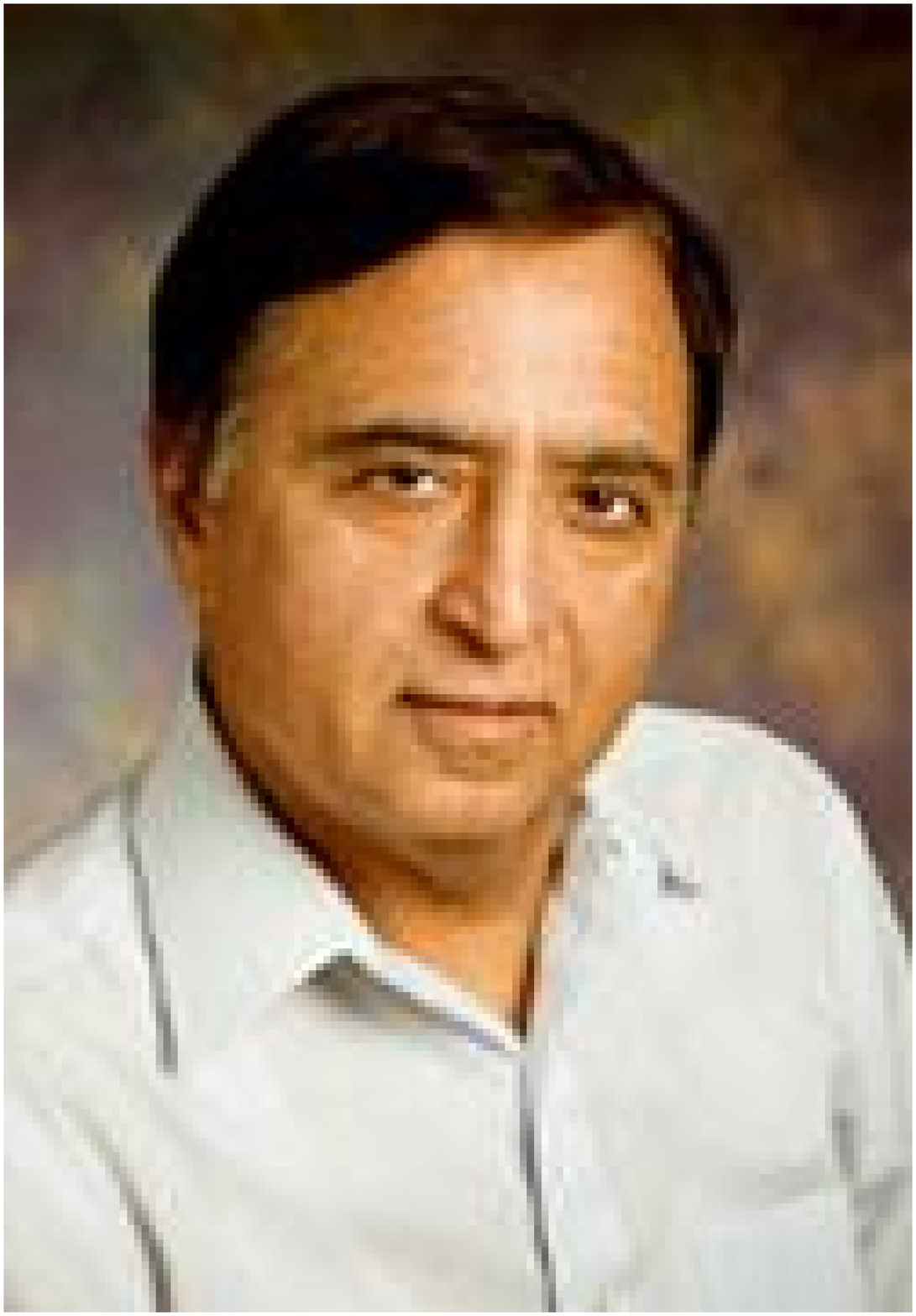}}]{Narendra
Ahuja} is the Donald Biggar Willet Professor with the
Department of Electrical and Computer Engineering,
University of Illinois at Urbana-Champaign, Urbana,
IL, USA.
He received the Ph.D. degree
from the University of Maryland, College Park, MD,
USA, in 1979.
%
%He was the Founding Director of the International Institute of Information Technology, Hyderabad, Hyderabad, India, where he continues to serve as a Director International.
%
He is a fellow of the American
Association for Artificial Intelligence, the International Association for
Pattern Recognition, the Association for Computing Machinery, the American
Association for the Advancement of Science, and the International Society for
Optical Engineering.
\end{IEEEbiography}

\vspace{-1.5em}
\begin{IEEEbiography}[{\includegraphics[width=1in,height=1.25in,clip,keepaspectratio]{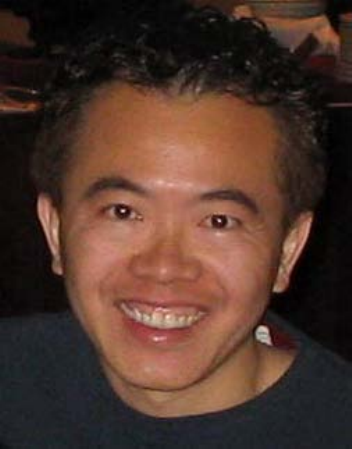}}]{Ming-Hsuan
Yang} is a professor in Electrical Engineering and
Computer Science at University of California, Merced. He received
the Ph.D. degree in computer science from the University of Illinois
at Urbana-Champaign in 2000. Prior to joining UC Merced in 2008, he
was a senior research scientist at the Honda Research Institute
working on vision problems related to humanoid robots.
Yang served as an
associate editor of the IEEE Transactions on Pattern Analysis and
Machine Intelligence from 2007 to 2011, and is an associate editor of the International Journal of Computer Vision, Image and Vision Computing and Journal of Artificial Intelligence Research.
He received the NSF CAREER award in 2012, the Senate Award for Distinguished Early Career Research at UC Merced in 2011, and the Google Faculty Award in 2009. 
He is a senior member of the IEEE and
the ACM.
\end{IEEEbiography}

\end{document}